\documentclass[runningheads]{llncs}
\usepackage{graphbox,graphicx}

\usepackage{tikz}
\usepackage{comment}
\usepackage{amsmath,amssymb} 
\usepackage{color}

\usepackage{bbm}
\usepackage{caption}
\usepackage[caption=false]{subfig}
\usepackage{wrapfig}
\usepackage{hyperref}
\usepackage{cleveref}

\usepackage[accsupp]{axessibility}  

\usepackage[width=122mm,left=12mm,paperwidth=146mm,height=193mm,top=12mm,paperheight=217mm]{geometry}
\newcommand{\etal}{\textit{et al}.}

\begin{document}
\pagestyle{headings}
\mainmatter

\title{When Deep Classifiers Agree: Analyzing Correlations between Learning Order and Image Statistics} 

\titlerunning{When Deep Classifiers Agree}
%
\author{Iuliia Pliushch\inst{1} \and
Martin Mundt\inst{2} \and
Nicolas Lupp\inst{1} \and
Visvanathan Ramesh\inst{1}}
\authorrunning{I. Pliushch et al.}
%
\institute{Goethe University Frankfurt, Germany \email{\{pliushch,vramesh\}@em.uni-frankfurt.de} \and
TU Darmstadt and hessian.AI, Germany \\
\email{martin.mundt@tu-darmstadt.de}}

\maketitle

\begin{abstract}
Although a plethora of architectural variants for deep classification has been introduced over time, recent works have found empirical evidence towards  similarities in their training process. It has been hypothesized that neural networks converge not only to similar representations, but also exhibit a notion of empirical agreement on which data instances are learned first. Following in the latter works' footsteps, we define a metric to quantify the relationship between such classification agreement over time, and posit that the agreement phenomenon can be mapped to core statistics of the investigated dataset. We empirically corroborate this hypothesis across the CIFAR10, Pascal, ImageNet and KTH-TIPS2 datasets. Our findings indicate that agreement seems to be independent of specific architectures, training hyper-parameters or labels, albeit follows an ordering according to image statistics.
\end{abstract}

\section{Introduction}

Can we make the learning process of a neural network more transparent? Is there any order in which a network is learning dataset instances? Various prior works posit that certain data instances are easier to learn than others. Arpit \etal \cite{Arpit2017} argue that neural networks prioritize simple patterns first during the learning process by analyzing the nature of the decision boundary. Mangalam and Prabhu \cite{Mangalam2019} extend the results, showing that neural networks first learn instances, classified correctly by shallow models like Random Forests and Support Vector Machines. Geirhos \etal \cite{Geirhos2019} and Shah \etal \cite{Shah2020} specify which patterns are simple for neural networks to learn, namely that they exhibit a \emph{simplicity bias} during training by relying on simple-but-noisy features like color and texture, instead of shape. Recently, Hacohen \etal \cite{Hacohen2020} have shown that there is agreement over learned examples throughout the entire learning process of a neural network, which is independent of initialization and batch-sampling and occurs also when changing different hyperparameter settings, like learning rate, optimizer, weight-decay and architecture. Given that certain examples are easier for a neural network to learn than others, it can be hypothesized that the difficulty of an example depends on dataset and image statistics to a larger degree than the learner itself. Hence, different neural networks would learn the same (easy) examples at the same time, despite random shuffling of the data during the training process in each epoch and strong regularization through noise of mini-batch stochastic gradient descent. In this spirit, we extend the analysis to a stricter agreement metric and not only replicate the presence of agreement, but also correlate it to chosen image statistics, in an attempt to clarify the reason why certain image examples are more difficult to learn than others. Our contributions are:
\begin{itemize}
    \item We design a strict instance-based agreement metric to quantify how learning progresses and similar data instances are classified correctly across neural networks.  
    \item With our metric, we replicate insights on classifier agreement from Hacohen \etal \cite{Hacohen2020} in terms of consistent neural network agreement across batch-sizes and architectures, as well as investigate the role of labels through a label randomization study.
    \item We select promising image statistics to correlate with neural network data instance agreement. We empirically corroborate correlations for metrics such as entropy, segment count, number of relevant frequency coefficients and summed edge strengths on popular image classification datasets: CIFAR10, Pascal, KTH-TIPS2b and ImageNet. 
\end{itemize}
 The analysis we conduct might help to better understand the tools - neural networks - we use for training. Further, as the next step, these insights could drive us towards the design of a \emph{learning curriculum} \cite{Kumar2010,Jiang2015,Bengio2009,Hacohen2019}, in which the dataset instances could be sampled according to a high correlation between observed agreement and the inspected metrics. We posit that this could lead to training speed ups and performance improvements. Our code is available at: \url{https://github.com/ccc-frankfurt/intrinsic_ordering_nn_training}.
\section{Problem statement and motivation} 
\label{problem_statement}
Let us first recall the general procedure of neural network training and the rationale behind it. Assume that we have a training set $X = {(x_{n}, y_{n})}^{N}_{n=1}$, where $x_{n}$ are our (i.i.d.) dataset instances, $y_{n}$ the corresponding labels and the number of dataset instances N. We also assume access to a similarly designed non-overlapping test set. We want to optimize a defined loss to measure and minimize the discrepancy between our network prediction and the ground truth. For that, we ideally want to integrate the loss L of our neural network - function $f_{\theta}$ with parameters $\theta$, over the dataset distribution:
\begin{equation}
    \int L(f_{\theta}(x),y) dP(x,y)
    \label{eq:approximation}
\end{equation}
In practice, we only have a limited amount of samples from the dataset distribution. Hence we compute an \emph{approximation}. Typically a noisy gradient estimate is leveraged for such empirical optimization, by presenting our data in mini-batches over several epochs $t$ and shuffling the dataset after every epoch, i.e. when the network has seen all the data (at least) once. Of course, when the networks have fully converged and learnt a sufficiently large amount of data, the dataset instances they have learnt trivially overlap. It is however not self-evident that different approximators would learn the data in a similar way or in other words that despite shuffling and mini-batch updating, neural networks would learn data in the same order. We first provide a definition for such agreement and then consider when networks necessarily start to agree on the dataset instances during learning.

\begin{figure}[t]
    \centering
    \includegraphics[width=0.9\textwidth]{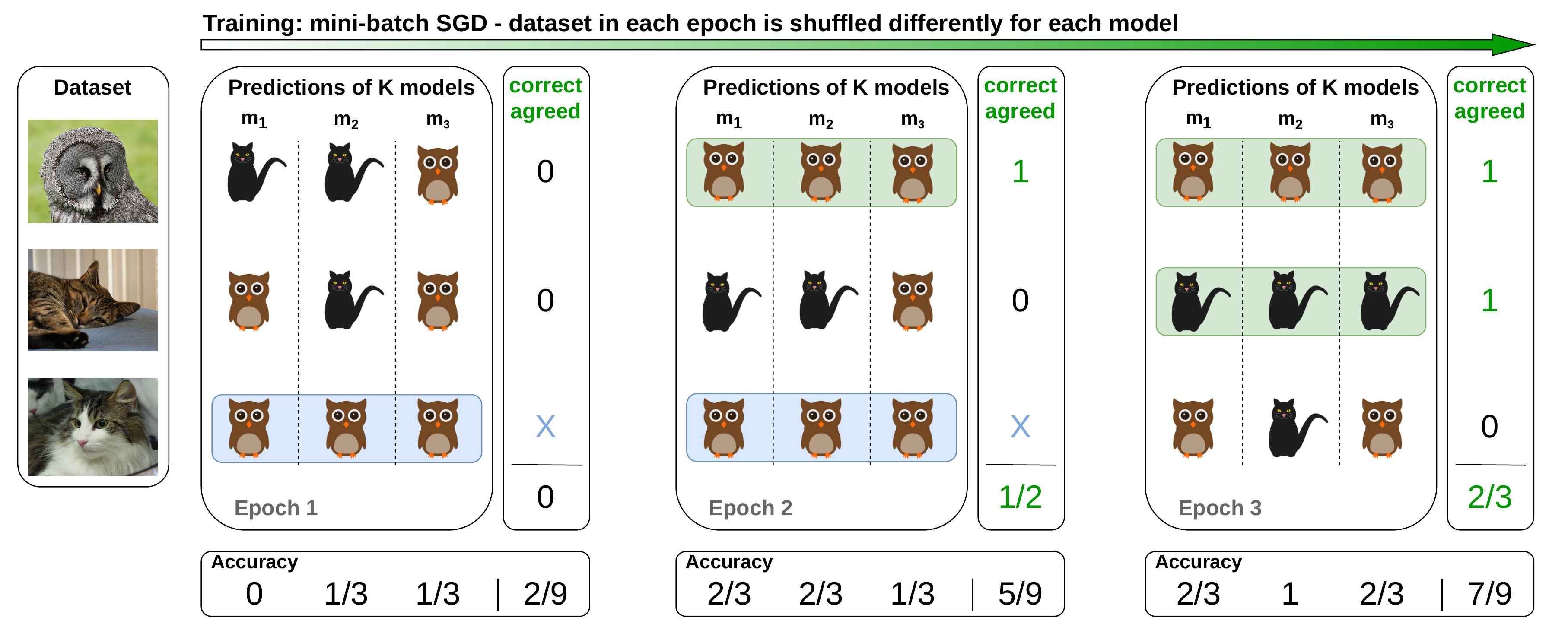}
    \caption{\textbf{Agreement visualization}: For each image the classification results are compared across networks and, in addition to \emph{average accuracy over networks}, \emph{true positive agreement} calculated, which is the ratio of images that all networks classify correctly per epoch to those that at least one network classifies correctly. Images are taken from ImageNet.}
    \label{fig:agreement}
\end{figure}
Hacohen \etal \cite{Hacohen2020} define one form of \emph{agreement} as ``the largest fraction of classifiers that predict the same label'' for the same data instance, as well as \emph{true positive agreement} as an ``average accuracy of a single example over multiple models''. In order to take the next step and try to quantify the difficulty of the images for training, we step back from agreement as an average and define \emph{(true positive)} agreement of an instance per epoch as an \emph{exact match}, such that all $K$ networks classify the same instance correctly in epoch $t$, where $K$ is the number of networks we have trained. \emph{True positive agreement per epoch} can then be computed as the sum of instances classified correctly by \emph{all} classifiers in that epoch, normalized by the sum of instances classified correctly by \emph{any} classifier in that epoch. Formally, true positive agreement $TPa$ per epoch $t$ can be defined as:
\begin{equation}
    TPa^{(t)}(x,y) = \frac{ \sum_{n \in N} \prod_{k \in K} \mathbbm{1}_{f^{(t)}_{k}(x_{n})=y_{n}}}{\sum_{n \in N} \max_{k \in K} \mathbbm{1}_{f^{(t)}_{k}(x_{n})=y_{n}}}
\end{equation}
During training, we now monitor the true positive agreement in every epoch for each training instance. Suppose that we train $K$ networks, as in \cref{fig:agreement}. In the first epoch, some models classify some dataset instances correctly (indicator function, $\mathbbm{1}_{f^{t}_{k}(x_{n})=y_{n}}$ being the condition for a prediction match), but for no instance it is the case that all models classify it correctly. In the second epoch - one dataset instance is classified correctly by all models. As during training more models classify instances correctly, if they learn the same instances first, then as soon as all models agree, it will be reflected in the agreement scores. Perfect agreement of 100\% can be reached if all models learn the same instances. So, agreement - fraction of correctly classified instances over all models compared to learned at all by at least one model - can be higher than average model accuracy per epoch, which is the fraction of correctly classified instances from the whole train set, averaged over the trained models. False positive agreement (all models misclassify an instance in the same way), as in the case of the partly white cat in the first two epochs (blue shaded box in \cref{fig:agreement}), is left out of the true positive agreement. However, one could, also analyze the false positive agreement in future work.

To assess, how trivial it is that neural networks agree \emph{before} convergence we consider the minimum possible agreement - the \emph{lower bound}. When do the models necessarily start to agree to learn the same instances? Let us assume that we train 2 networks and every one is 50\% correct at some epoch $t$. Then every network can learn the fraction of data that the other one did not. The same applies if we train 3 networks and the accuracy is 2/3 per network, because if we split the data into 3 portions, every network can learn 2 portions of the data such that there is not one portion common to all 3 networks. The same occurs for $K$ networks with accuracy being $\frac{K-1}{K}$. In all these cases the sum of errors $err^{e}_{k}$ all networks make is $K*\frac{1}{K}=1$ or $100 \%$. As the accuracy rises higher than that and the error gets lower, the networks will necessarily start to learn the fractions of data others are learning. Hence, the lower bound is 0\% when the sum of errors the networks make is bigger or equal to 100\%. The lower bound fraction, reported in the remainder of the paper in percent, can be defined as:
\begin{equation}
    LBa^{t}(x,y) = 1 - \min (\sum_{k \in K} \underbrace{1 - acc^{e}_{k}}_{err^{e}_{k}}, 1).
\end{equation}
In particular, the difference between agreement and lower bound shows us the portion of agreement which could not have been predicted on the basis of the lower bound alone. 

Lastly, since one of the datasets we test our hypotheses on is \emph{multilabel}, in this scenario we extend the definition of agreement and lower bound such that agreement and accuracy are calculated on the basis of \emph{exact match}, meaning that for all present and absent labels in an image, the prediction should exactly match: the presence and absence of a label should be predicted correctly for all labels.  This is a criterion which is non-forgiving: if one of the two labels has been predicted correctly, the exact match still classifies this image as wrongly predicted. On the other hand, it is a strong criterion to test agreement on, since what we are after is a \emph{full} and not \emph{partial} agreement. It is also a much stronger criterion than investigated in Hacohen \etal \cite{Hacohen2020}. Notably, our TP-agreement is thus also different from an \emph{observed agreement}, i.e.~the sum over instances given estimators classify correctly (true positives) and incorrectly (true negatives) divided by the total number of instances \cite{Hallgren2012}. Our choice to separate out true positive agreement is intended to avoid potential confusion, as the number of true positives and negatives is not equal during training and can complicate the assessment of reliability across estimators. The alternative Cohen's kappa, for example, suffers from the prevalence problem: it is non-representatively low in case of uneven frequencies of events. An alternative PABAK measure counteracting the prevalence bias \cite{Byrt1993} is a linear function of observed agreement, which takes both true positives and negatives into account. We provide a deeper discussion on alternatives in the appendix, in context of the preliminary results shown in the upcoming section.

\section{Agreement for different batch-sizes and architectures, as well as for random labels} 
\label{ablation}

To reiterate, the general procedure is to train several (in our case 5) networks on the same dataset, as well as track, which examples have been classified correctly per epoch per network. Upon training several networks for the same amount of epochs, \emph{agreement per epoch} is defined as the sum of images \emph{every} network classified correctly, normalized by the sum of images \emph{any} network classified correctly. We have seen on the example of the \emph{lower bound} that a quite high accuracy is necessary for the networks to unavoidably learn the same instances and that this relation depends on the number of networks trained: the more networks, the higher accuracy each of those has to obtain to start learning the same portion of the data. Hence, it is not self-evident that agreement does occur during neural network training.

If networks agree on what to learn,  which factors does this agreement depend on? To provide an intuition for above paragraphs, as well as to replicate prior insights obtained by Hacohen \etal \cite{Hacohen2020}, we first test for the overall presence of agreement. Then, we extend our experiments to different batch-sizes and architectures. Presence of agreement in all these diverse conditions supports the hypothesis that agreement is not dependent on the \emph{model}, but on the (sampled) dataset population itself in terms of the nature of the true unknown underlying distribution and the intrinsic characteristic of the difficulty of classification \cite{Caelen2017}. To further investigate whether the \emph{joint} dataset and label distribution leads to agreement, or whether it is independent of labels, we test the presence of agreement in case of random labels.

\begin{figure*}[th]
     \centering
     \subfloat[Without random labels \label{fig:cifar10_without_random_labels}]{
         \includegraphics[width=0.31\textwidth]{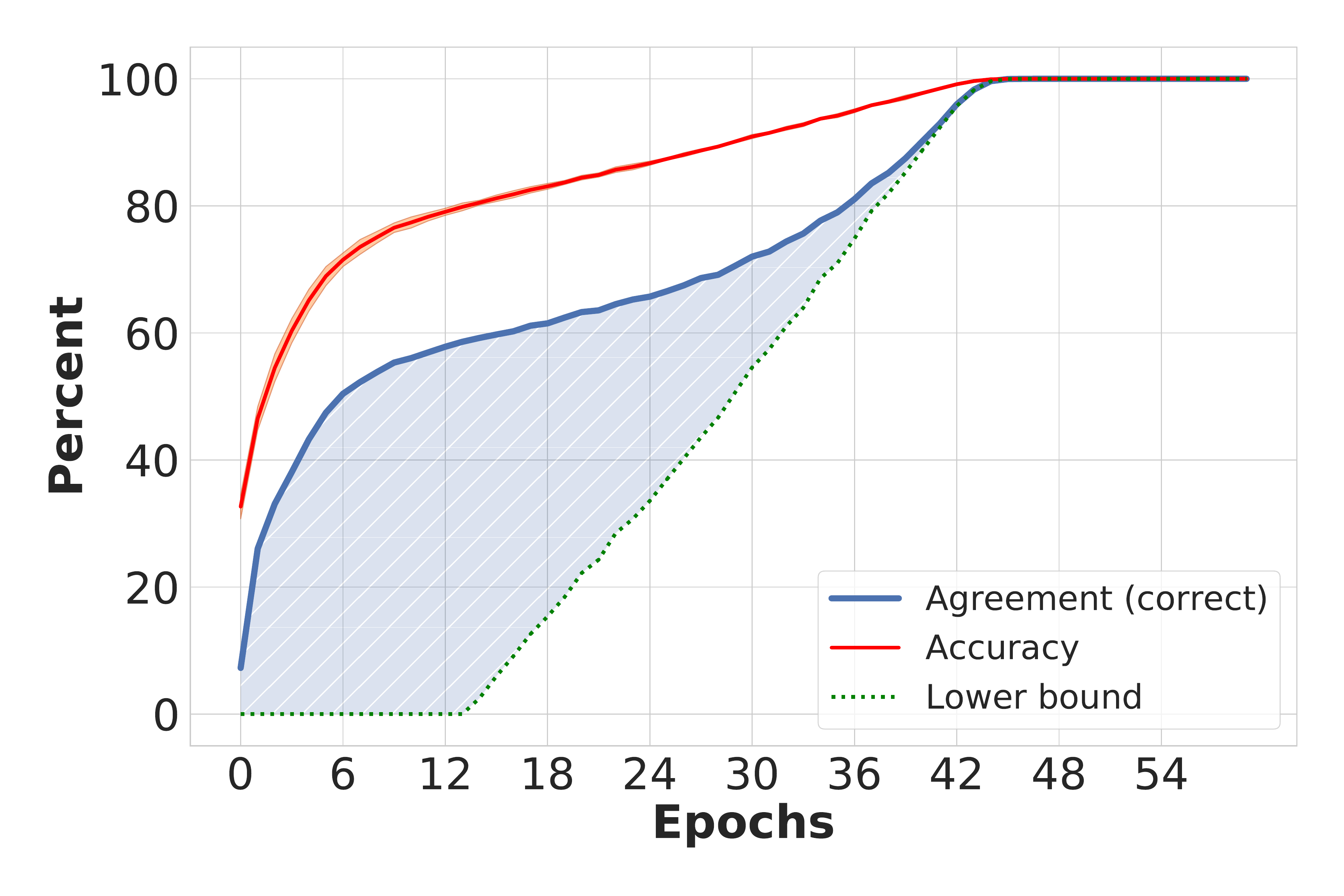}
    }
    \subfloat[With random labels \label{fig:cifar10_random_labels}]{
         \includegraphics[width=0.31\textwidth]{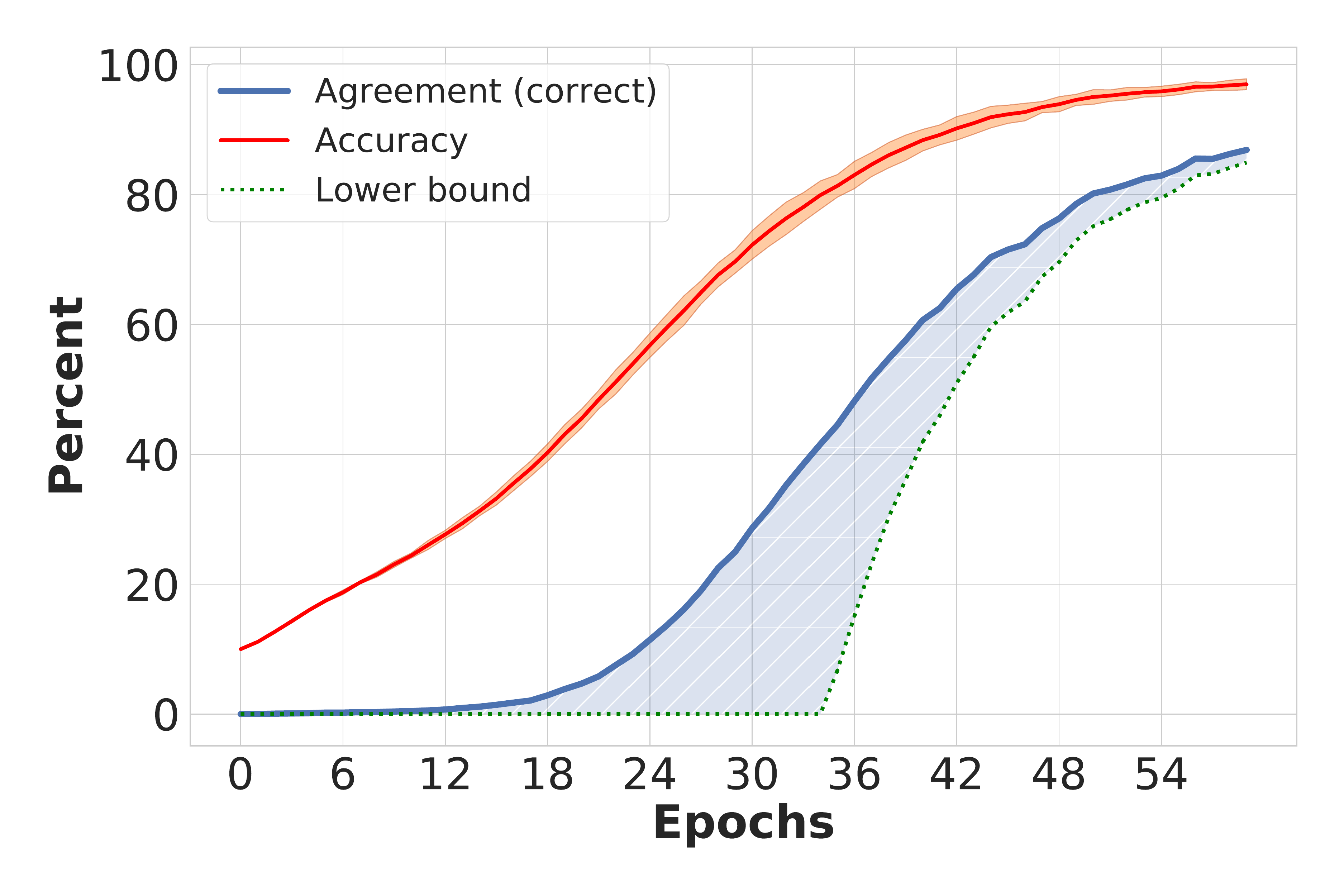}
    }
    \subfloat[Different architectures \label{fig:cifar10_archs}]{
         \includegraphics[width=0.31\textwidth]{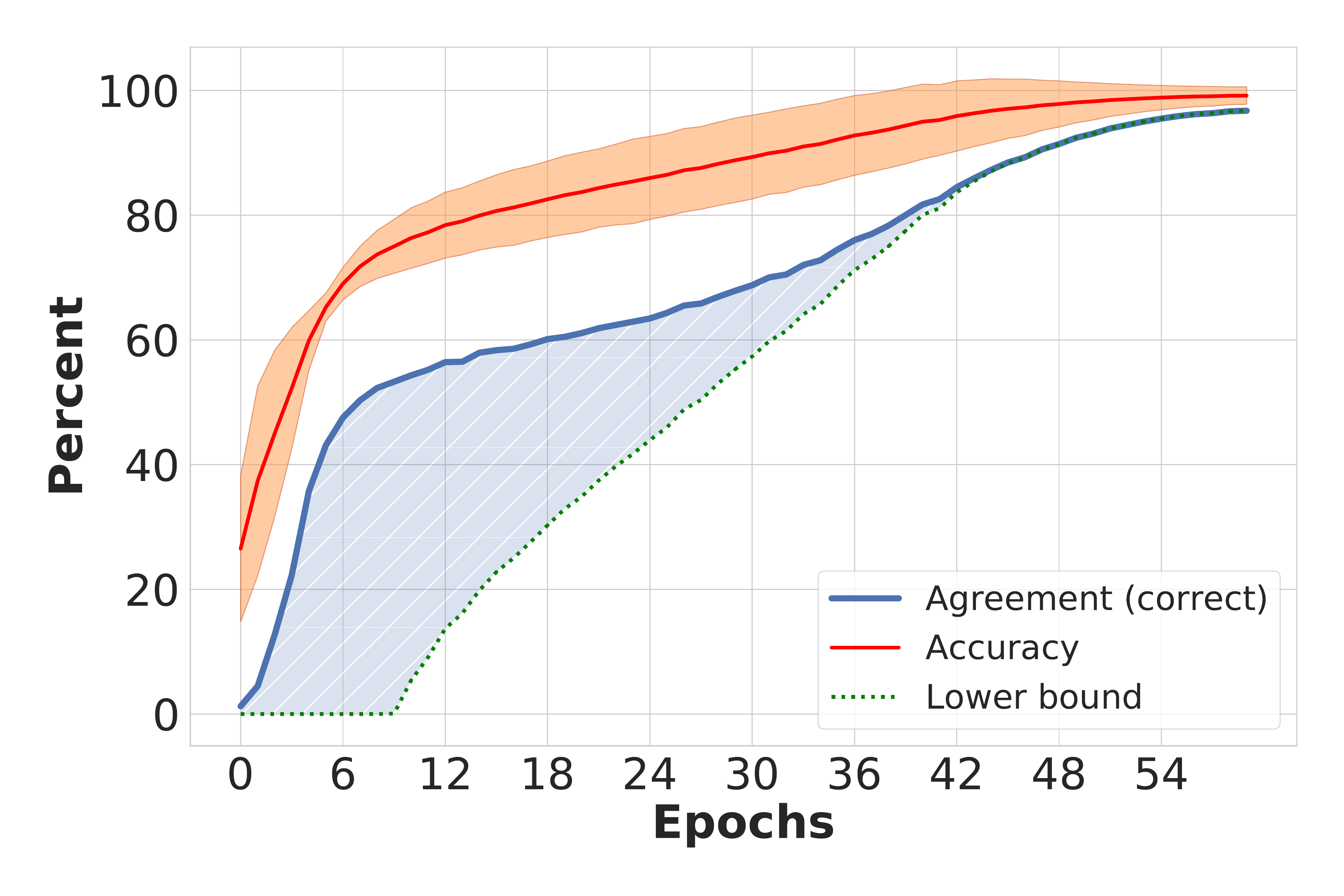}
    }
     \caption{Ablation study on \textbf{CIFAR10}: training DenseNet121 with and without randomization, as well as with different architectures: LeNet5, VGG16, ResNet50, DenseNet121. The \textbf{blue area} demonstrates the difference between agreement and lower bound. The \textbf{red area} is the epoch-wise standard deviation from average accuracy across trained networks.\label{cifar10_ablation}}
\end{figure*}

As hypothesized, mirroring prior hypothesis of Hacohen \etal \cite{Hacohen2020}, but using our strict agreement criterion, we found visually clear agreement during training of CIFAR10 \cite{Krizhevsky2009} on DenseNet121 \cite{Huang2017}, presented in \cref{fig:cifar10_without_random_labels}. Particularly in the first epochs, as the accuracy grows, the area between agreement and the lower bound, shaded in blue, is the most prominent. It shows that throughout training we observe growing agreement on the learned instances, i.e.~certain data instances are labelled correctly in earlier stages than others, which is most remarkable in the training epochs before networks converge and the accuracy plateaus.

Agreement persists also when training different architectures. We have chosen 4 diverse architectures, ranging from simple ones like LeNet5 \cite{LeCun1998} and VGG16 \cite{Simonyan2015} to the more complex ones like ResNet50 \cite{He2016} and DenseNet121 \cite{Huang2017}. Training CIFAR10 on them for the same amount of epochs, we could observe visually prominent agreement as well, see \cref{fig:cifar10_archs} (training details and additional plots in the appendix). The agreement is similar to that for the same architecture. The differences in learning speed are reflected in the standard deviation across different model accuracies, shaded in red. Note that accuracy deviation has been almost negligible when the same architectures have been trained.

We have also replicated the presence of agreement for different batch-sizes in the appendix. A comparison shows that agreement curves are similar for simpler architectures and smaller batch-sizes, as well as more complex architectures and larger batch-sizes, suggesting that model capacity and enhanced randomness when training with smaller batch sizes slow down the learning and, as an effect, agreement. 

Above experiments suggest that batch-size or architecture type are not the underlying causes for agreement during training. Next, we test whether it is the structure of the data itself, or the relationship between the training data and its human assigned labels that account for it. To test this hypothesis, we assess agreement under label randomization. State-of-the-art convolutional networks can fit random labels with ease \cite{Zhang2017}. Maennel \etal \cite{Maennel2020} argue further that during training with random labels an alignment between the principal components of data and network parameters takes place. Hence, if the dataset structure is responsible for agreement, it should be visible also in case of random labels. \cref{fig:cifar10_random_labels} supports this hypothesis. We observe that the accuracy grows at a slower pace and that agreement grows more slowly than during training with ground truth labels. This may reflect the fact that it takes longer for the network to start learning the dataset structure without label guidance. Nonetheless, there is still sufficient agreement once accuracy starts to rise. Interestingly, here we disagree with \cite{Hacohen2020}, who did not find agreement with randomized labels. 

To conclude, we have observed a clear gap between theoretical lower-bound and observed agreement, independence on semantic labels and architecture, even coherence in between them. This supports the existence of a fundamental core mechanism linked to more elemental dataset properties. To further support and strengthen our early results, we show in the appendix that observed trends of true positive agreement clearly surpassing the lower-bound persist even when comparint it to the \emph{expected random agreement}.  The latter is computed as the product of accuracies for a given epoch. It is based on the assumption that networks classify instances independently of each other, which does not seem to be the case, partly because, as we show, the dataset structure plays an important role.   In addition, we also compute the standard deviation on agreement for Pascal in the appendix,  showing that it is negligibly small.
\section{Dataset metrics} 
\label{dataset_metrics}
In the face of the insight that dataset properties are the probable cause of agreement, we proceed by choosing several diverse datasets, as well as dataset metrics to establish a correlation between them and training agreement. We have chosen the following datasets to validate our agreement hypothesis: a tiny-sized CIFAR10 \cite{Krizhevsky2009} for ablation experiments, a diverse dataset with objects differing in illumination, size and scale - Pascal Visual Object Classes (VOC) 2007 and 2012 \cite{Everingham2010,Everingham2015}, a large-scale ILSVRC-2012 (ImageNet) \cite{JiaDeng2009,Russakovsky2015}, as well as a texture dataset KTH-TIPS2b \cite{Caputo2005}. CIFAR10, Pascal and ImageNet have been gathered by means of search engines and then manual clean-up. In case of ImageNet, classes to search for were obtained from the hierarchical structure of WordNet and the manual clean-up proceeded by means of the crowdsourcing platform Amazon Mechanical Turk. On the example of the person category, Yang \etal \cite{Yang2020} elaborate that a dataset gathered in such a manner is as strong as the semantic label assumptions and distinctions on which it is based, the quality of the images obtained using search engines (e.g. lack of image diversity), as well as the quality of the clean-up (annotation) procedure. In distinction to the above three datasets, though KTH-TIPS2b is a quite small dataset, it has been carefully designed, controlling for several illumination and rotation conditions, as well as varying scales.

\textbf{Prior work}: What surrogate image statistics may be used to study the agreement between classifiers on the order of learning? Let us look at some prior works in this direction. First, some dataset properties make learning difficult in general, namely the diverse nature of the data itself: the difficulty of assigning image categories due to possible image variation, e.g. rotation, lighting, occlusion, deformation \cite{Pinto2008}. Second, learning algorithms also show preferences for particular kinds of data. Russakovsky \etal \cite{Russakovsky2013} analyze the impact onto ImageNet classification and localization performance of several dataset properties on the image and instance level, like object size, whether the object is human- or man-made, whether it is textured or deformable. Their insights are consistent with the intuition that object classification algorithms rely more on texture and color as cues than shape. They further argue that object classification accuracy is higher for natural than man-made objects. Hoiem \etal \cite{Hoiem2012} analyze the impact of object characteristics on errors in several non-neural network object detectors, coming to the conclusion that the latter are sensitive to object size and confusion with semantically similar objects.

In an attempt to mimic the decisions of a human learner, several image features have been related to image memorability and object importance. Isola \etal\cite{Isola2014a} investigate memorability of an image as a stable property across viewers and its relation to basic image and object statistics, like mean hue and number of objects in the image. Spain and Perona \cite{Spain2008}, on the other hand, establish the connection between image object features and the importance of this object in the image, which is the probability that a human observer will name it upon seeing the image. Berg \etal \cite{Berg2012} extends this analysis to encompass semantic features like object categories and scene context. 

Knowing which image cues the learning process is sensitive to gives room for improving it. Alexe \etal \cite{Alexe2012} design an objectness measure generic over classes from image features (multi-scale saliency, color contrast, edge density and superpixels straddling) which can be used as a location prior for object detection. Extending the latter work, Lee and Grauman \cite{Lee2011} use this objectness measure, as well as an additional familiarity metric (whether it belongs to a familiar category) to design a learing procedure which first considers easy objects. Liu \etal \cite{Liu2011} fit a linear regression model with several image features, like color, gradient and texture, to estimate the difficulty of segmenting an image. In \cite{Vijayanarasimhan2009} an active learner is designed, which partly on the basis of edge density and color histogram metrics, proposes which instances to annotate and estimates the annotation cost for the multi-label learning task. Notably the question of whether the networks learn the same examples first is different from the question whether they learn the same representations \cite{Wang2018,Li2016}, but if the former is correct, then the latter is more probable. 

\textbf{Our choice}: Several works have correlated basic image statistics to various human-related concepts, like memorability \cite{Isola2014a}, importance \cite{Spain2008} or image difficulty \cite{Ionescu2016}, or directly attempted to find out the influence of such metrics onto object classification \cite{Russakovsky2013,Alexe2012}. Inspired by these approaches, we have chosen 4 image statistics to correlate agreement to: segment count \cite{Felzenszwalb2004}, (sum of) edge strengths \cite{Isola2014b}, (mean) image intensity entropy \cite{Frieden1972,Skilling1984} and percentage of coefficients needed to reconstruct the image based on the DCT coefficient matrix \cite{Ahmed1974}. First 3 metrics are shown in \cref{fig:img_metrics} (DCT coefficients matrix in the appendix).

The choice of the first two is inspired by \cite{Ionescu2016}, who correlate several image properties, including segment count and (sum of) edge strengths, to the \emph{image difficulty score}, defined as a normalized response time needed for human annotations to detect the objects in the image. Their hypothesis is that segments divide the image into homogeneous textural regions, such that the more regions - the more cluttered an image might be (and the more difficult an object to find). The same line of reasoning goes for the sum of edge strengths: the more edges, the more time might be needed to get a grasp of the image. If the way humans search for objects in the image corresponds to some degree to the way a neural networks learns to predict their presence, segment count and the (sum of) edge strengths might be predictive for the agreement during training. 
Mean image entropy and DCT coefficients have been chosen for a similar line of reasoning. Both entropy and DCT coefficients, similar to edge strengths and segment count, provide a measure of variability: how uniform the variation in image pixels is in the case of entropy and how many (vertical and horizontal) frequencies are needed to describe an image in the case of DCT coefficients. Recently, \cite{Ortiz-Jimenez2020} have analyzed how certain dataset features relate to classification, stating, for instance, that for MNIST and ImageNet the decision boundary is small for low frequency and high for high frequency components. In other words, the classifier develops a strong invariance along high frequencies. Hence, frequency-related information might influence agreement.

\begin{figure}[t]
    \centering
    \subfloat{%
        \includegraphics[width=.23\linewidth]{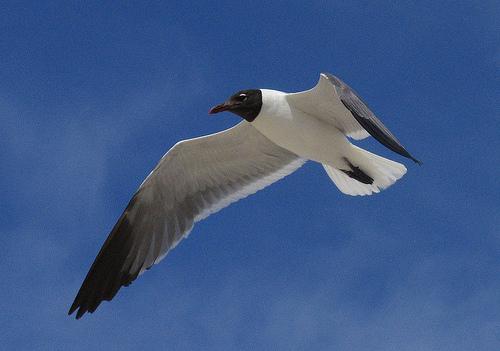}
    } 
    \subfloat{%
        \includegraphics[width=.23\linewidth]{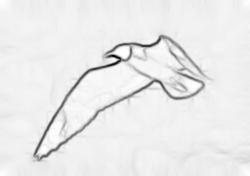}
    }
    \subfloat{%
        \includegraphics[width=.23\linewidth]{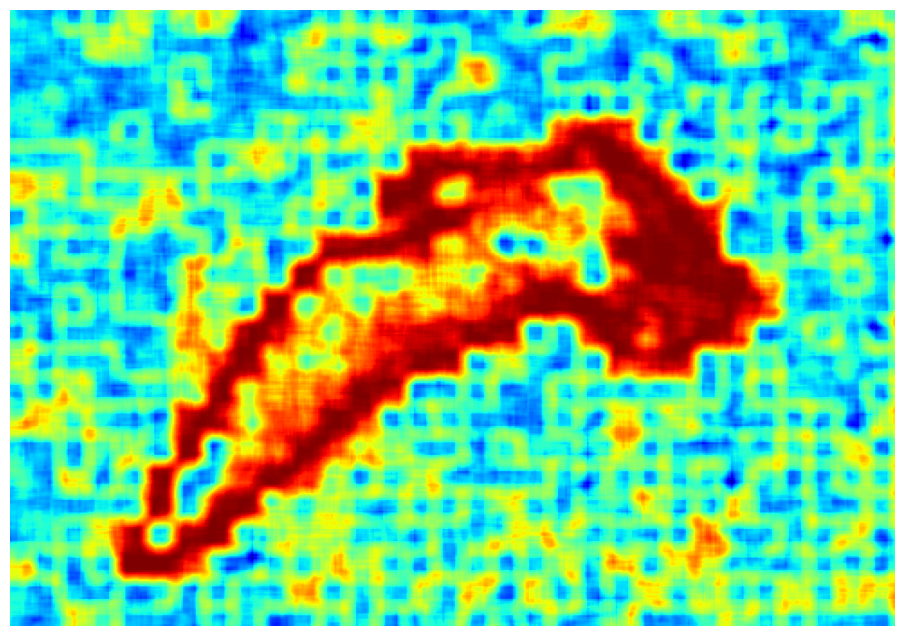} 
    }
    \subfloat{%
        \includegraphics[width=.23\linewidth]{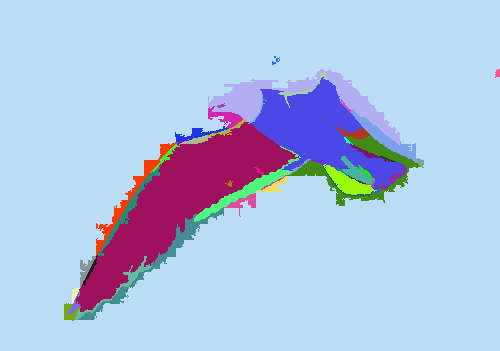}
    }\\
    \centering
    \setcounter{subfigure}{0}
    \subfloat[Image]{
        \includegraphics[width=.23\linewidth]{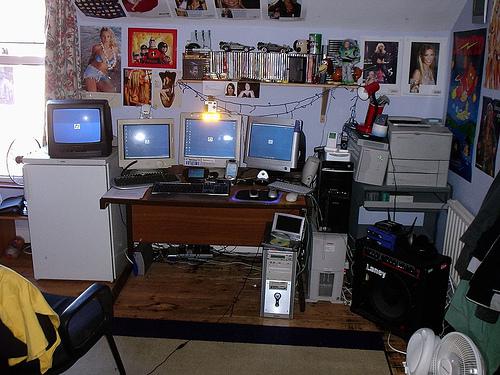}
    }
    \subfloat[Edge strengths]{%
         \includegraphics[width=.23\linewidth]{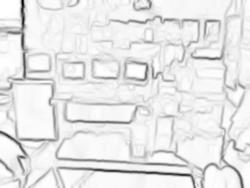}
    }
    \subfloat[Entropy]{%
         \includegraphics[width=.23\linewidth]{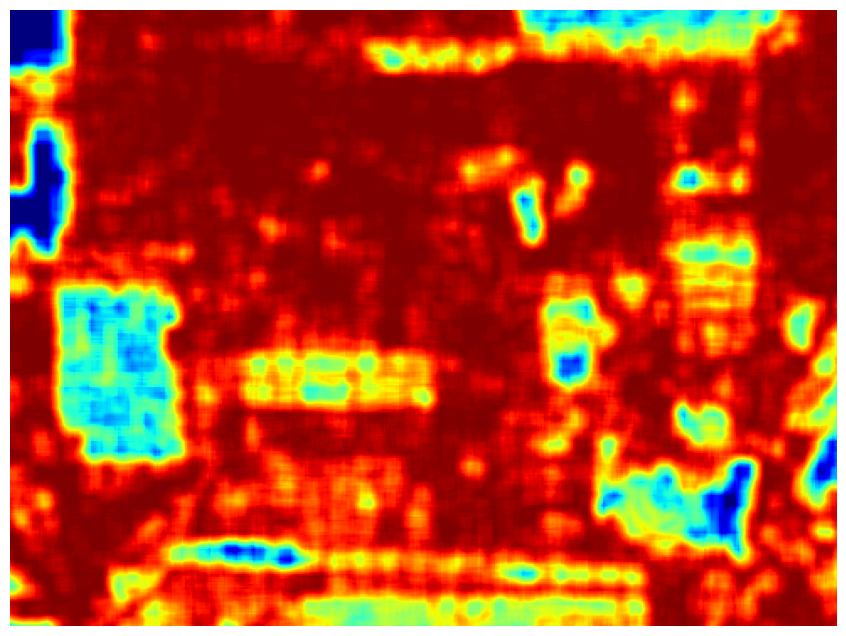}
    }
    \subfloat[Segments]{%
         \includegraphics[width=.23\linewidth]{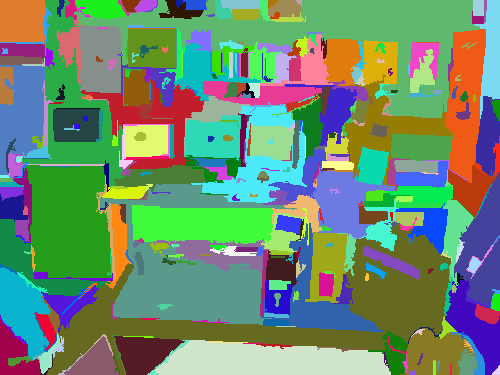}
    }
	\caption{Visualization of selected metrics on Pascal examples. For intuition, we have chosen an ''easy'' and a ''difficult'' image according to Ionescu \etal \cite{Ionescu2016}. Implementation details can be found in the appendix.\label{fig:img_metrics}}
\end{figure}
In addition, for Pascal, which contains further annotations, we considered the \emph{number of object instances} (since it is a \emph{multilabel} dataset such that several label instances may be present in the same image) and \emph{the bounding box area} (ratio of the area taken by objects divided by the image size), as well as the \emph{image difficulty scores} described above, computed by \cite{Ionescu2016}. KTH-TIPS2b, is constructed in such a way that for each texture type, there are 4 texture samples, each of which varies in \emph{illumination}, \emph{scale} and \emph{rotation}, which we also considered. Finally, for the CIFAR10 test set, Peterson \etal \cite{Peterson2019} have computed soft labels, reflecting human uncertainty that the given target class is in the image, to test the hypothesis that networks trained on soft labels are more robust to adversarial attacks and generalize better than those trained on hard one-hot labels. \emph{Soft label entropy} shows weak negative correlation with test agreement (see Fig. 5 of the appendix).

\section{Do basic image statistics correlate with training learning dynamics?}
Our initial investigation of section \ref{ablation} suggests that neither the labels, nor the precisely chosen neural architecture or optimization hyper-parameters seem to be the primary source for agreement, eliminating all factors of \cref{eq:approximation} other than the data distribution itself.  As such,  we investigate the question \emph{``do image statistics provide a sufficient description in correlation for agreement?''} on four datasets: Pascal \cite{Everingham2010}, CIFAR10 \cite{Krizhevsky2009}, KTH-TIPS2b \cite{Caputo2005}, and ImageNet \cite{JiaDeng2009}.

\begin{figure}
\centering
 \includegraphics[width=0.65\textwidth]{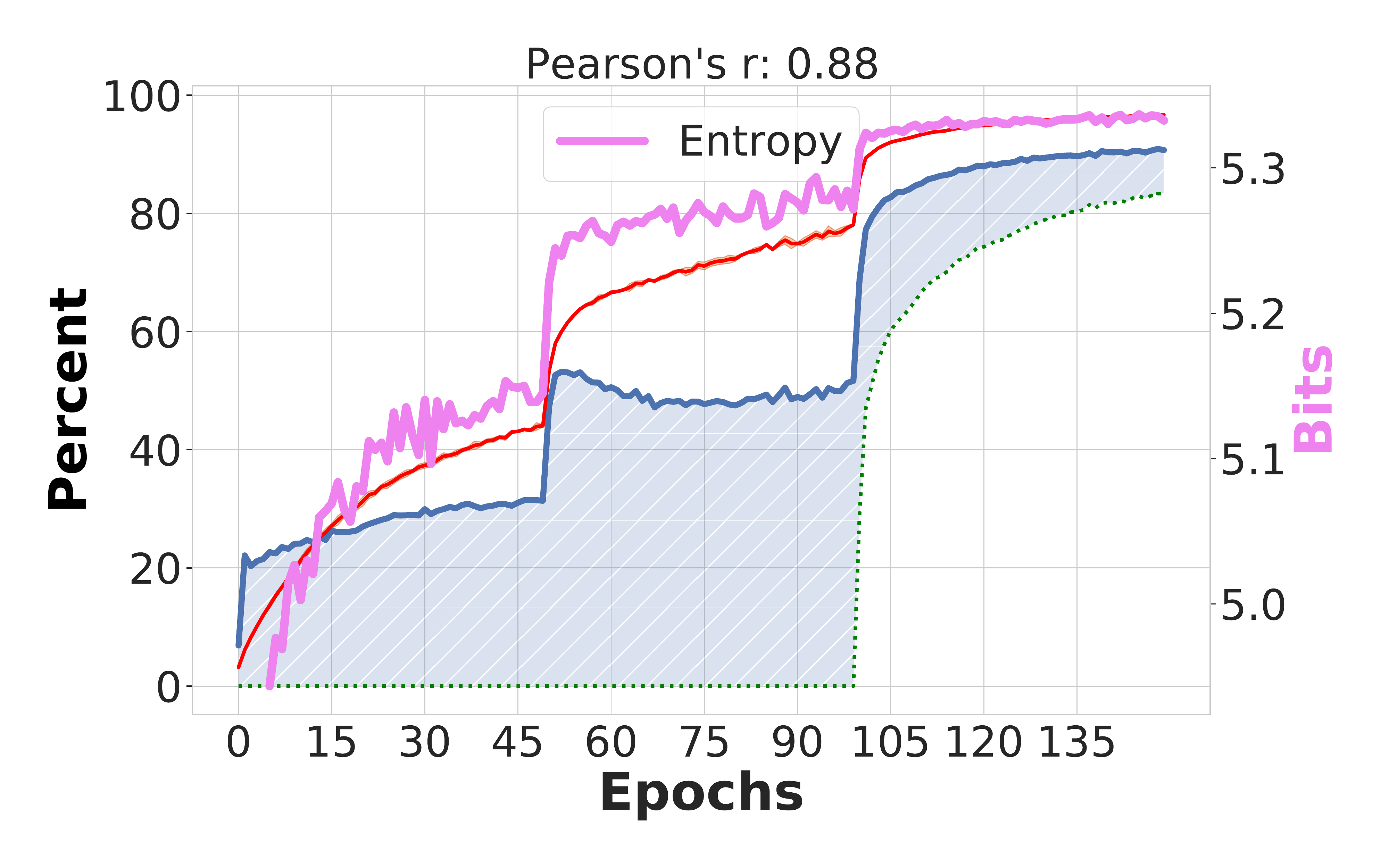}
 \caption{\textbf{Pascal DenseNet}: Agreement visualization on \emph{train set}. In addition to accuracy (\textbf{red}), agreement (\textbf{blue} curve) and its difference to lower-bound  (shaded \textbf{blue} area) (left y-axis), dataset metric values are shown in \textbf{purple} (right y-axis). \emph{Pearson's correlation coefficient} between agreement (\textbf{blue} curve) and metric values (in \textbf{purple}) quantitatively supports the visible correlation.}
 \label{pascal_densenet_entropy}
\end{figure}
To make the full upcoming results easier to follow, we first start by visualizing and discussing one example metric, namely entropy on Pascal in \cref{pascal_densenet_entropy}. In a given epoch, the agreement (blue) and accuracy (red) curves can be compared to the average entropy of the agreed upon instances (in purple). Again, the shaded blue area accentuates the difference between lower bound and agreement. Since we have chosen a step-wise learning rate scheduler for training, in order to reach roughly 50\% exact match accuracy on the test set, the step-wise learning is reflected in the accuracy and agreement curve.  As a new addition to our previously shown figures, we now notably also observe a strong positive correlation with the dataset entropy metric.  This correlation between network agreement and the entropy of the correspondingly agreed upon instances is further quantified through a high Pearson correlation coefficient of 0.88. 

\begin{figure}
\centering
\subfloat[\textbf{Pascal}  \label{pascal_agreement}]{%
        \includegraphics[width=0.33\textwidth]{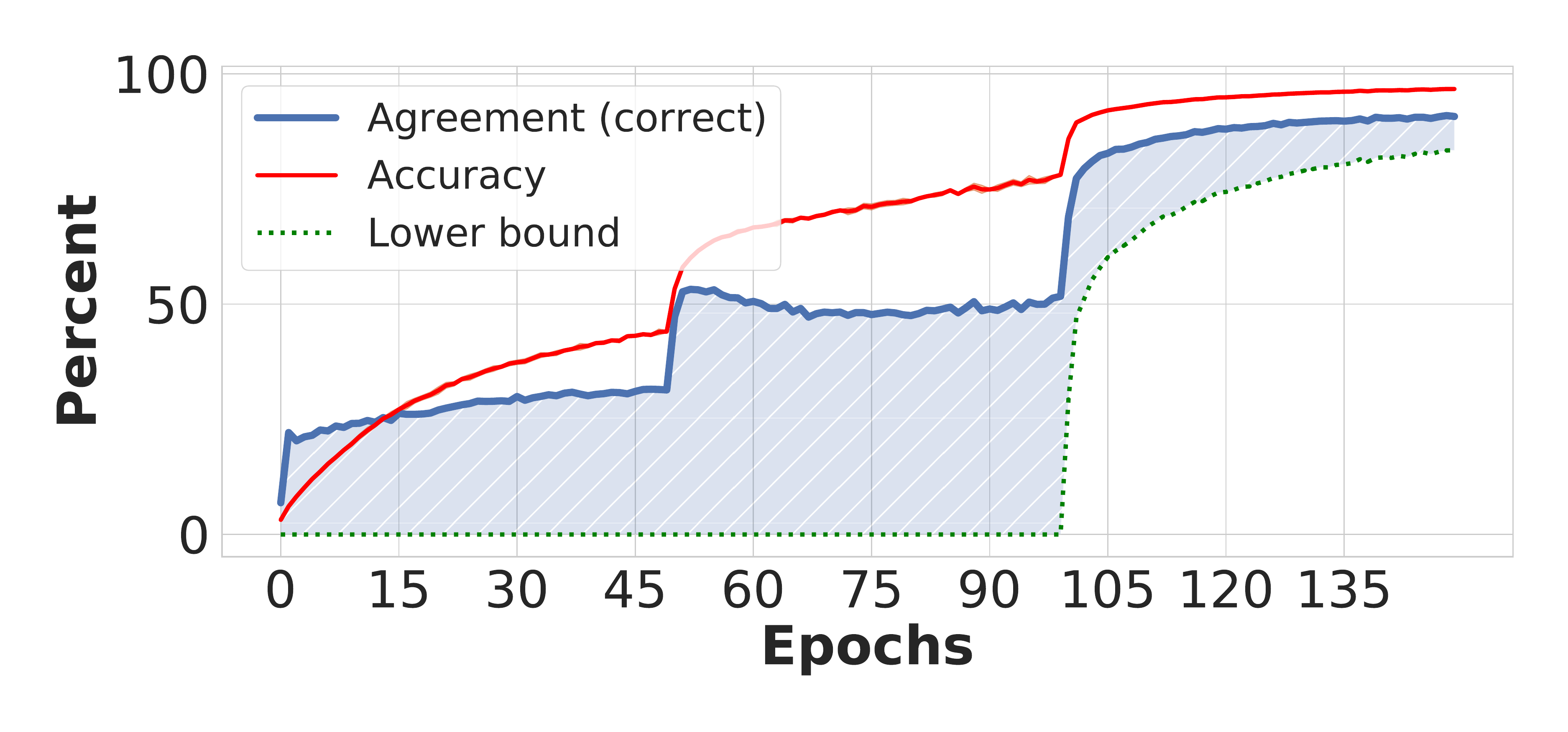}
    } 
    \subfloat[\textbf{KTH-TIPS2b}  \label{kth_tips_agreement}]{%
        \includegraphics[width=0.33\textwidth]{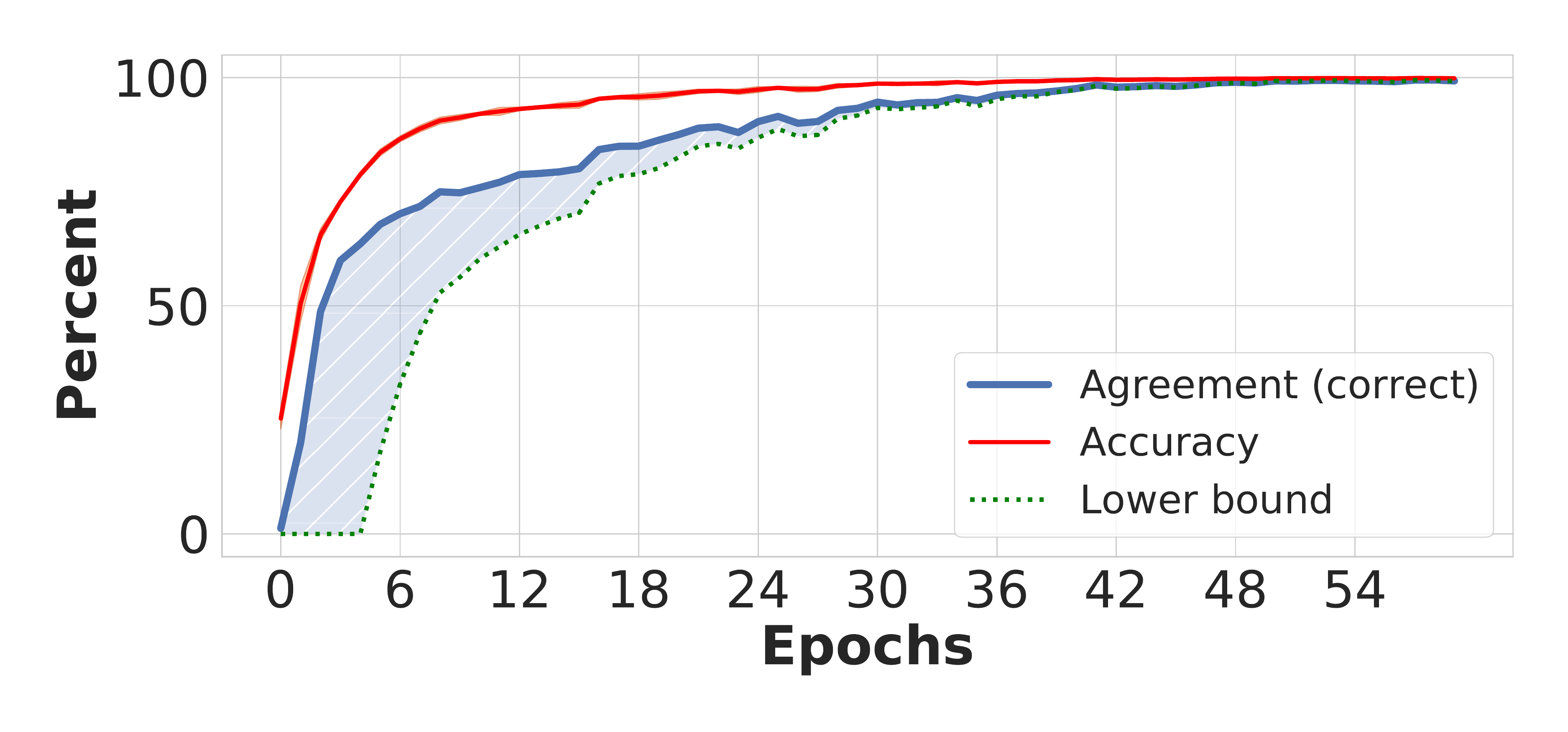}
    }
    \subfloat[\textbf{ImageNet} \label{imagenet_agreement}]{%
        \includegraphics[width=0.33\textwidth]{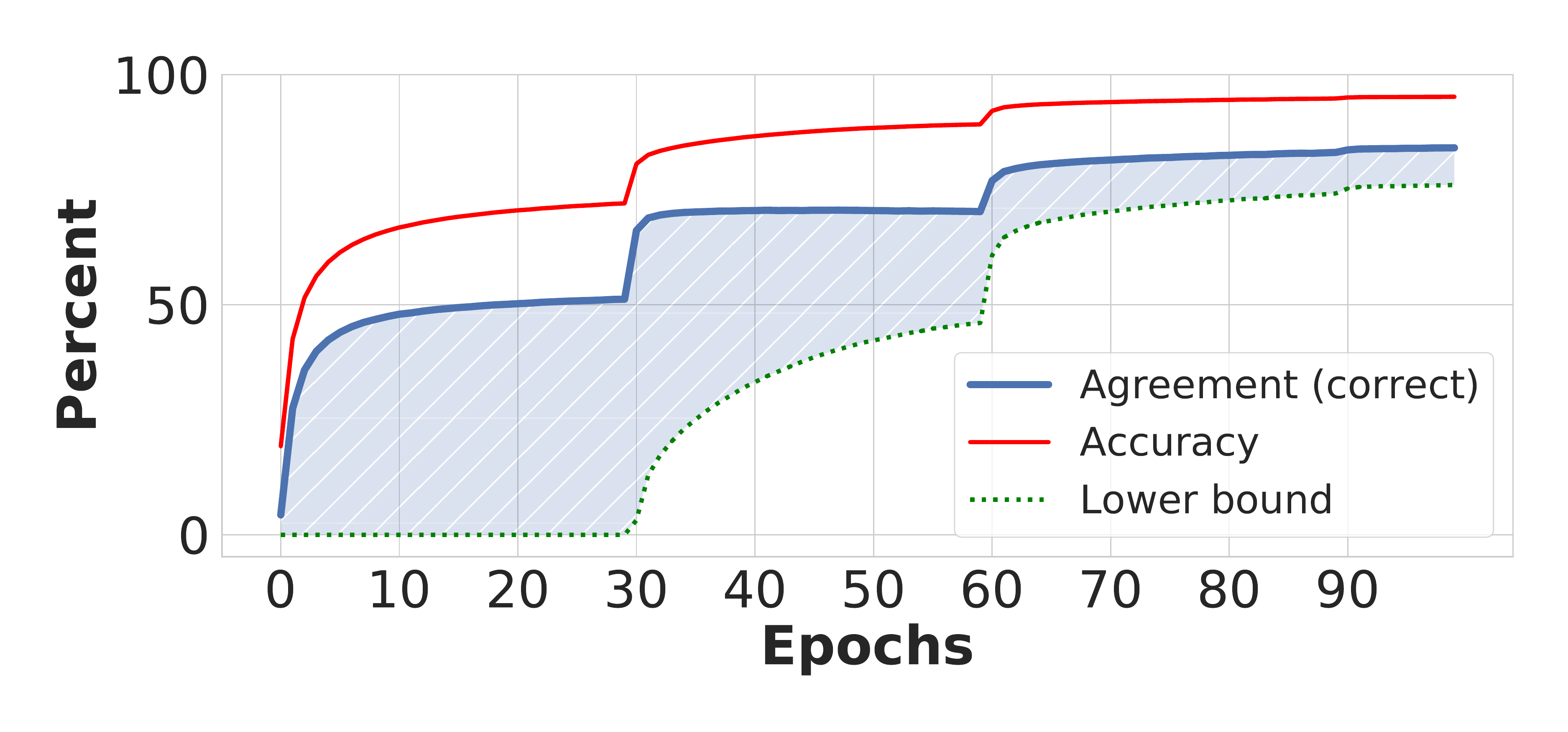}
    }
    \caption{\textbf{On DenseNet}: Agreement, accuracy and lower bound as in \cref{pascal_densenet_entropy}}
    \label{agreement_datasets}
\end{figure}
\begin{figure}
\centering
    \begin{minipage}[l]{0.38\textwidth}
    \subfloat[\textbf{Pascal DenseNet}: Correlation between agreement (in blue) to per epoch averaged  dataset metric values for correctly agreed upon training instances,  on a shared x-axis (in analogy to the purple curve in \cref{pascal_densenet_entropy}).  Pearson correlation coefficient between agreement and the metric is provided.  It is in rectangular brackets if the 2-tailed p-value is $>=$ 0.001, see appendix for details.  \label{pascal_densenet_metrics}]{%
        \includegraphics[width=0.99\textwidth]{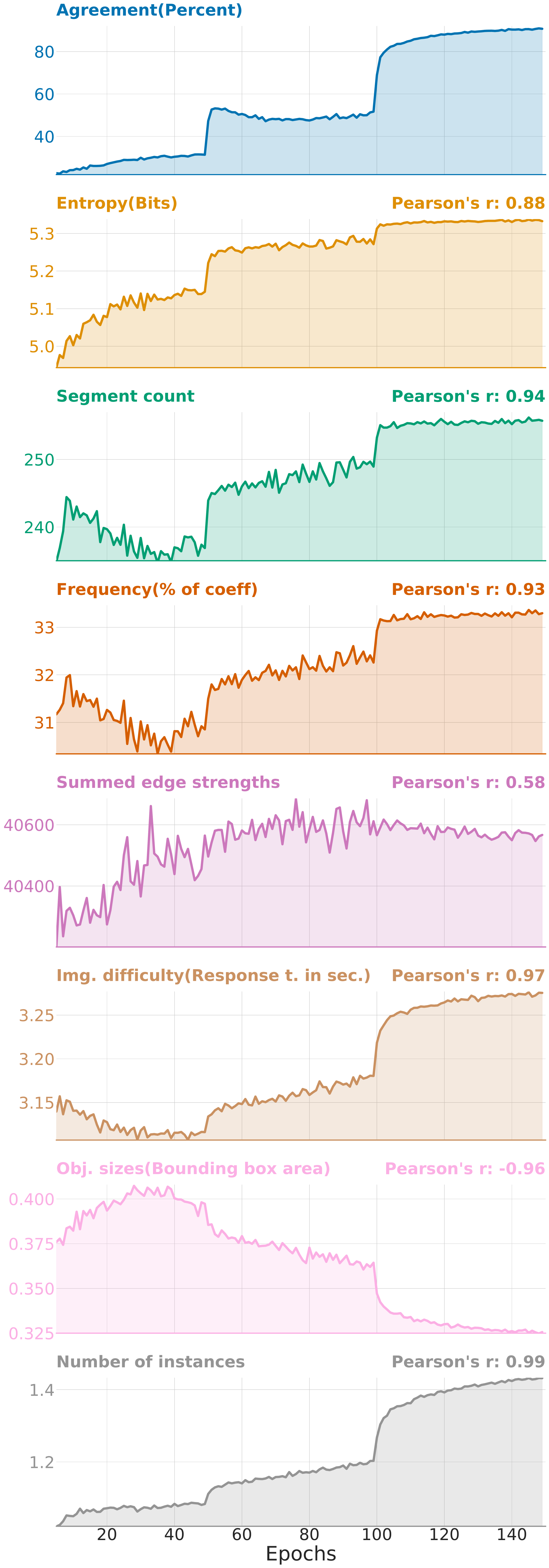}
    }
    \end{minipage} \hfill
    \begin{minipage}[r]{0.38\textwidth}
    \subfloat[\textbf{KTH-TIPS2b DenseNet}: Agreement and metric values in analogy to \cref{pascal_densenet_metrics} \label{kth_tips_densenet_metrics}]{%
        \includegraphics[width=0.99\textwidth]{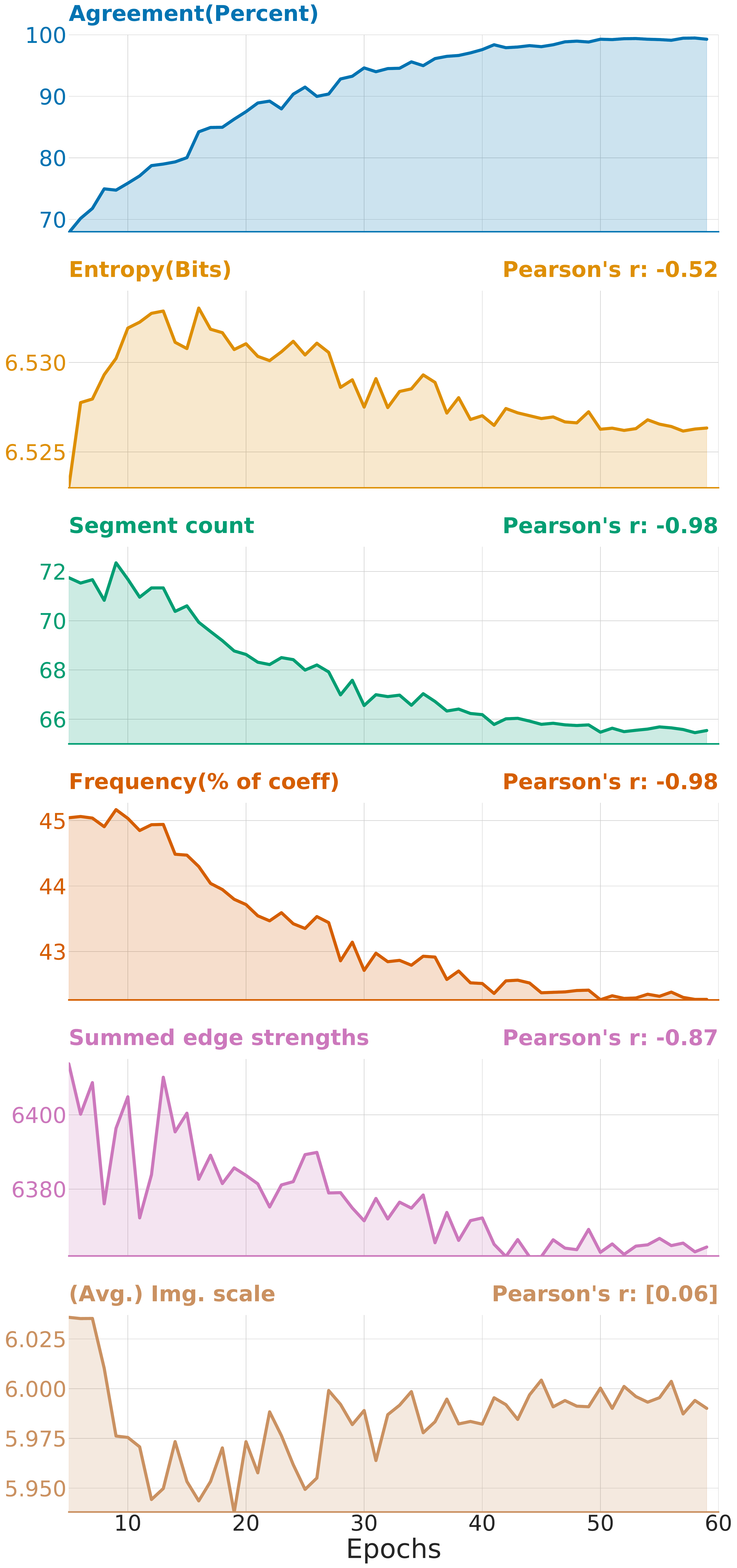}
    }
    \newline
    \subfloat[\textbf{KTH-TIPS2b DenseNet}: For illumination and rotation types, the percentage learned per category per epoch has been visualized separately for immediate comparison. \label{kth_tips_densenet_categorical}]{%
        \includegraphics[width = 0.99\textwidth]{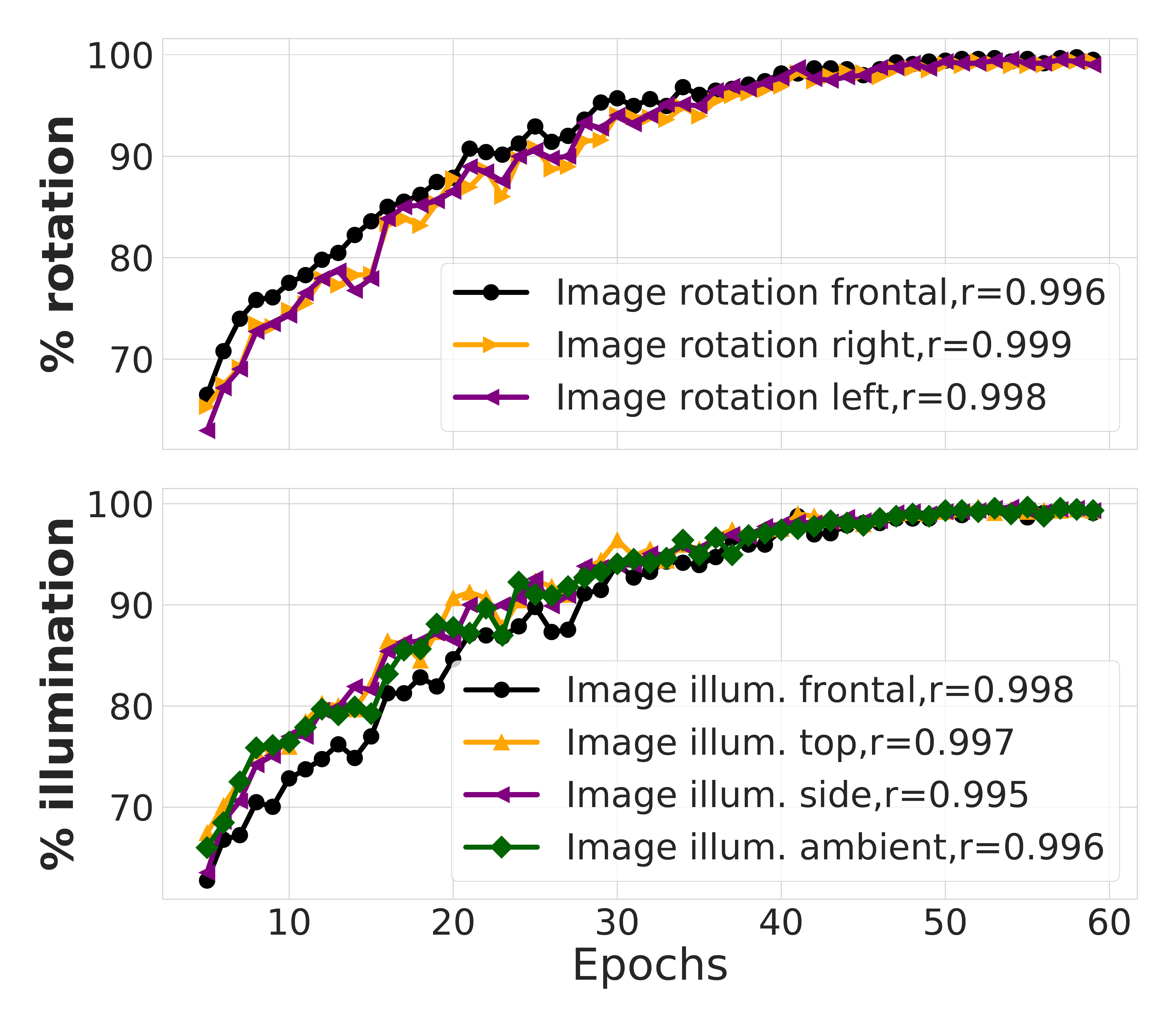}
    }
    \end{minipage}
    \caption{Visualization of correlations between agreement and dataset metrics on \emph{train sets} for \textbf{Pascal} and \textbf{KTH-TIPS2b}. \label{results_grid_pascal_kth}}
\end{figure}
\begin{figure}
\centering
    \subfloat[\textbf{ImageNet DenseNet}: In analogy to \cref{pascal_densenet_metrics}. \label{imagenet_densenet_metrics}]{%
            \includegraphics[align=t,width=0.38\textwidth]{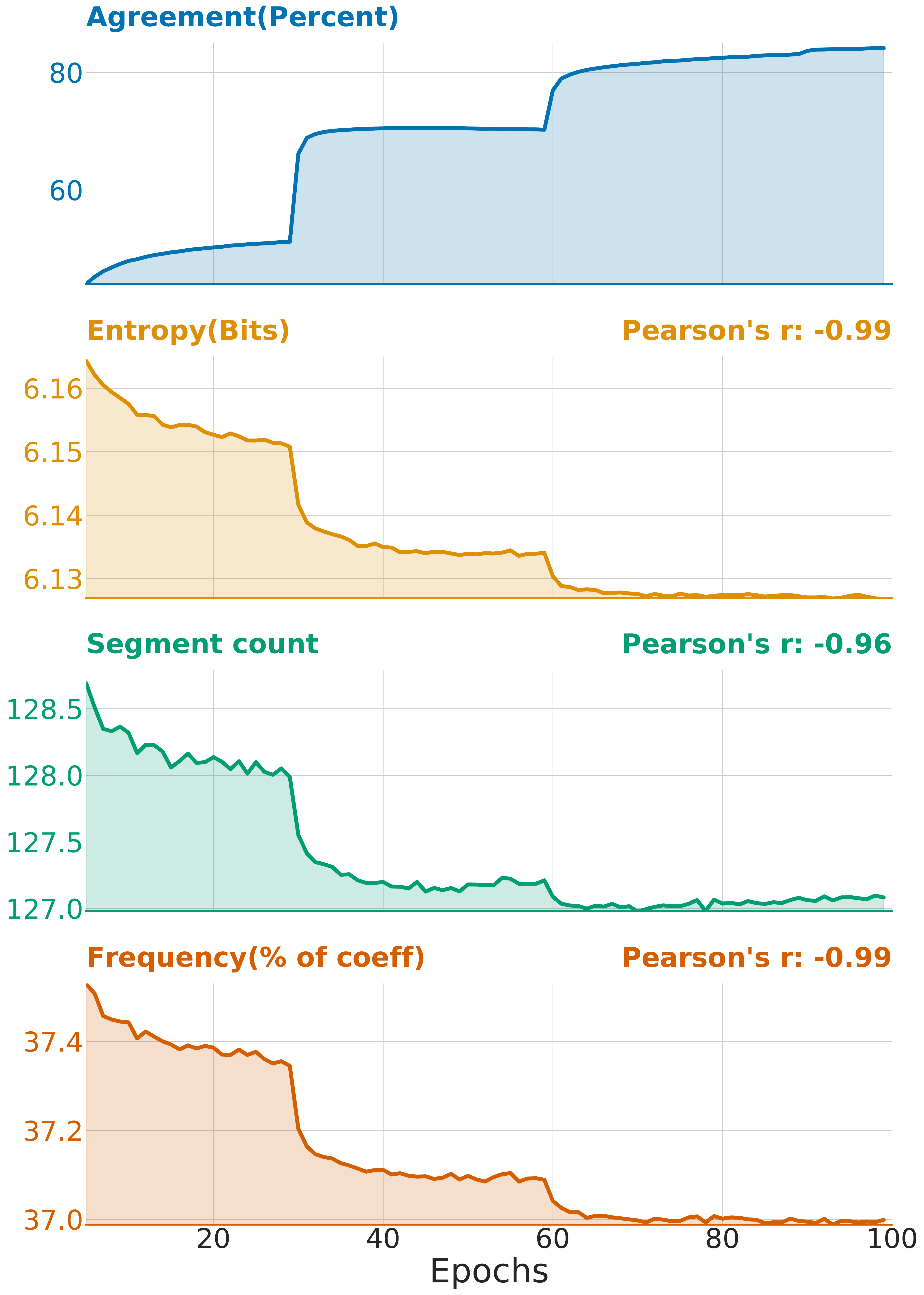}
        }\hfill
    \subfloat[\textbf{CIFAR10 DenseNet}: In analogy to \cref{pascal_densenet_metrics}. \label{cifar_densenet_metrics}]{%
        \includegraphics[align=t,width=0.38\textwidth]{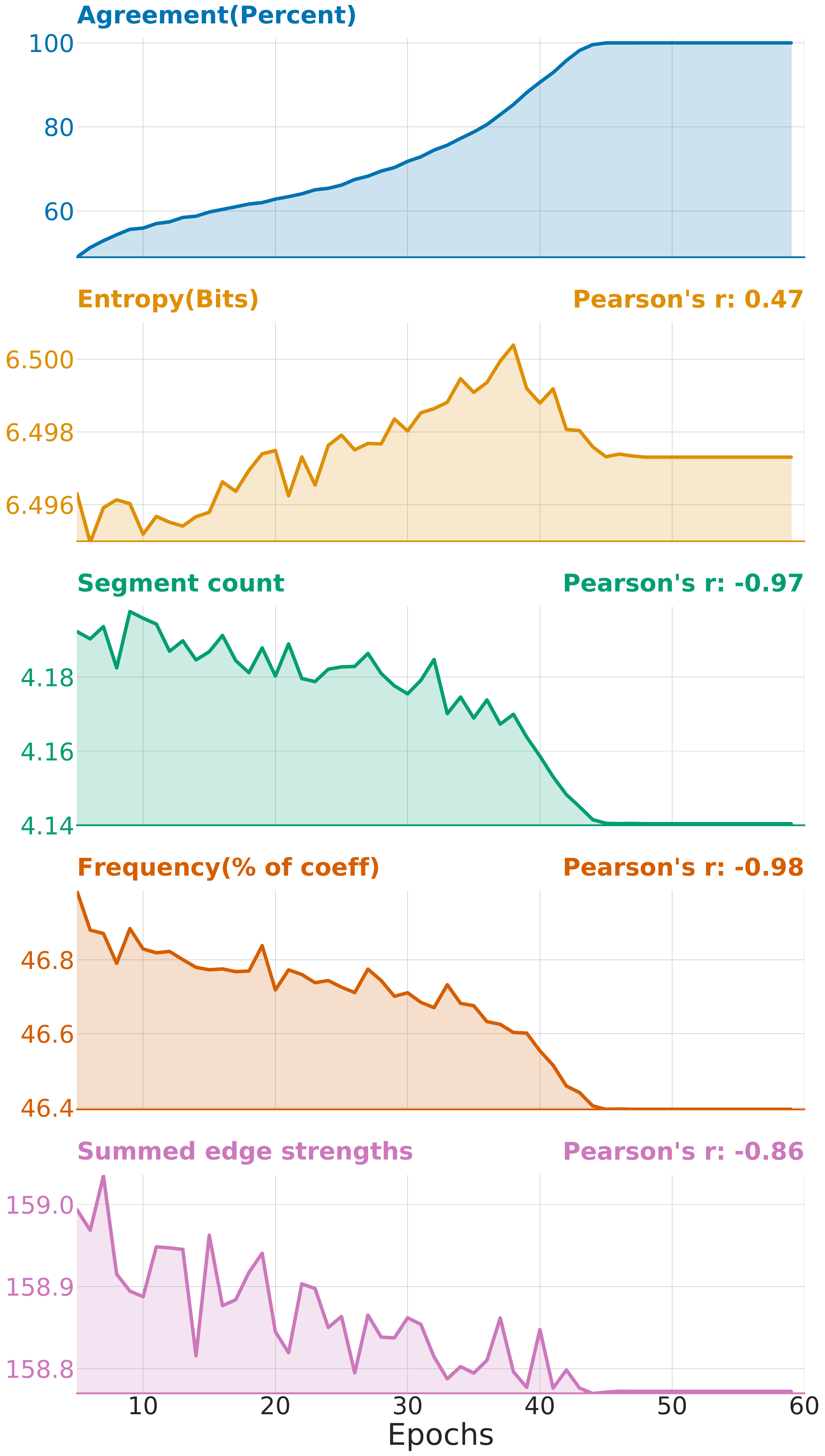}
    }
    \caption{Visualization of correlations between agreement and dataset metrics on \emph{train sets} for \textbf{ImageNet} and \textbf{CIFAR10}. \label{results_grid_imagenet_cifar}}
\end{figure}

We continue our analysis in this form for all other mentioned metrics and dataset combinations in \cref{results_grid_pascal_kth} and \cref{results_grid_imagenet_cifar}. To better evaluate the shown correlations, we have visualized the distribution (in the form of a histogram) of the values for each dataset metric on the train sets in the appendix.  Since the metrics fluctuate a lot in the first epochs due to predictions being primarily random, we omit plotting them for the first 5 epochs.  For reference,  we provide the agreement, accuracy and lower-bound curves in the style of our previous figures in \cref{agreement_datasets}.  To clarify potential correlations,  these curves are then followed by visualizations and quantitative Pearson's r values of only agreement to the set of chosen image metrics in \cref{results_grid_pascal_kth} and \cref{results_grid_imagenet_cifar}. 

For the \textbf{Pascal} dataset,  shown in \cref{pascal_densenet_metrics}, the correlations between agreement and average dataset metrics are apparent (apart from sum of edge strengths). The correlations suggest that as the models learn, they in progression first learn dataset instances with lower entropy, segment count, number of significant DCT coefficients and number of object instances, meaning that less labels are present in the same image. The distribution for the number of instances (in the appendix) shows that the number of labels in most images is only 1 (equivalent to the single label classification), less for 2 and 3. This is reflected in the correlation, which goes up from 1.1 to 1.4. There is also a correlation between image difficulty (how much time humans need to find the objects in the image) and agreement, which is consistent with the other metrics, namely that easy examples are learnt first. This result is further supported by the bounding box size, where we see an inverse correlation, such that large objects are learnt first, in agreement with insights presented in the dataset metrics section. 
However, note that particularly for segment count, frequency coefficients and "image difficulty" the dataset metric curve first goes down, before it reverses its direction for the remainder of epochs. What we observe is that when accuracy is low and the model is trying to find an optimal trajectory to learn, there are more random fluctuations, due to the stochasticity of the learning process. In the appendix we show the correlations for Pascal ResNet with the same training setup. There, the trend of metric values first going down in first epochs (until the agreement approximatively reaches 20\%) and then up is even more pronounced for some metrics.

Correlations are also present on \textbf{KTH-TIPS2b}, visualized in \cref{kth_tips_densenet_metrics}. However, for this texture dataset, the tendency observed in Pascal is reversed, such that entropy, summed edge strengths, segment count and frequency percentage are inversely correlated with the dataset metric. In addition to these metrics, the way in which the dataset has been designed allows to extract additional ones, namely several illuminations, rotations and scales. The corresponding correlations are visualized in \cref{kth_tips_densenet_categorical}. For illumination and rotation, instead of building an average over the metric values per epoch, we calculate for each illumination kind and rotation direction the fraction of values agreed upon, normalized by all metric values of that type. For frontal illumination, for example, we count the instances that models agree on and divide by the number of instances of that type in the train set. We observe that texture patterns \emph{illuminated} from the front are agreed on slower than other illumination types, while texture patterns \emph{captured} from the front are agreed on quicker than other rotation directions. Correlation for texture scale seems absent.

We have seen that correlations on Pascal and KTH-TIPS2b diverge. In \cref{imagenet_densenet_metrics} we also see that on \textbf{ImageNet} correlations, albeit on a small scale, are present and they seem to be congruent with those on KTH-TIPS2b rather than Pascal. Namely, first images with higher segment count, entropy and number of relevant frequency coefficients are learned. Pascal and ImageNet are datasets of objects, while KTH-TIPS2b of texture. Why the difference? If we follow recent hypotheses on neural networks exhibiting simplicity bias and primarily using texture to discriminate \cite{Shah2020,Geirhos2019}, then correlation directions on KTH-TIPS2b and ImageNet would be the same (which our results indicate). On both Pascal and ImageNet random-crop is used to get train images of the same size, which has been argued by \cite{Hermann2019} to enhance texture bias. The way objects are presented in Pascal and ImageNet differs however, for ImageNet objects are centered and not much background is present, which is not the case for Pascal, making latter classification more challenging, but also effects of random-crop 
eventually different.

On \textbf{CIFAR10}, the direction of correlations is more consistent with KTH-TIPS2b and ImageNet (see \cref{cifar_densenet_metrics}). The correlations are marginal, however, given that metric value range is very small. The presence of agreement with little correlation can also be an indication that there are some other metrics which explain it, though it may be some humanly non-interpretable noise patterns \cite{Ilyas2019}.

\section{Discussion on limitations and prospects}
Although agreement is present for \emph{every} dataset, the previous sections have exposed a dataset dependency with respect to how precisely correlations between agreement and dataset metrics manifest. This means that the order, in which dataset instances are learned, is to a large degree independent of network parameters, but instead dependent on the general dataset statistics. Hence, in order to better understand the learning process of neural networks, agreement should be computed for carefully designed datasets, for which the generative process and dataset statistics are well known. There should be further analysis into suitable dataset metrics, which reflect the learning process.

Another research direction is to assess generalization and analyze the agreement on unknown \emph{test sets}. Though these are often assumed to follow the same distribution as their training counterparts, it is practically not always the case. Hypothetically, if the subset statistics are sufficiently similar, then agreement and metrics correlations should similarly manifest, as we exemplary show in the appendix.

Last but not least, as mentioned in the introduction, the insights on agreement could help us design a more efficient \emph{learning curriculum}, for which an appropriate pacing function \cite{Hacohen2019} should be chosen with care.

\section{Conclusions and outlook}
In this paper we have defined a new notion of agreement, characterising the learning process of neural networks in a more detailed way. We have demonstrated agreement on the train (and test set) for CIFAR10, Pascal, ImageNet and KTH-TIPS dataset. We have further correlated agreement on these datasets to several image statistics, in an attempt to explain why neural networks prefer to learn dataset instances in the way they do. Our results have shown several positive and negative correlations to dataset metrics, though different for each dataset. For future research is left the opportunity to test the results on further datasets, to test the correlation for further metrics, as well as to design curricula for training neural networks based on these insights.

\begin{wrapfigure}[5]{l}{0.21\textwidth}
\centering
\includegraphics[width=0.1\textwidth]{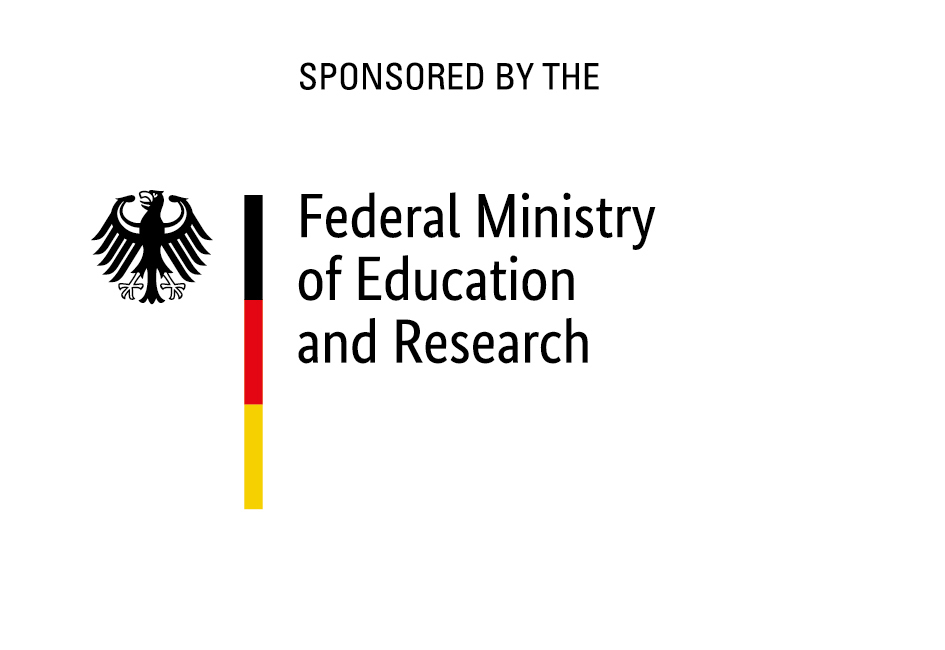} 
\includegraphics[width =0.1\textwidth]{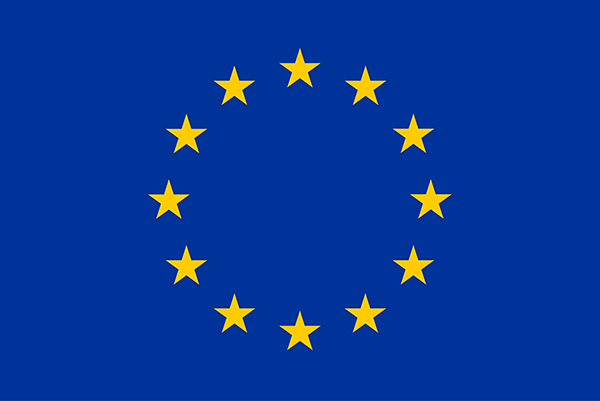}
\end{wrapfigure}
\section{Acknowledgements} This work was supported by the German Federal Ministry of Education and Research (BMBF) funded project 01IS19062 "AISEL" and the European Union's Horizon 2020 project No. 769066 "RESIST".

\clearpage
\title{Appendix: When Deep Classifiers Agree: Analyzing Correlations between Learning Order and Image Statistics} 

\titlerunning{When Deep Classifiers Agree}
%
\author{Iuliia Pliushch\inst{1} \and
Martin Mundt\inst{2} \and
Nicolas Lupp\inst{1} \and
Visvanathan Ramesh\inst{1}}
\authorrunning{I. Pliushch et al.}
%
\institute{Goethe University Frankfurt, Germany \email{\{pliushch,vramesh\}@em.uni-frankfurt.de} \and
TU Darmstadt and hessian.AI, Germany \\
\email{martin.mundt@tu-darmstadt.de}}

\maketitle

\appendix
\addcontentsline{toc}{section}{Appendix}
\renewcommand{\thesubsection}{\Alph{subsection}}

In this supplementary material we provide additional details, experimental setup and descriptions for the employed methodology of the main body. The structure is as follows:

\begin{itemize}
 \item[\textbf{A.}] Experimental setup and training hyper-parameters.
 \item[\textbf{B.}] Additional plots for initial experiments on different batch-sizes and architectures, as well as expected random agreement and a discussion of an alternative agreement definition.
 \item[\textbf{C.}] Additional discussion of the reasons for the weakness of correlations for CIFAR10, as well as experimental results for Pascal trained on ResNet, omitted in the main body. In this context also an explanation of the relationship between Pearson correlation coefficient and the p-value.
 \item[\textbf{D.}] More precise description of the computed dataset metrics, as well as additional visualization thereof.
 \item[\textbf{E.}] Visualization of dataset metrics histograms
\end{itemize}
\section{Experimental details}
\label{experimental}
Since our aim is to analyze the training process on the original images, we did not use data augmentation techniques, apart from random cropping for train and center cropping for test images on Pascal, ImageNet and KTH-TIPS2b due to the difference in size between images in these datasets. For Pascal and ImageNet, we resize the smaller size to 256 and randomly crop to obtain patches of width and height 224 pixels \cite{Krizhevsky2009,Simonyan2015}. For KTH-TIPS2b, in analogy we resize to 200 and then randomly crop to the size 190 pixels. Note that we perform dataset metrics computation on the original non-cropped (training) images. Only for ImageNet's entropy and frequency calculation we downsample the images to 128x128 for computational reasons.

For \textbf{CIFAR10}, we trained (5 times) LeNet5 (with added batch normalization after each layer), VGG16, ResNet50 and DenseNet121 on original labels using SGD with momentum 0.9 for 60 epochs with batch-size 128, batch-normalization $10^{-5}$ and weight-decay $5*10^{-4}$, cosine annealing scheduler \cite{Loshchilov2017} with initial learning rate 0.1 and minimal learning rate $5*10^{-4}$, which lowers the learning rate from the initial to the minimal one over the training epochs (without warm restarts). For the random label experiment, we have lowered the initial learning rate to 0.001 to ensure a quicker convergence. We use Kaiming normal weight initialization \cite{He2015} for all experiments.

For \textbf{KTH-TIPS2b}, we used the \emph{sample a} of each class for testing and the rest for training. We trained DenseNet121 for 60 epochs with batch-size 64, Adam with momentum 0.9, batch-norm $10^{-5}$, weight-decay $10^{-5}$ and a one cycle learning rate scheduler \cite{Smith2017,Smith2018} in which the learning rate first increases from a minimal one to a maximal one of $10^{-4}$ and then decreases over the rest of epochs to an even lower minimum. Standard Pytorch implementation parameters for OneCycleLR have been used to determine the initial and final learning rate.

For \textbf{Pascal}, we used train and validation splits of 2007 and 2012 for training and 2007 test split (in which we disregarded difficult label instances) for testing. We trained DenseNet121 and ResNet50 for 150 epochs with batch-size 128, SGD with momentum 0.9, batch-norm $10^{-5}$, weight-decay $5*10^{-4}$ and a step learning rate scheduler \cite{Loshchilov2017} which lowers the initial learning rate of 0.1 every 50 steps by a factor of 0.2. For \textbf{ImageNet}, we trained DenseNet121 for 100 epochs with batch-size 128, SGD with momentum 0.9, batch-norm $10^{-5}$, weight-decay $10^{-5}$ and a step learning rate scheduler which lowers the initial learning rate of 0.1 every 30 steps by a factor of 0.1. The training procedure for Pascal and ImageNet is inspired by Huang \etal \cite{Huang2017}. We used single NVIDIA A100 GPU to run Pascal/ImageNet style experiments with DenseNet or ResNet.

\section{Initial experiments: additional plots}
\label{ablation_additional}

\begin{figure}[!h]
     \centering
     \subfloat[\textbf{CIFAR10} \label{fig:cifar10_2bounds}]{%
        \includegraphics[width=0.32\textwidth]{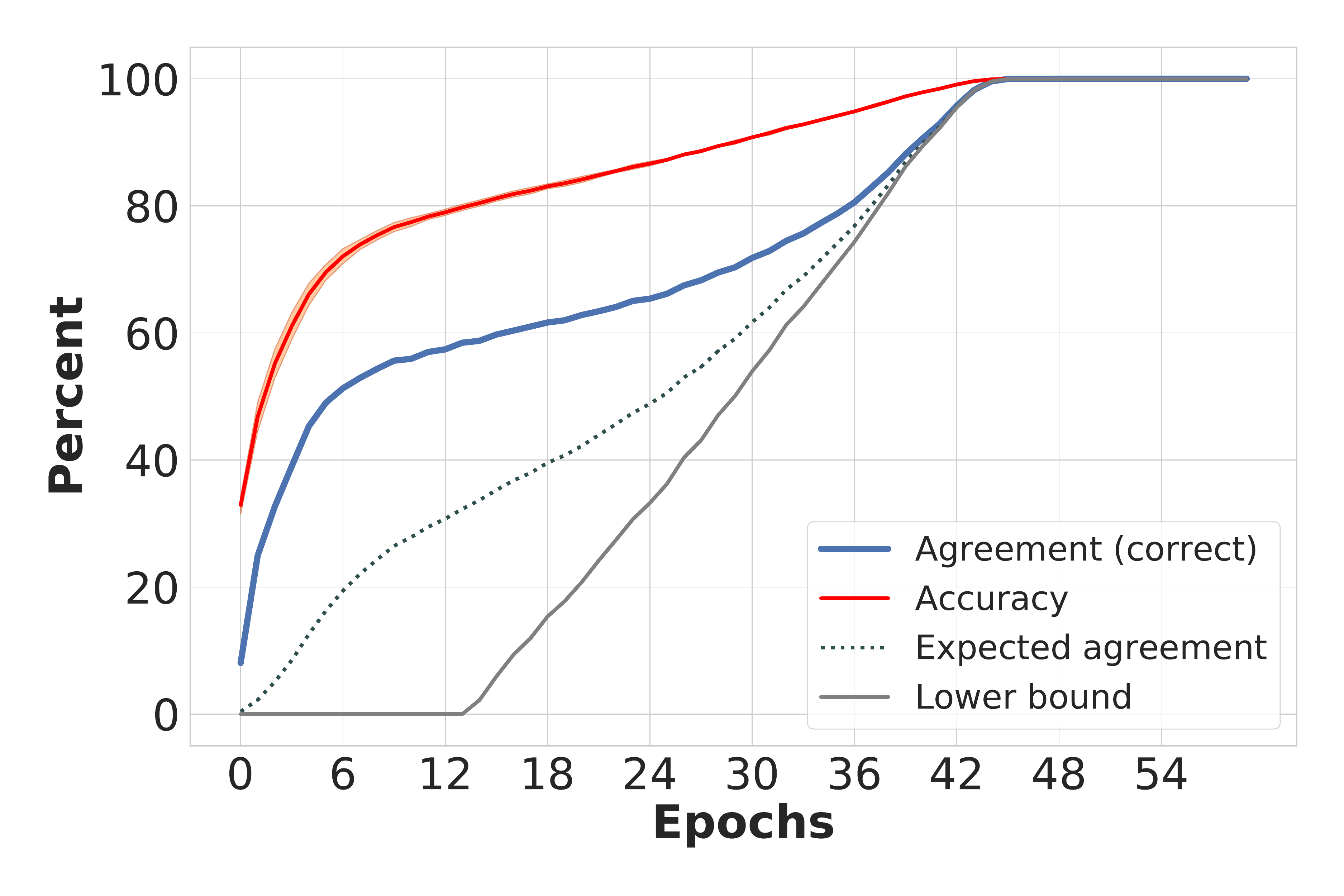}
    }
    \subfloat[\textbf{KTH-TIPS2b} \label{fig:kth_tips_2bounds}]{%
        \includegraphics[width=0.32\textwidth]{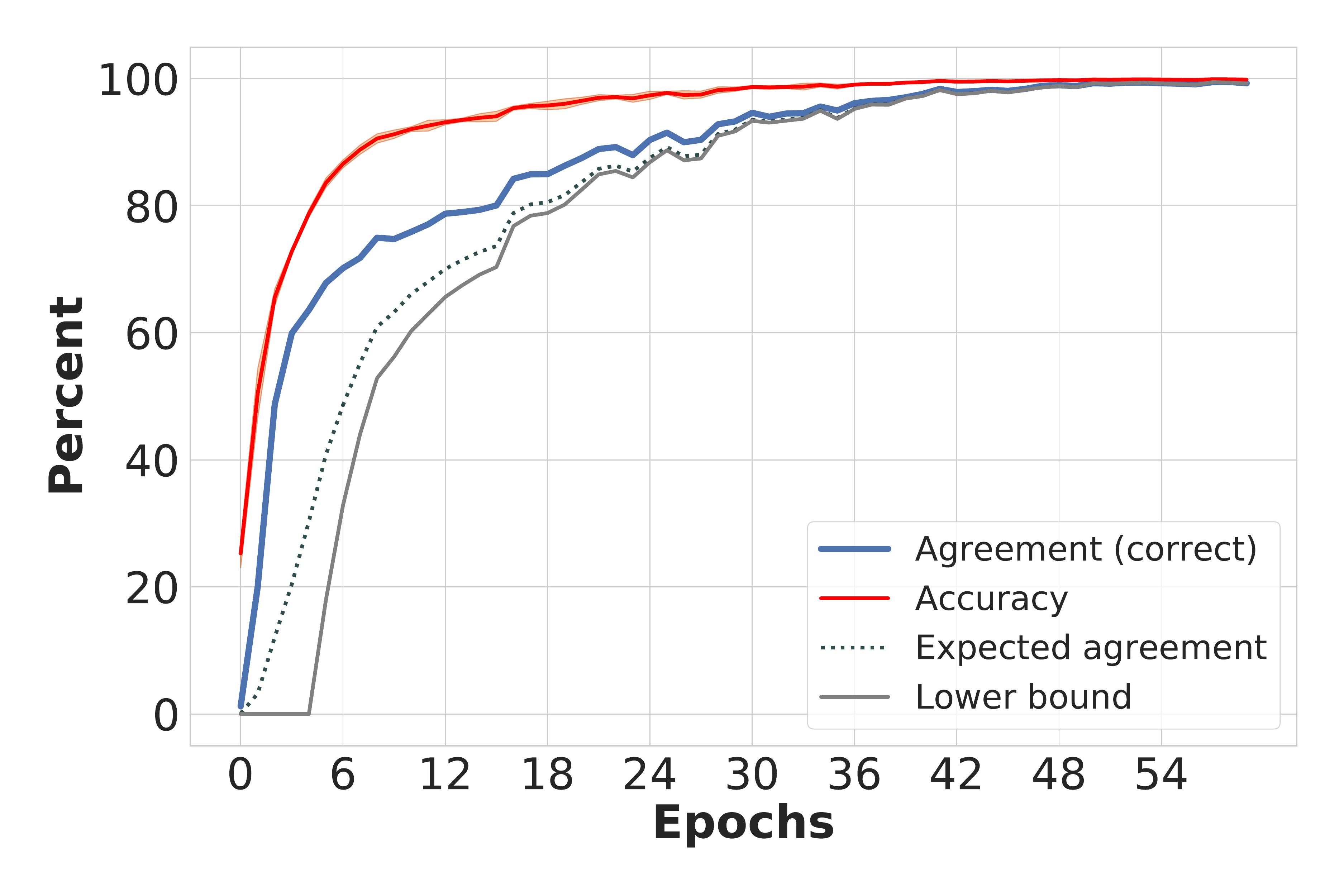}
    }
    \subfloat[\textbf{ImageNet} \label{fig:imagenet_2bounds}]{%
        \includegraphics[width=0.32\textwidth]{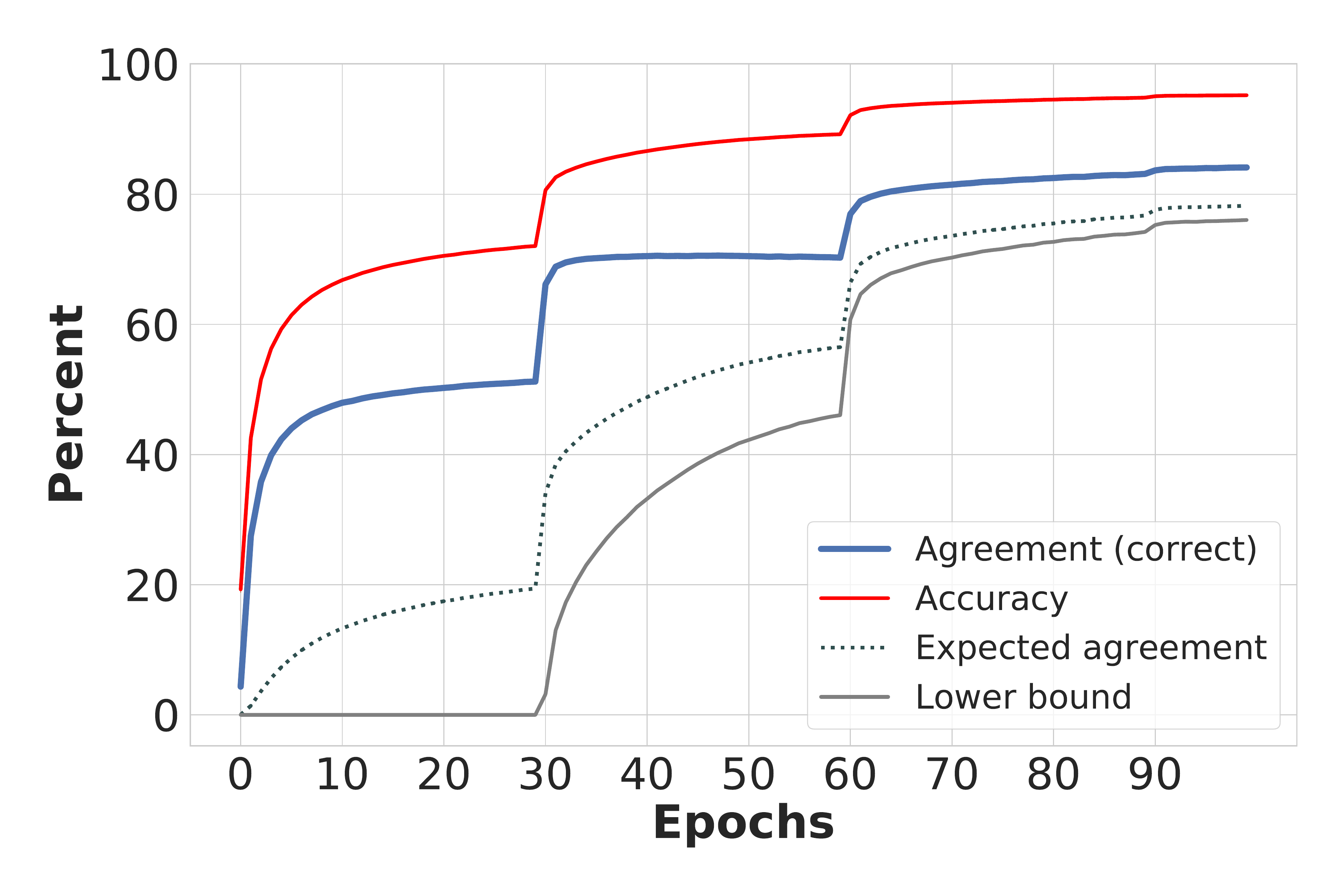}
    }
     \caption{Ablation study: Computing \emph{expected random agreement}, in comparison to agreement and lower bound. Expected random agreement is higher than the lower bound, but still lower than agreement. For Pascal we see in \cref{pascal_expected_random_agreement} that the deviation on agreement and the expected random agreement is rather small too.}
     \label{2bounds}
\end{figure}

First and foremost, in addition to the lower bound presented in the paper, we also computed \emph{expected random agreement} by multiplying the network accuracies (divided by 100 to the range between 0 and 1) in a given epoch. This gives us an assessment of how probable it is that networks randomly agree on dataset instances which they classify correctly, assuming that they classify dataset instances independently. We observe in \cref{2bounds} that expected random agreement is higher than our lower bound, but still lower than the actual agreement.

\begin{figure}[!h]
    \centering
    \includegraphics[width = 0.47 \textwidth]{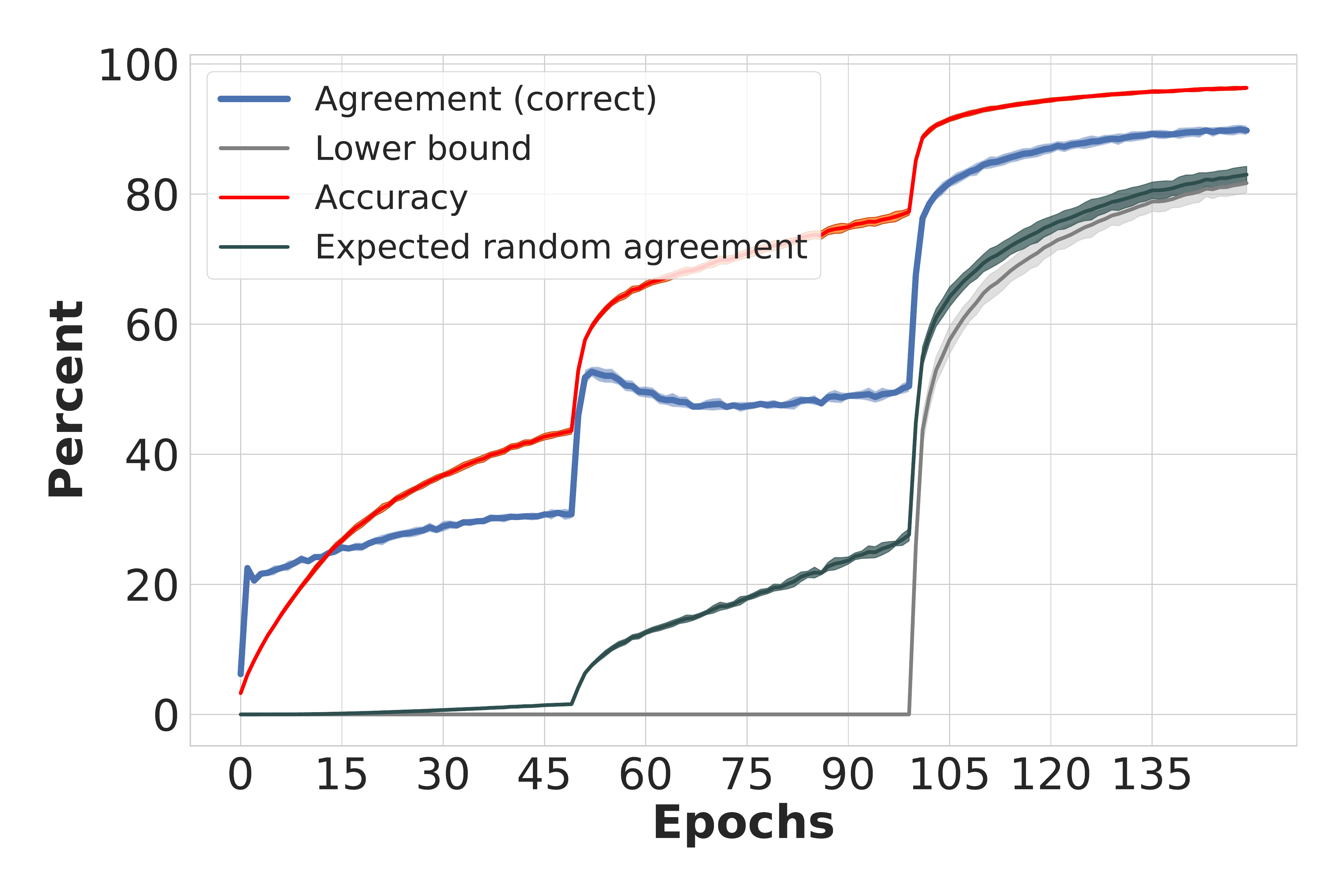}
    \caption{\textbf{Pascal DenseNet}: Agreement visualization on \emph{train set}, with expected random agreement, as well as standard deviation on agreement, expected random agreement and the lower bound.}
    \label{pascal_expected_random_agreement}
\end{figure}

As mentioned in the main body, we also conducted an experiment to calculate the standard deviation on agreement, similarly to the way we computed the deviation on accuracy. For Pascal, we ran the experimental setup 5 times, hence training 25 neural networks in total, to be able to calculate the deviation on agreement (and the lower bound). \cref{pascal_expected_random_agreement} visualizes that it is quite small.

\begin{figure*}[!h]
    \centering
    \subfloat[\textbf{Pascal DenseNet}\label{pabak_pascal}]{%
        \includegraphics[width = 0.49 \textwidth]{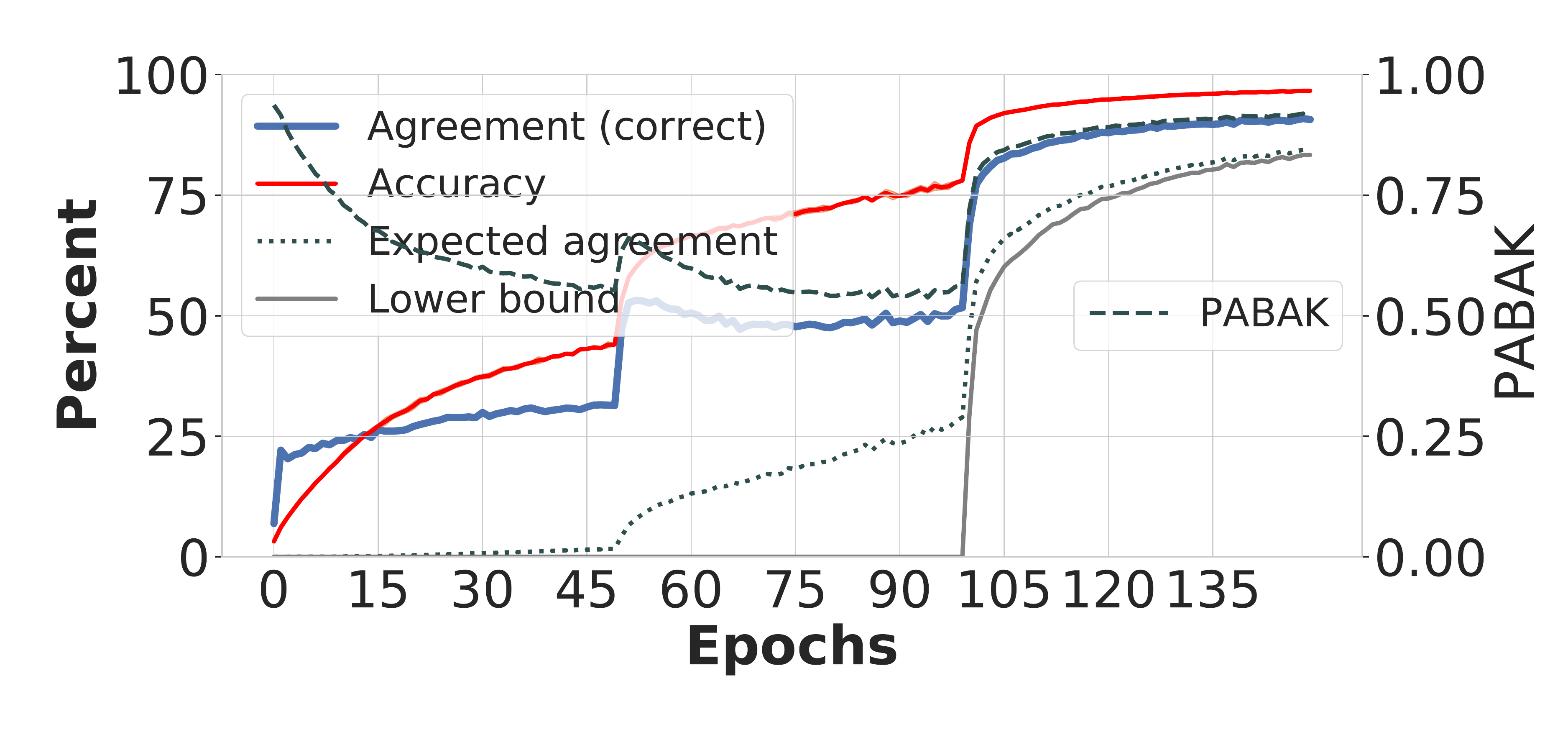}
    }
    \subfloat[\textbf{KTH-TIPS2b DenseNet} \label{pabak_kth_tips}]{%
        \includegraphics[width = 0.49 \textwidth]{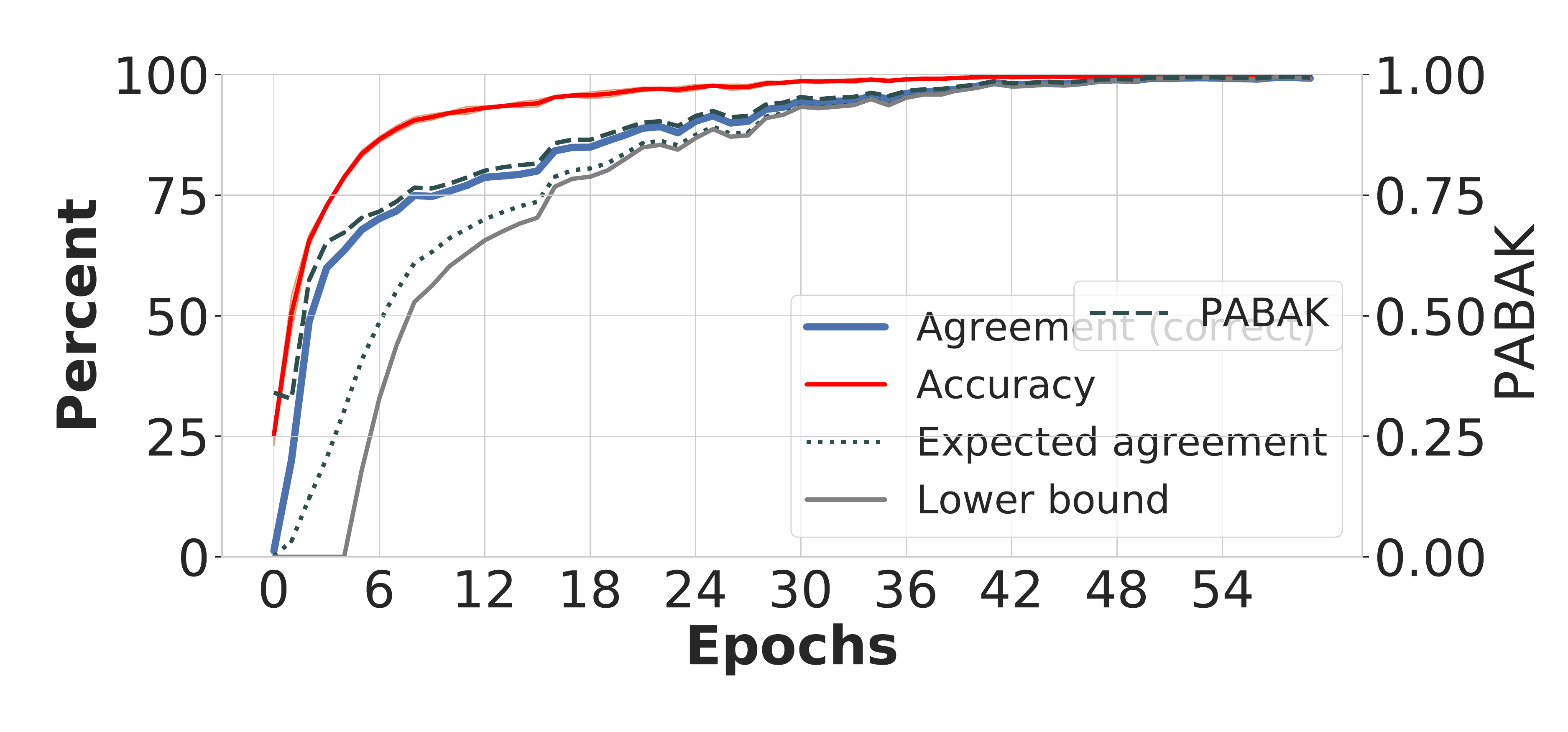}
    }
    \caption{Visualization of \textbf{PABAK} on \emph{train set} for a 2-class scenario (correctly classified vs. incorrectly classified). PABAK 's range is between -1 and 1. PABAK measure is 0 when observed agreement is 50\%.}
\end{figure*}

Second, let us strenghten the argument in favor of our definition of agreement even further. In the main body we have mentioned Cohen's kappa and PABAK as measures for the reliability of agreement. Usually, both operate on the notion of observed agreement, which considers not only true positives, but also true negatives. We focus only on true positives, because already taking into account true negatives makes the analysis more complex, since several trends are evaluated simultaneously. In addition, Cohen's kappa and PABAK operate over only 2 estimators. Since we have 5 networks, we have to either choose another measure, or to compute the average over all pairs of estimators. One measure for more than 2 estimators is \emph{Fleiss kappa}, but it assumes that instead of a fixed number of estimators, estimators are sampled from a larger pool such that it is not the case that every dataset instance is classified by the same estimators. 

\begin{figure}[!h]
     \centering
     \subfloat[Batch-size 16 \label{fig:cifar10_batch16}]{%
        \includegraphics[width=0.32\textwidth]{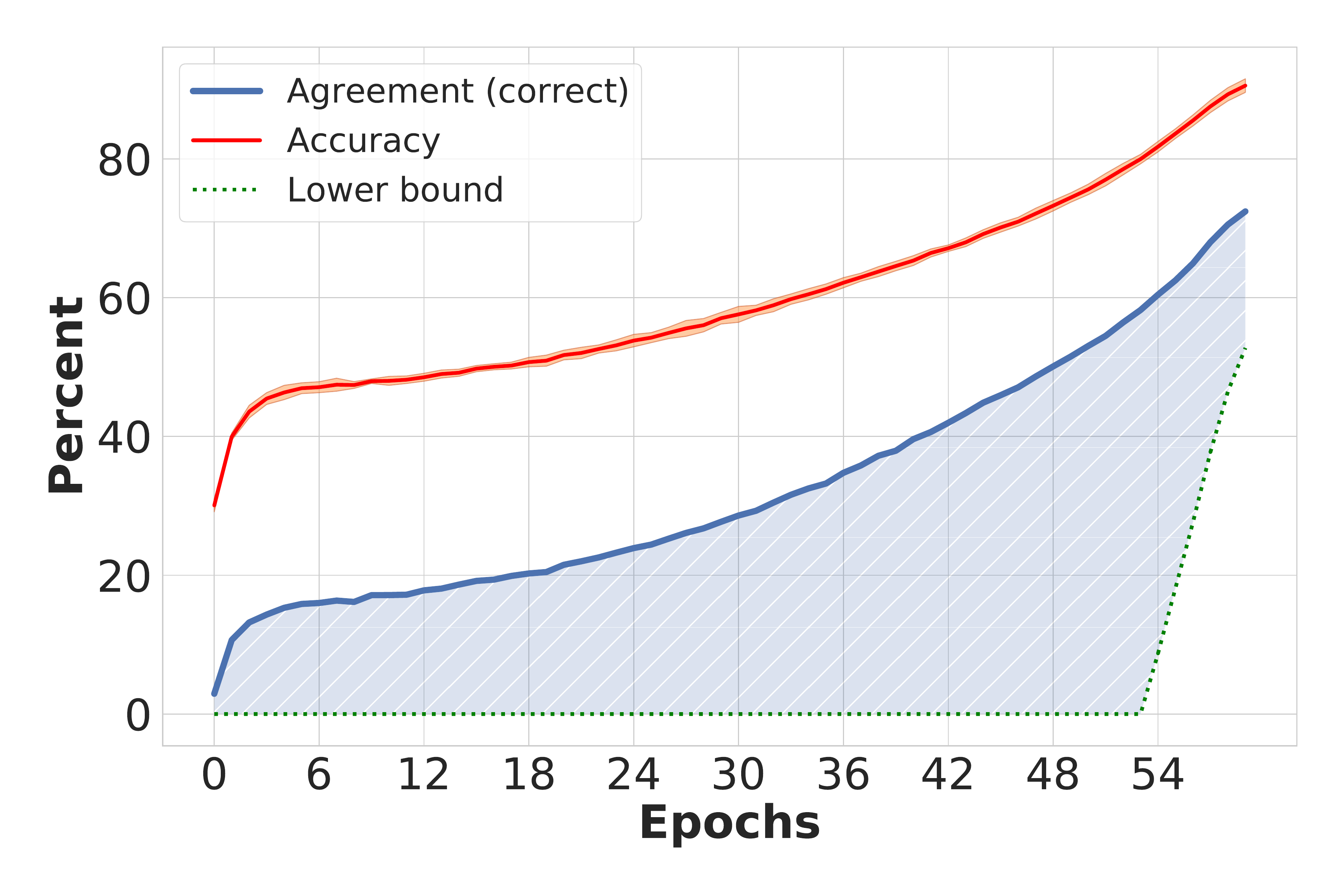}
    }
    \subfloat[Batch-size 64 \label{fig:cifar10_batch64}]{%
        \includegraphics[width=0.32\textwidth]{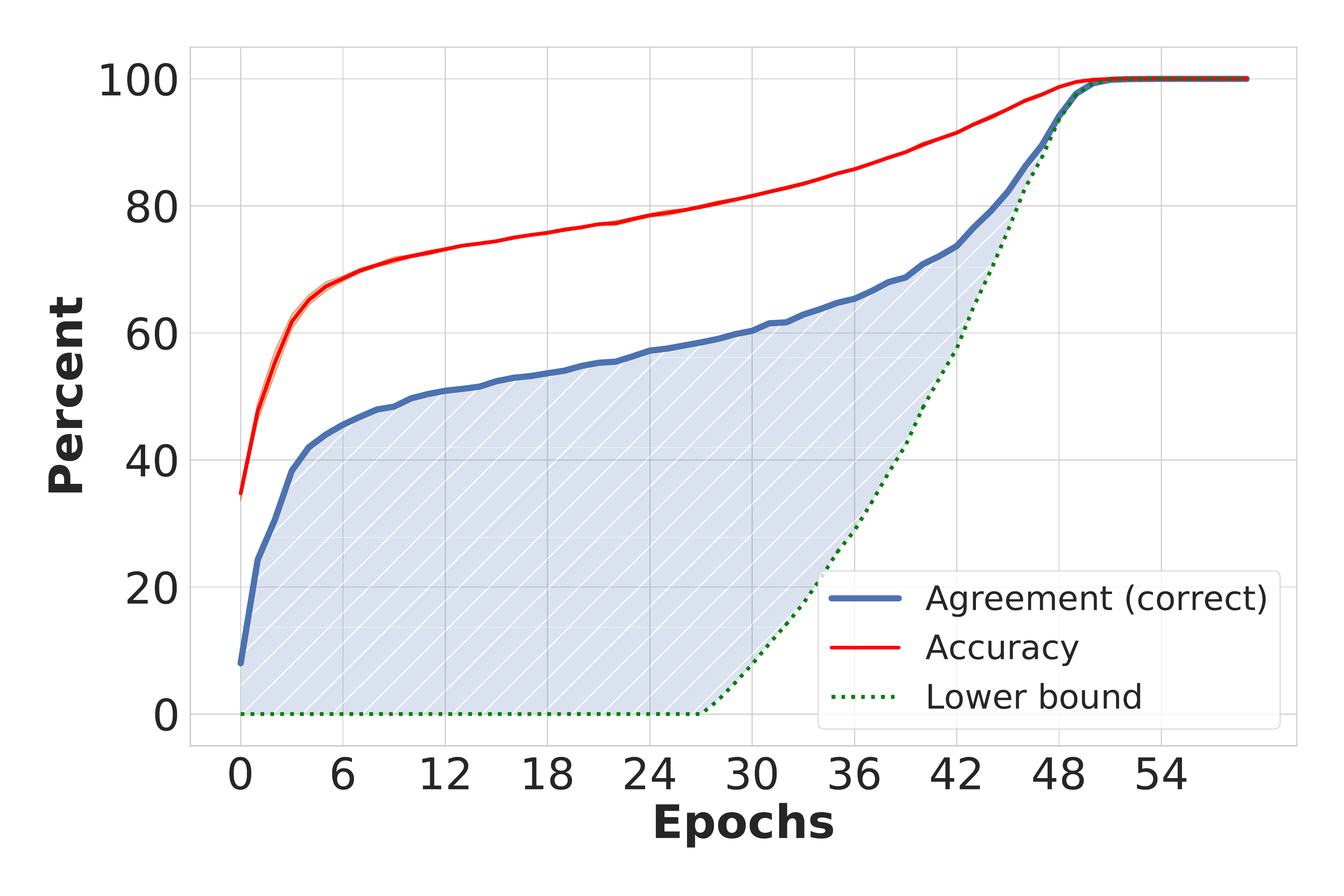}
    }
    \subfloat[Batch-size 256 \label{fig:cifar10_batch256}]{%
        \includegraphics[width=0.32\textwidth]{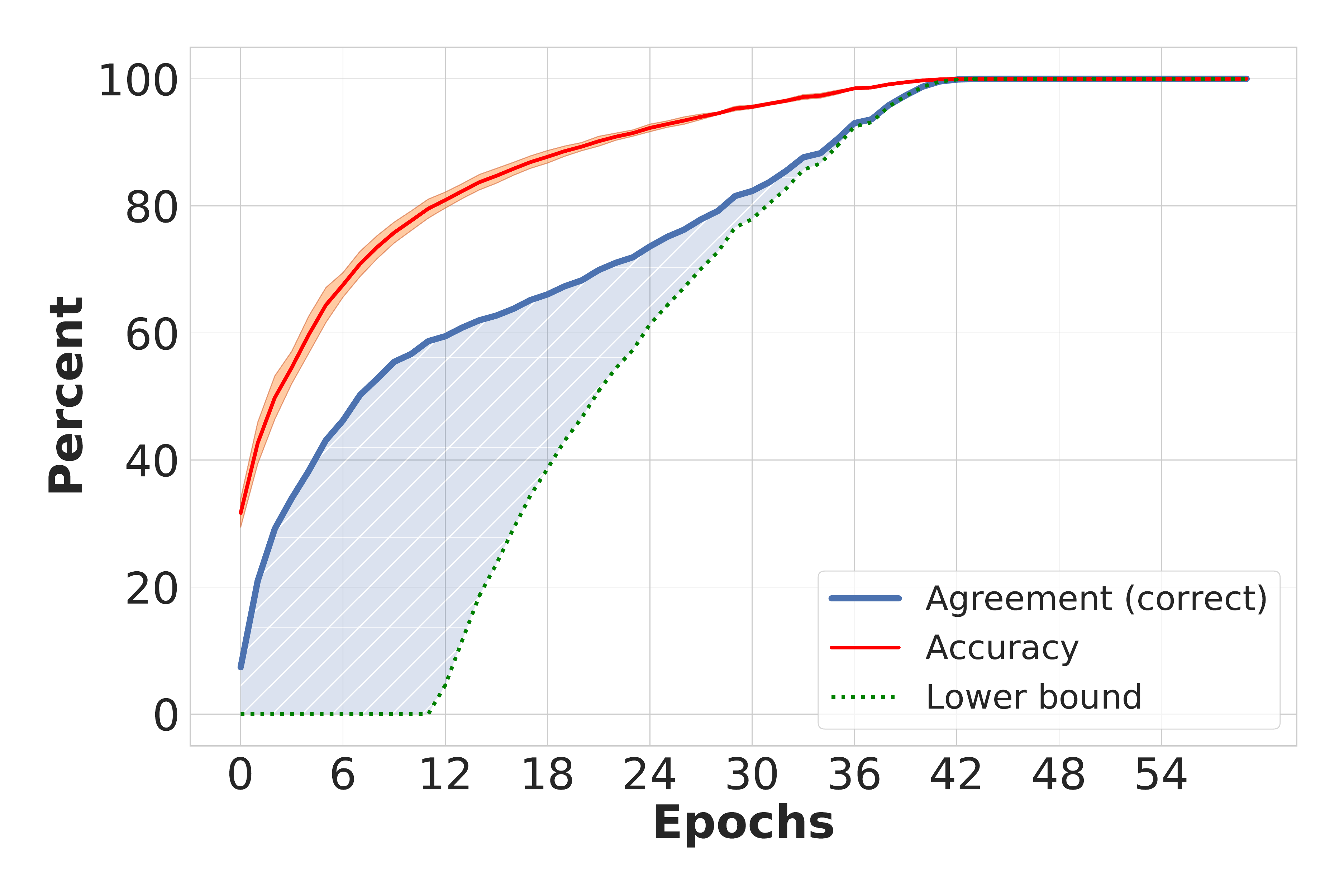}
    }
    \caption{Ablation study on \textbf{CIFAR10}: training with different batch-sizes}
    \label{batch_exp}
\end{figure}
\begin{figure}[!ht]
     \centering
     \subfloat[LeNet5, batch-size 128 \label{fig:cifar10_lenet5}]{%
        \includegraphics[width=0.32\textwidth]{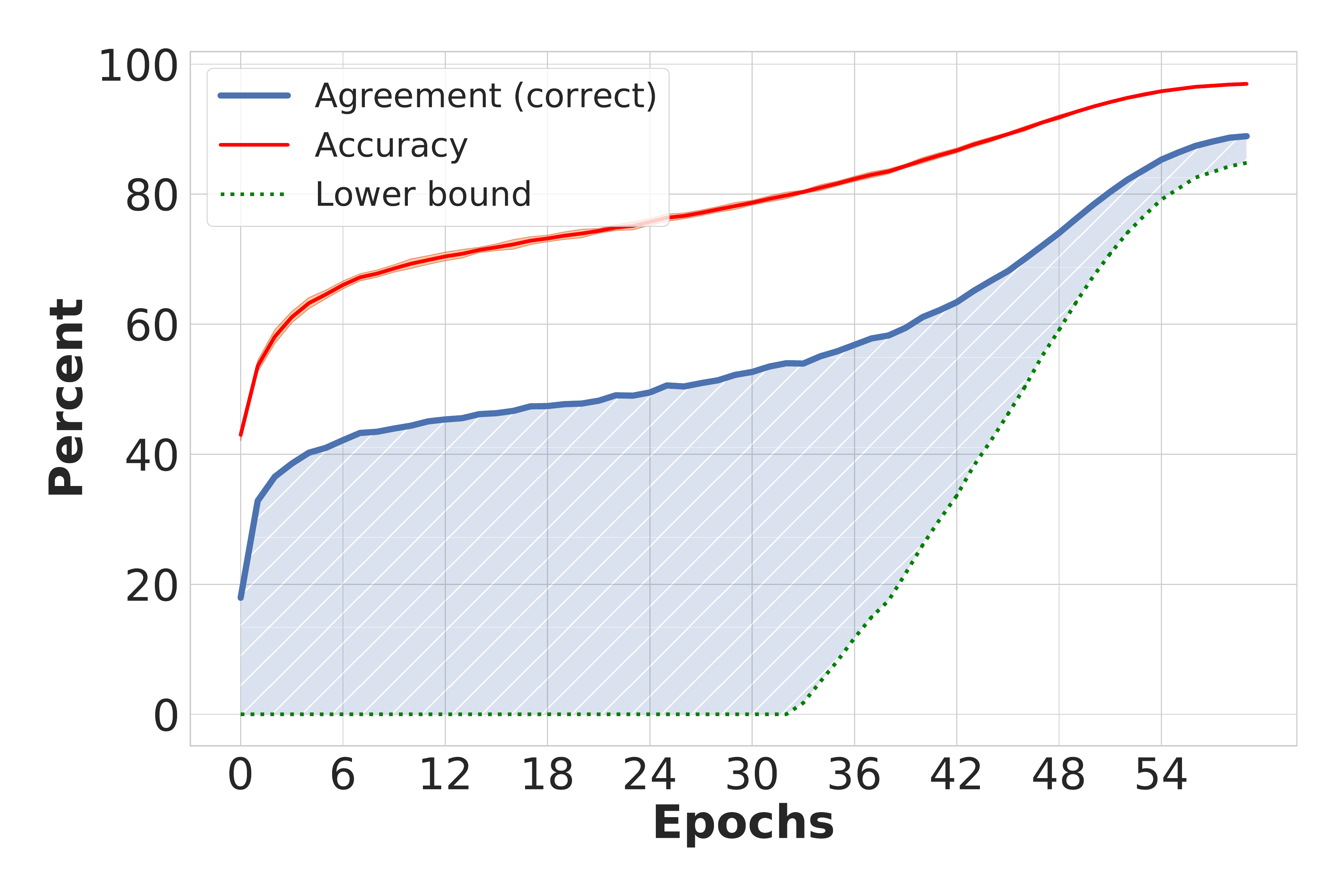}
    }
    \subfloat[VGG16, batch-size 128 \label{fig:cifar10_vgg}]{%
        \includegraphics[width=0.32\textwidth]{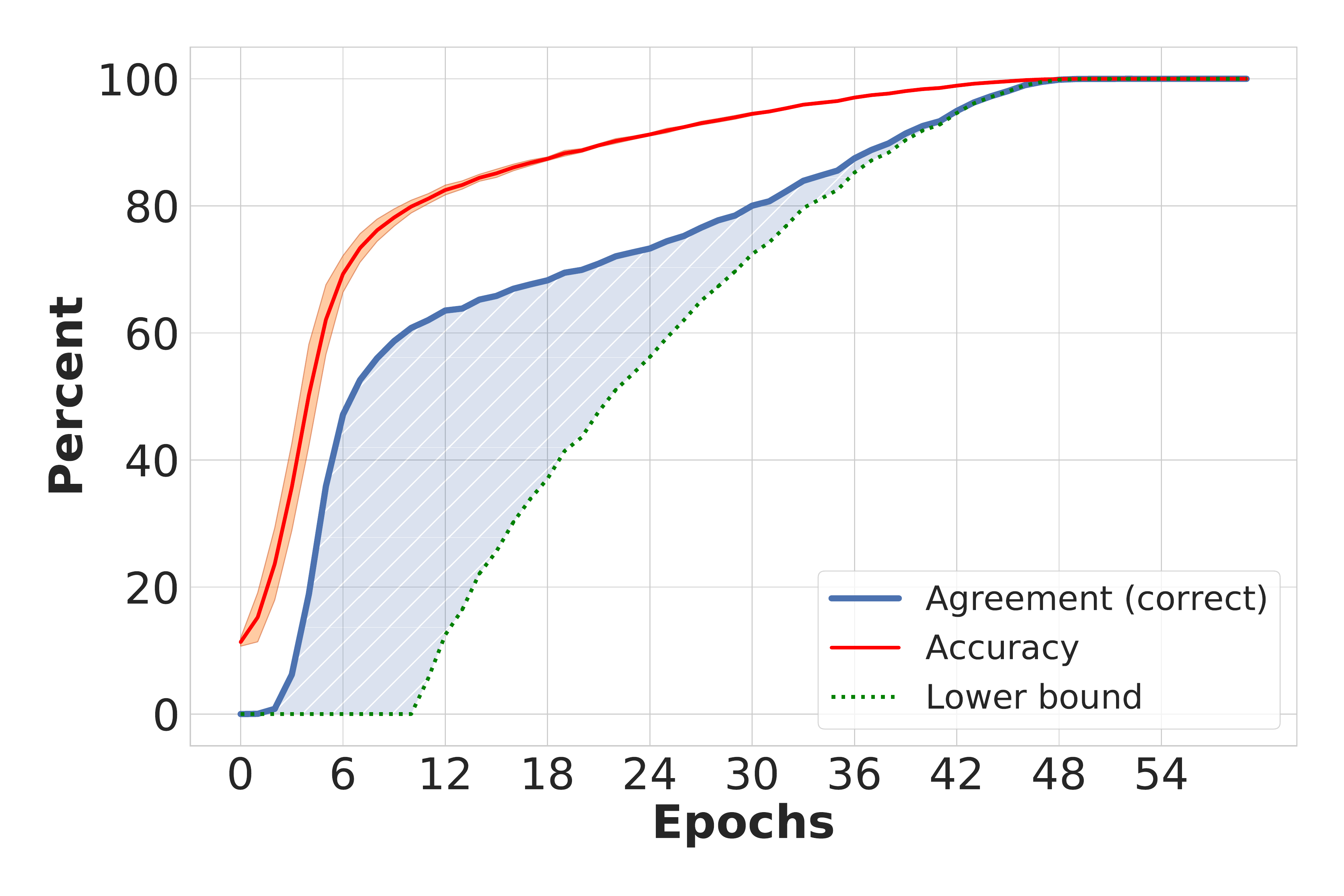}
    }
    \subfloat[ResNet50, batch-size 128 \label{fig:cifar10_resnet}]{%
        \includegraphics[width=0.32\textwidth]{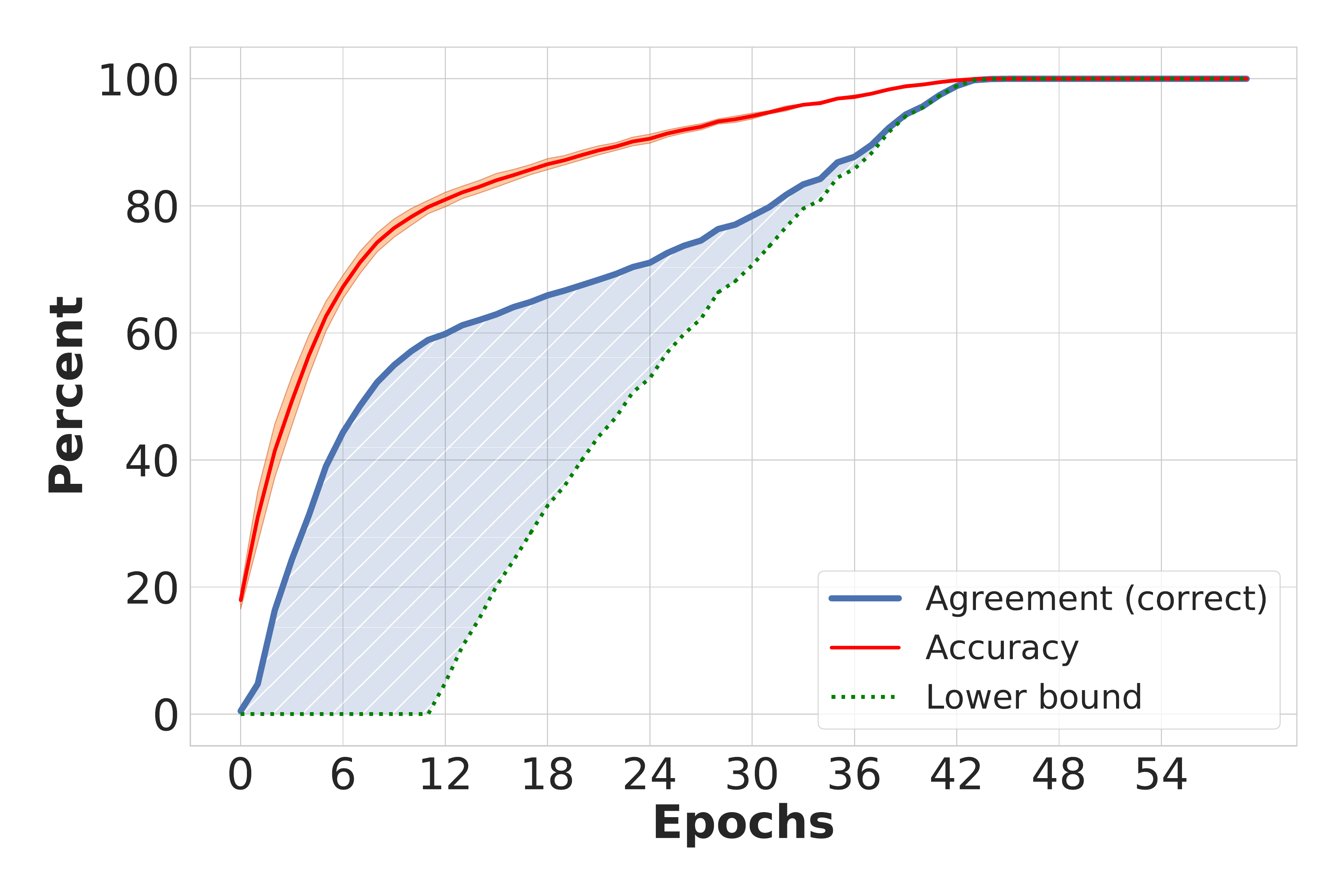}
    }
    \caption{Ablation study on \textbf{CIFAR10}: training with differnt architectures}
    \label{arch_results}
\end{figure}
To get a feel for observed agreement, let us consider a simplified scenario in which there are 2 classes - correctly classified and wrongly classified. We can then sum instances both estimators classify correctly and incorrectly, normalize by the total number of instances. We then linearly transform it to counteract the prevalence bias as described in \cite{Byrt1993} and average over pairs of estimators. We see in \cref{pabak_pascal} that if the accuracy grows slowly, we get a U-shape. First, PABAK is high due to the number of true negatives - it is the case when both estimators classify wrongly,- then it gets higher due to the number of instances pairs of estimators classify correctly. In \cref{pabak_kth_tips} we can see that if accuracy grows fast, PABAK curve resembles the true positive agreement we defined. However, the exact values of agreement we defined and PABAK cannot be compared as easily, because PABAK ranges between -1 and 1 and is 0 when observed agreement (which incorporates true positives and negatives) is 50\%. Note that the 2 class scenario is a crude simplification, as we would actually want to know in a multi-scenario, whether estimators missclassify \emph{in the same way} (into the same wrong class).

Third, we further conducted agreement experiments for CIFAR10 on DenseNet for several batch-sizes (5 networks for every batch-size, in analogy to the main body experiments), see \cref{batch_exp}. We also trained 5 networks each for CIFAR10 on LeNet5, VGG16 and ResNet50, in addition to DenseNet, see \cref{arch_results}. Comparison of both figures shows that agreement is present for different batch-sizes and architectures, as well as that the agreement curve changes similarly for growing batch-sizes and architecture complexity.

\begin{figure}[!h]
    \centering
    \includegraphics[width = 0.44 \textwidth]{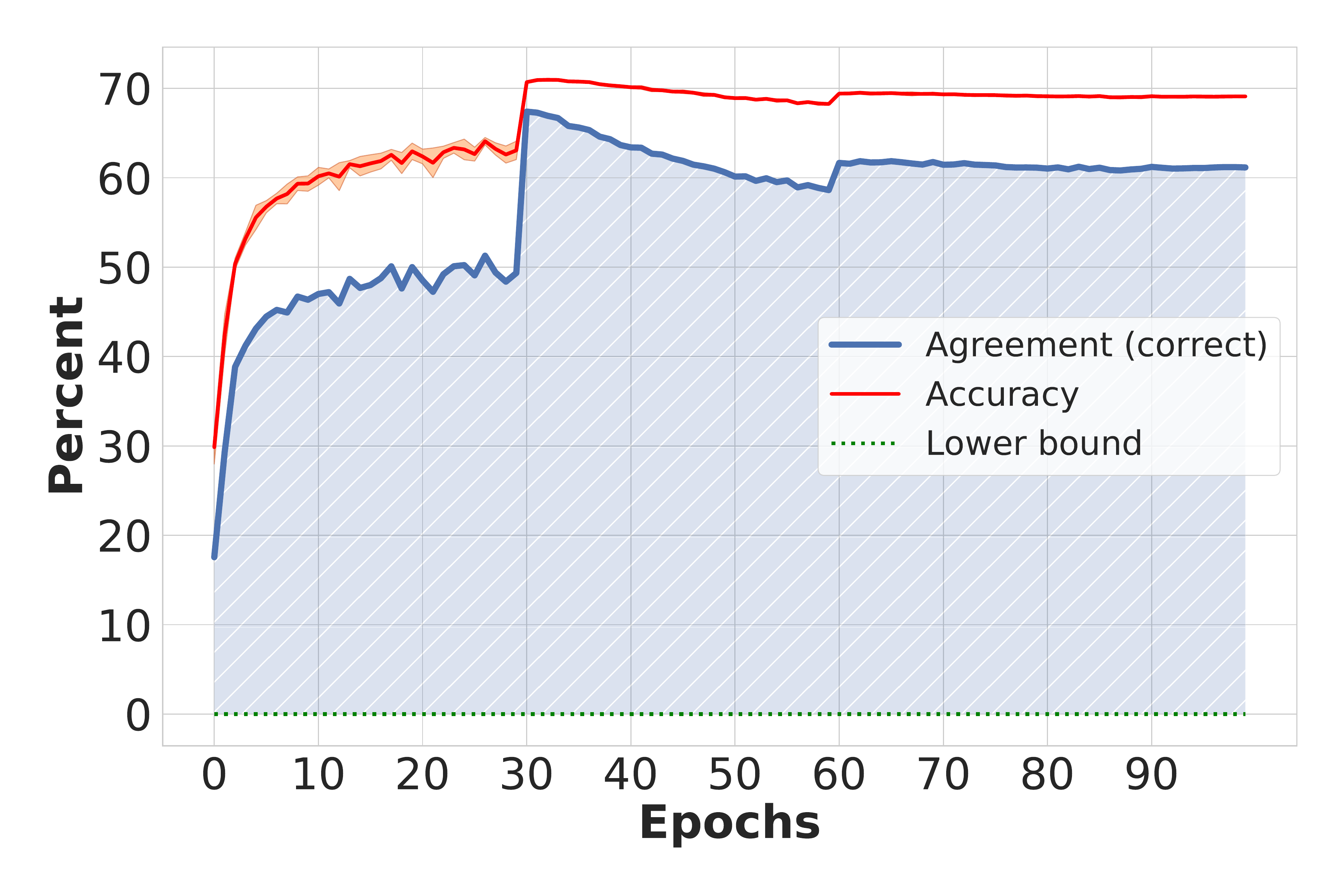}
    \caption{\textbf{ImageNet DenseNet}: Agreement visualization on \emph{test set}}
    \label{imagenet_densenet_test}
\end{figure}
\begin{figure*}[!th]
    \centering
    \subfloat[\textbf{CIFAR10} Uncertainty \label{cifar_pred_uncertainty}]{%
        \includegraphics[width = 0.32 \textwidth]{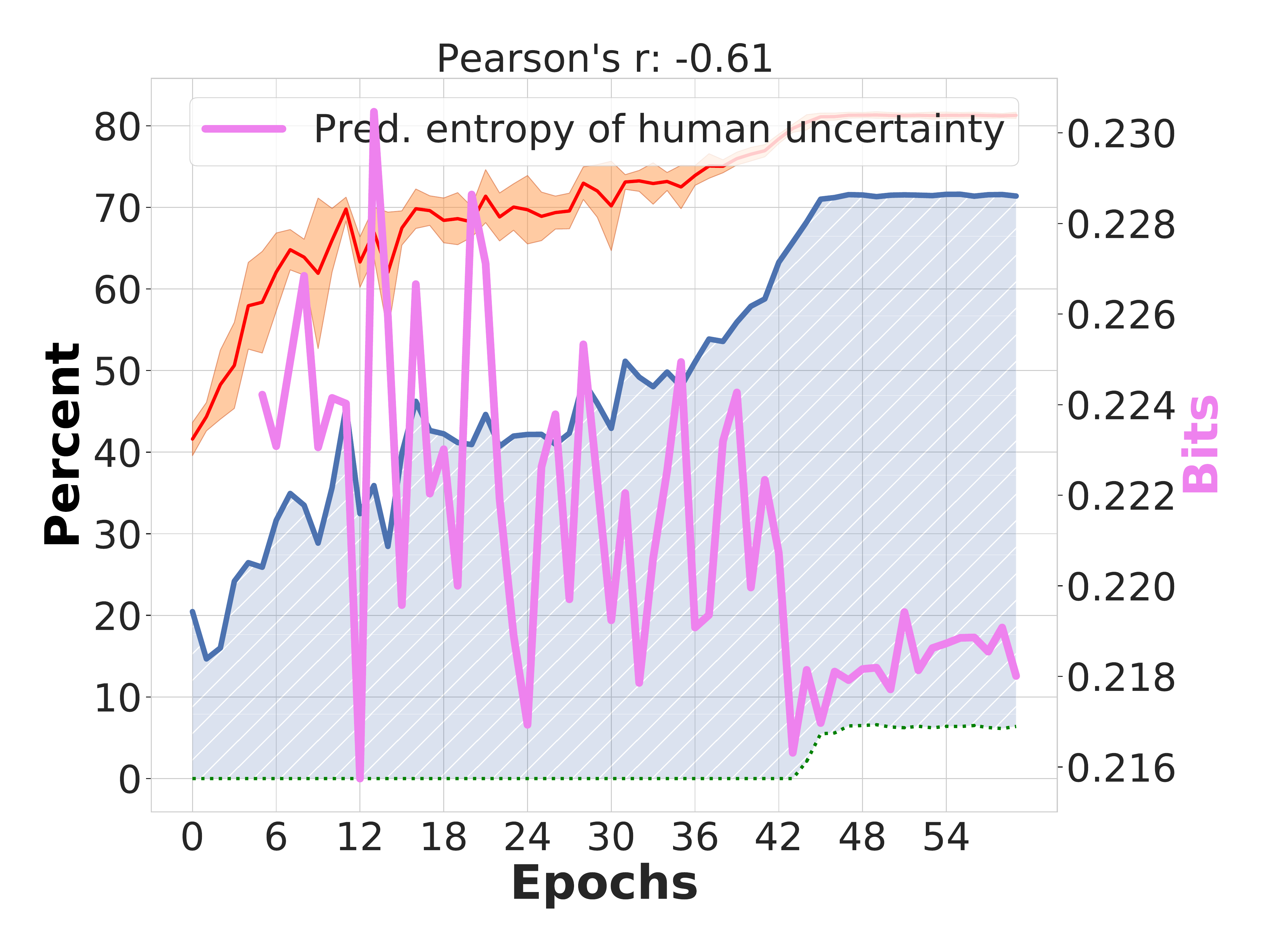}
    }
    \subfloat[\textbf{Pascal} Image Entropy ]{%
        \includegraphics[width = 0.32 \textwidth]{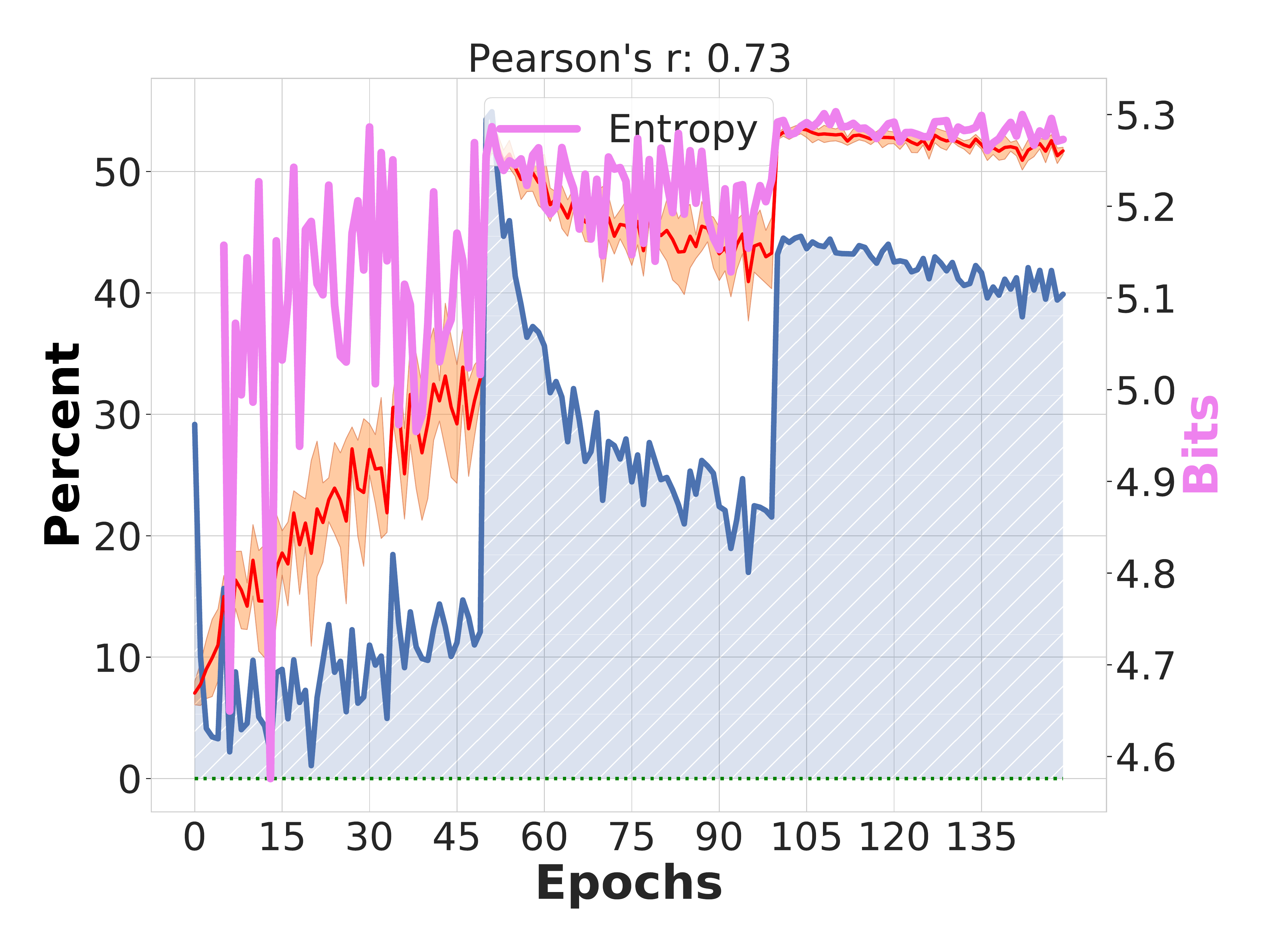}
    }
    \subfloat[\textbf{KTH-TIPS2b} Illumination]{%
        \includegraphics[width = 0.32 \textwidth]{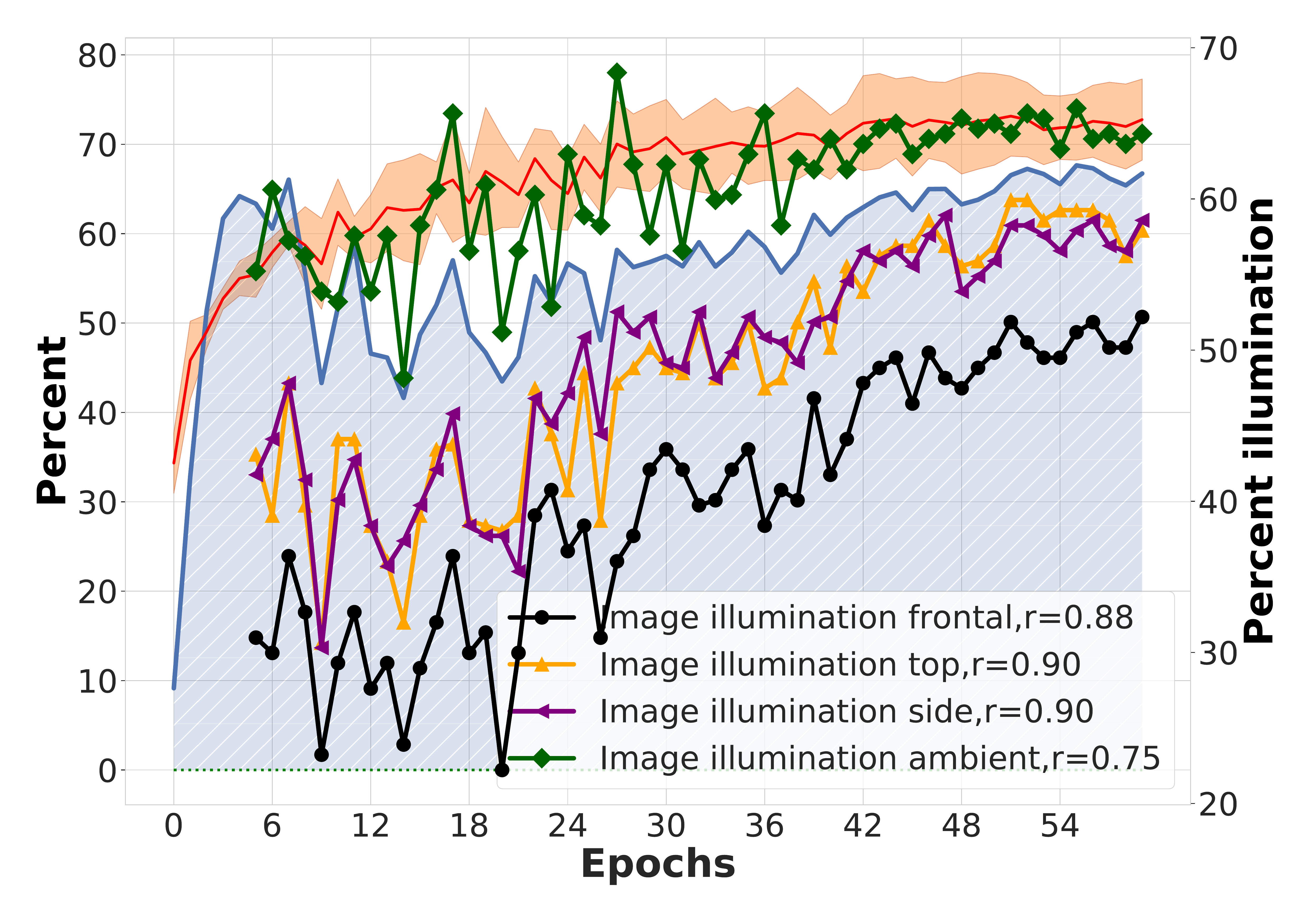}
    }
	\caption{Agreement and selected dataset metrics on the \emph{test sets} of CIFAR10, Pascal, and KTH-TIPS2b, based on DenseNet. Metric values are shown in \textbf{purple} (right y-axis), in correlation to accuracy (\textbf{red}), agreement (\textbf{blue} curve) and its difference to lower-bound  (shaded \textbf{blue} area) (left y-axis).}
	\label{fig:kth_tips_cifar_selected_test_metrics}
\end{figure*}

In \cref{fig:kth_tips_cifar_selected_test_metrics}, we exemplary visualize the test agreement for the three datasets CIFAR10, Pascal and KTH-TIPS2b and in \cref{imagenet_densenet_test} for ImageNet. We observe that for all four datasets there is sufficiently high agreement on the test set. Not surprisingly, the standard deviation of the accuracy is higher than for all train sets. In analogy to Pascal train set results, we see jumps in accuracy and agreement where the learning rate has been lowered in steps. Tentatively, for CIFAR10, Pascal and KTH-TIPS we visualize some dataset metric correlations on the test set too. For CIFAR10, we visualize the entropy of the soft labels as a metric. It has been computed by Peterson \etal \cite{Peterson2019} only for the test set. We see a slight downward tendency such that the entropy of soft labels decreases over the course of training. For Pascal, we see that similarly to the train entropy in fig. 4 of the main body, there is a correlation present for the test entropy. Even more interesting is the correlation of illumination on the KTH-TIPS2b dataset. Further, in fig. 5e of the main body we have seen that frontal illumination is learned slower than other kinds of illumination on the train set, in \cref{fig:kth_tips_cifar_selected_test_metrics} we see that for the test set this tendency is even more nuanced such that agreement is highest on the ambient illumination type and lowest on the frontal illumination type. A thourough analysis though, when dataset metric correlations are present/absent on the test data and how well they correlate with those on the train data is left for future work.

\section{Additional evaluation of correlations for CIFAR10 and Pascal}
As mentioned in the main section, the range of fluctuations of CIFAR10 dataset metrics is negligible and therefore it is hard to judge the correlations between agreement and dataset metrics. To elaborate, the entropy is almost the same around 6.5, while sum of edge strenghts, segment count and percentage of DCT coefficients decrease slightly. The CIFAR10 distributions of dataset metrics indicate that for entropy, uncertainty and segment count, the distribution of values centers on a couple of values and, hence, there is no diversity, which can be reflected in agreement correlations. Further, since the dataset contains highly downsampled images, neither the presence of high frequencies, nor meaningful edge strengths are expected. Hence, the direction of correlations is the same as for the texture dataset KTH-TIPS2b and opposite of Pascal, which also contains objects as CIFAR10 does. 

\begin{figure}[!t]
    \centering
    \begin{minipage}{0.47\textwidth}
        \includegraphics[width = 0.99 \textwidth]{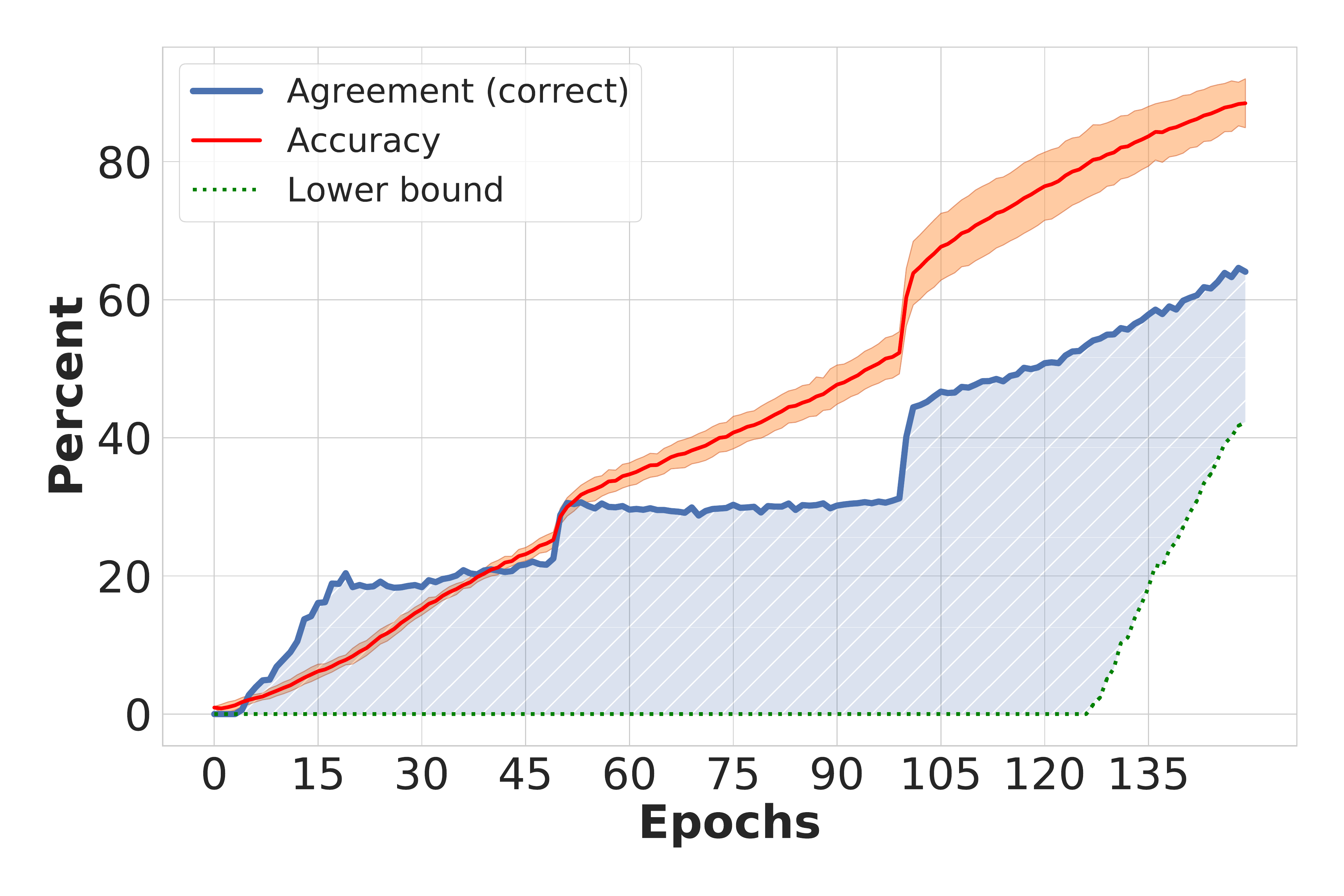}
	\caption{\textbf{Pascal ResNet}: Agreement visualization on \emph{train set}}
	\label{pascal_resnet_agreement}
    \end{minipage}
    \begin{minipage}{0.47\textwidth}
	\includegraphics[width = 0.8 \textwidth]{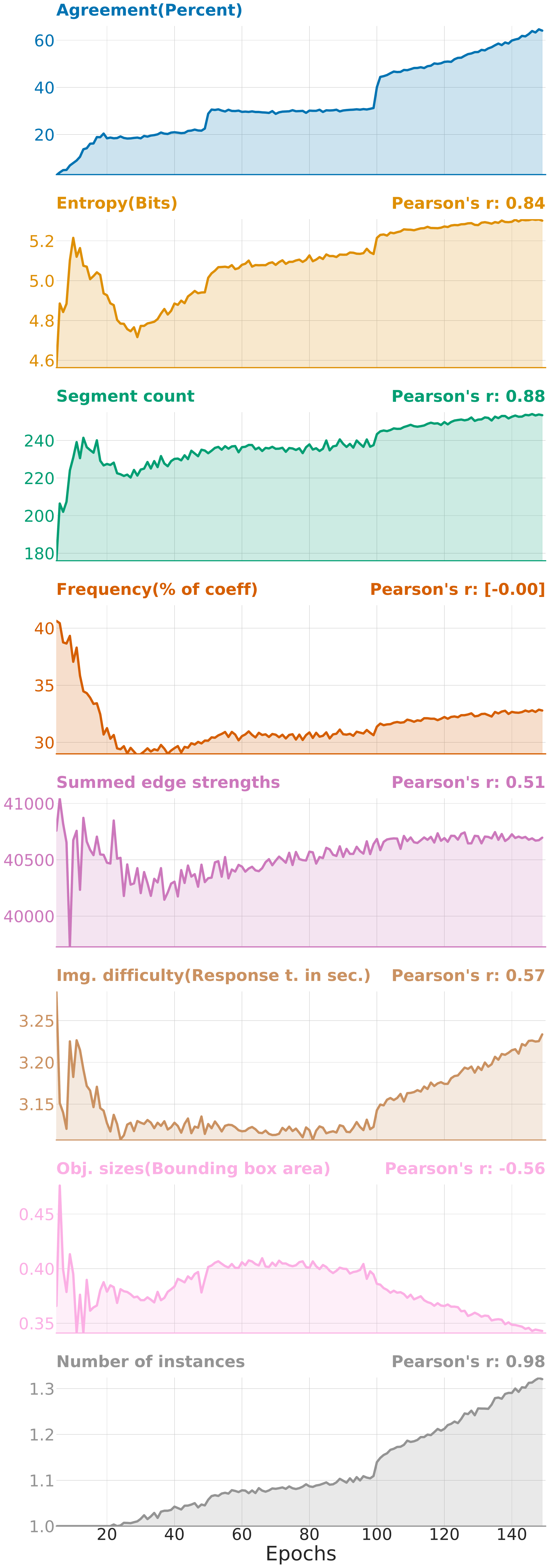}
	\caption{\textbf{Pascal ResNet}: Dataset metrics correlations on \emph{train set}}
	\label{pascal_resnet_metrics}
	\end{minipage}
	\caption{\textbf{Pascal ResNet}: Agreement and dataset metrics correlations}
\end{figure}

To further support our results we, in addition to DenseNet, trained 5 ResNet50 networks on Pascal and computed with the same experimental setup the agreement (see \cref{pascal_resnet_agreement}), as well as the dataset metrics correlations (see \cref{pascal_resnet_metrics}). We see that ResNet learns more slowly than DenseNet (with the experimental setup chosen for DenseNet), but the general metrics tendency, when agreement approximately reaches 20\%, remains the same as for DenseNet in fig.~4 of the main body. The strength of the correlation, measured by the Pearson correlation coefficient, is not as high for ResNet, as for DenseNet. For the frequency dataset metric it is even absent. The value is in brackets, because the corresponding \emph{2-tailed p-value} is bigger than 0.001. For Pearson correlation coefficient between agreement and the given dataset metric, the null hypothesis is that both are uncorrelated. The higher the p-value, the more the null hypothesis is supported, The lower p-value supports the presence of a correlation.

It would be interesting to further study both the initial learning phase when agreement is low, as well as the subsequent learning phase which this paper primarely was focused on.

\newpage
\section{Dataset metrics}
\label{metrics}
In this section we first give more details on how exactly we computed the dataset metrics and then visualize them on the example of ImageNet, in addition to the Pascal examples presented in the main paper, as well as visualize the matrices of DCT coefficients for those examples.

Let us start with the computation of the dataset metrics, evaluated in the main paper:
\begin{itemize}
\item \textbf{Segment count}: Felzenszwalb and Huttenlocher \cite{Felzenszwalb2004} introduce a graph-based image segmentation algorithm into regions, which can be summed up to get a segment count - a numer of segments in the image. First, images are smoothed with a Gaussian kernel of $\sigma$ standard deviation, then image regions are compared for similarity at a certain scale \emph{k} and merged if similar, subsequently small regions of size \emph{min} are filtered out. Hence, the most important parameter is the scale \emph{k}, larger value means preference for larger components. We used default parameters for the segmentation.

\item \textbf{Sum of edge strengths}: Isola \etal \cite{Isola2014b} compute semantically meaningful boundaries (between objects) in an image based on statistical pixel dependencies (pointwise mutual information). The resulting edge strengths (edge contours) can be summed to get one value characterizing the amount of edges in the image.

\item \textbf{Mean image intensity entropy}: Image intensity entropy for grayscale images is computed by sliding a window of a certain size \emph{k} (in our case 10) and then averaging the local entropies. Similar to the case of segment count, the window-size reflects how much noise to ignore in the image.

\item \textbf{Percentage of important DCT coefficients}: DCT coefficient matrix quantifies the spatical frequency in vertical and horizontal directions. Usually, lower frequency coefficients exhibit greater values. On the basis of this matrix we compute the percentage of DCT coefficients which contain 99.98\% of the energy in the image, computed by comparing the norm of the first \emph{c} sorted absolute values of DCT coefficients against the norm of all coefficients. In other words, this metric calculates how many coefficients are needed to reconstruct the image to a sufficiently high degree. 
\end{itemize}

\begin{figure*}[!ht]
    \centering
    \subfloat{%
        \includegraphics[width=.22\linewidth]{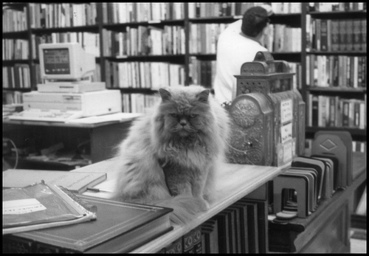}
    }
    \subfloat{%
        \includegraphics[width=.22\linewidth]{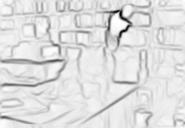} 
    }
    \subfloat{%
        \includegraphics[width=.22\linewidth]{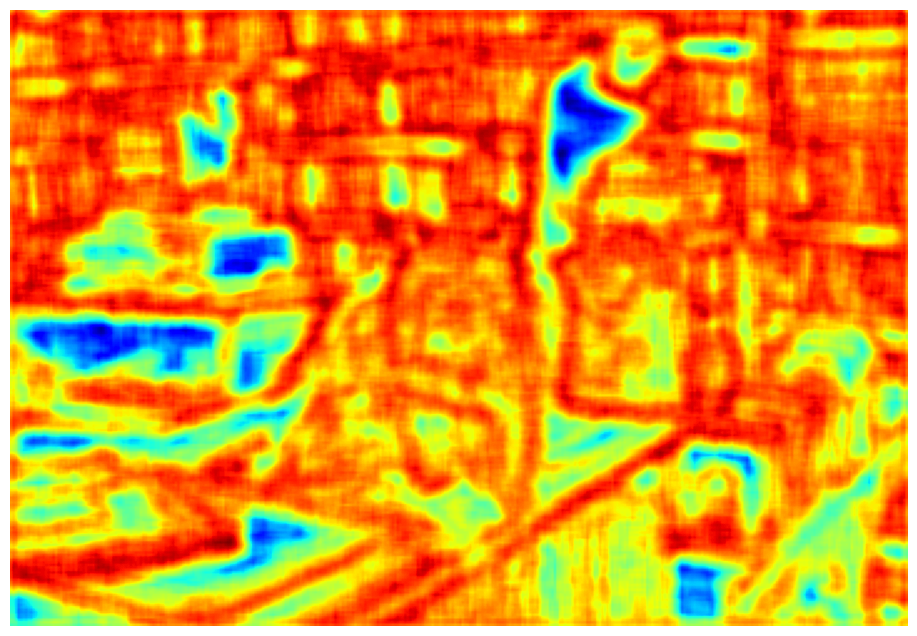}
    }
    \subfloat{%
        \includegraphics[width=.22\linewidth]{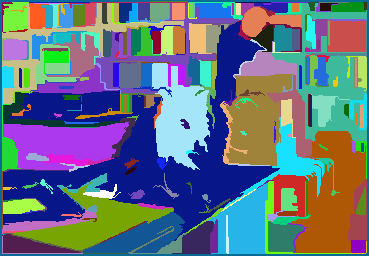}
    }\\
    \centering
    \setcounter{subfigure}{0}
     \subfloat[Image]{%
        \includegraphics[width=.22\linewidth]{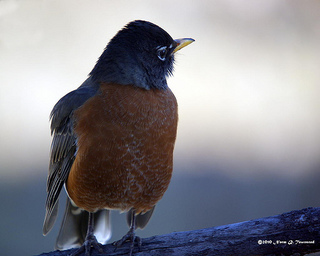}
    }
    \subfloat[Edge strengths]{%
        \includegraphics[width=.22\linewidth]{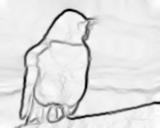} 
    }
    \subfloat[Entropy]{%
        \includegraphics[width=.22\linewidth]{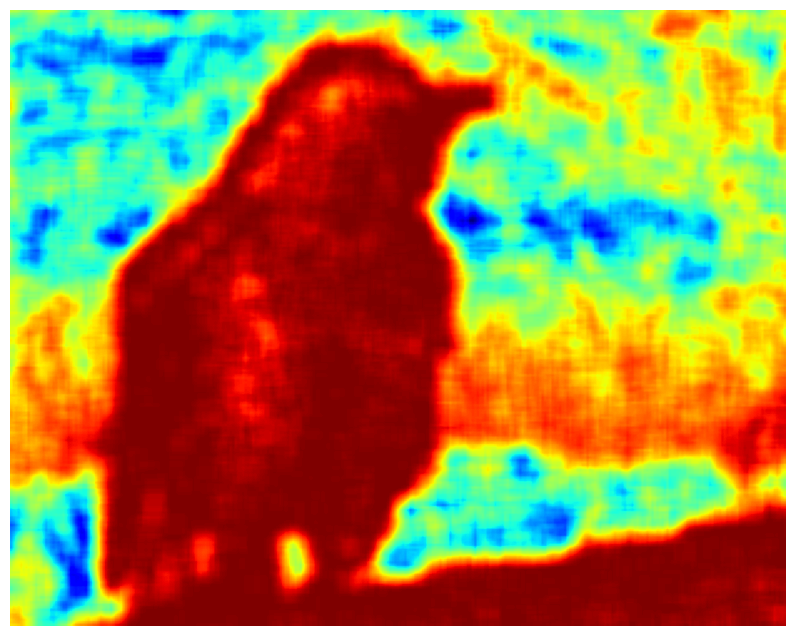}
    }
    \subfloat[Segments]{%
        \includegraphics[width=.22\linewidth]{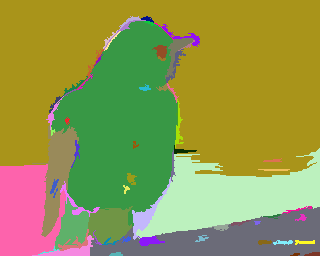}
    }
	\caption{Visualization of metrics on ImageNet}
	\label{fig:metrics_imagenet}
\end{figure*}
\begin{figure*}[!ht]
    \centering
    \begin{minipage}{0.24\textwidth}
        \includegraphics[height=.65\linewidth]{plots/metrics_examples/ILSVRC2012_val_00005712.JPEG}
        \includegraphics[height=.65\linewidth]{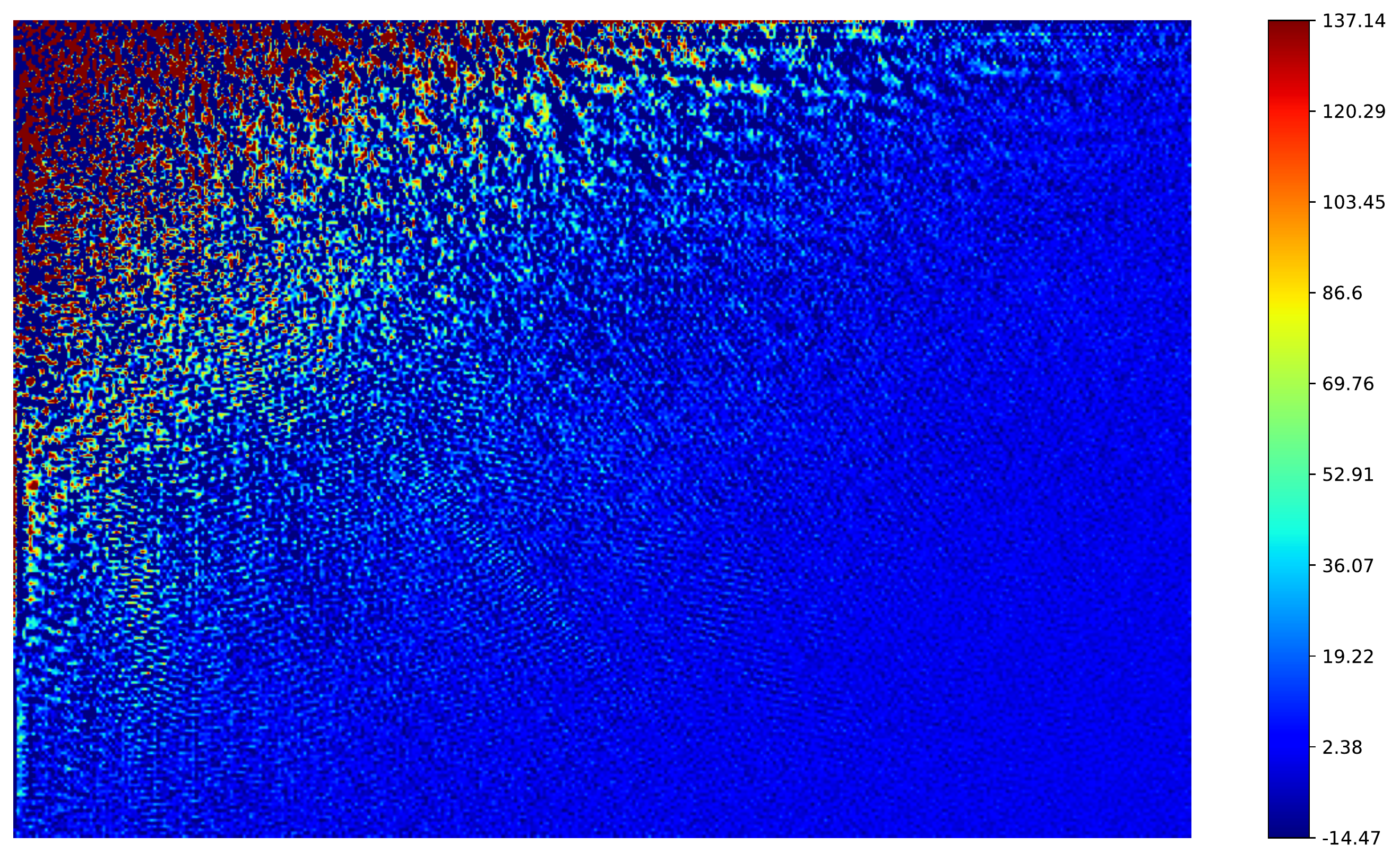}
    \end{minipage}
    \begin{minipage}{0.24\textwidth}
        \includegraphics[height=.65\linewidth]{plots/metrics_examples/ILSVRC2012_val_00041944.JPEG}  
         \includegraphics[height=.65\linewidth]{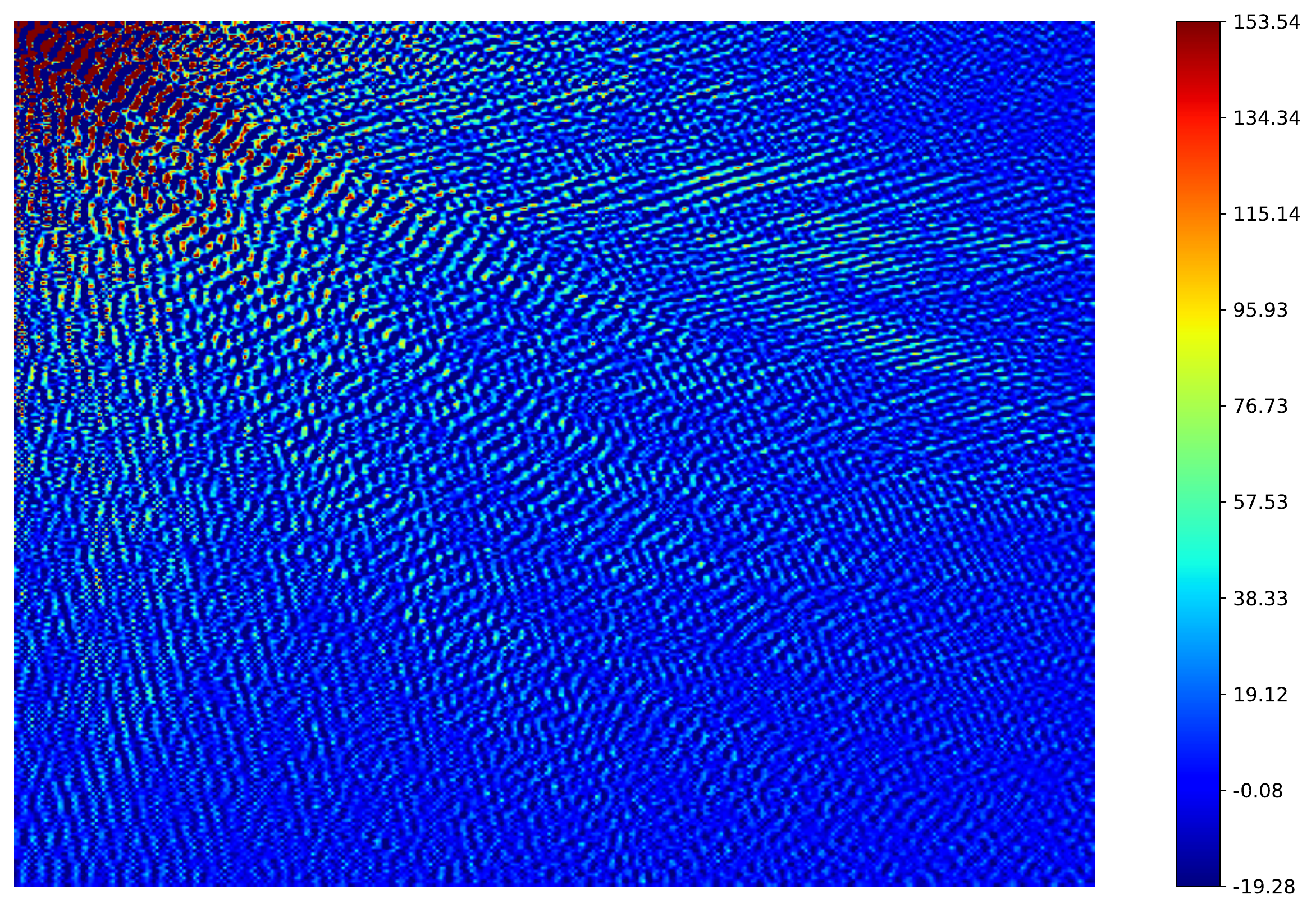} 
    \end{minipage}
    \begin{minipage}{0.24\textwidth}
        \includegraphics[height=.65\linewidth]{plots/metrics_examples/1__2008_006164.jpg}
         \includegraphics[height=.65\linewidth]{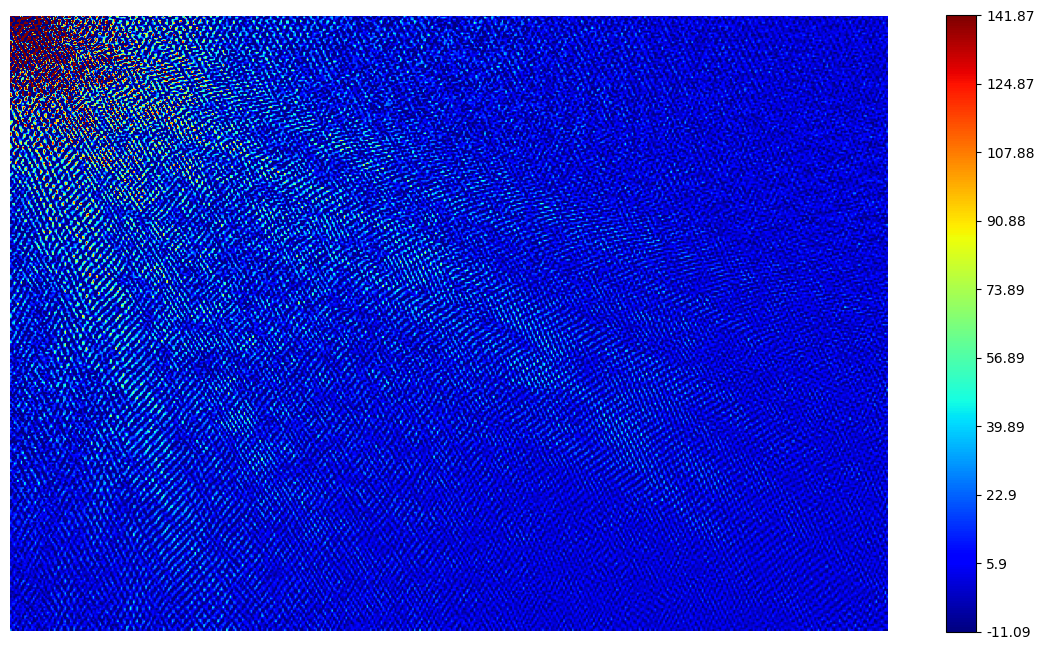}
    \end{minipage}
    \begin{minipage}{0.24\textwidth}
        \includegraphics[height=.65\linewidth]{plots/metrics_examples/19_15_14_0__2008_004506.jpg}
         \includegraphics[height=.65\linewidth]{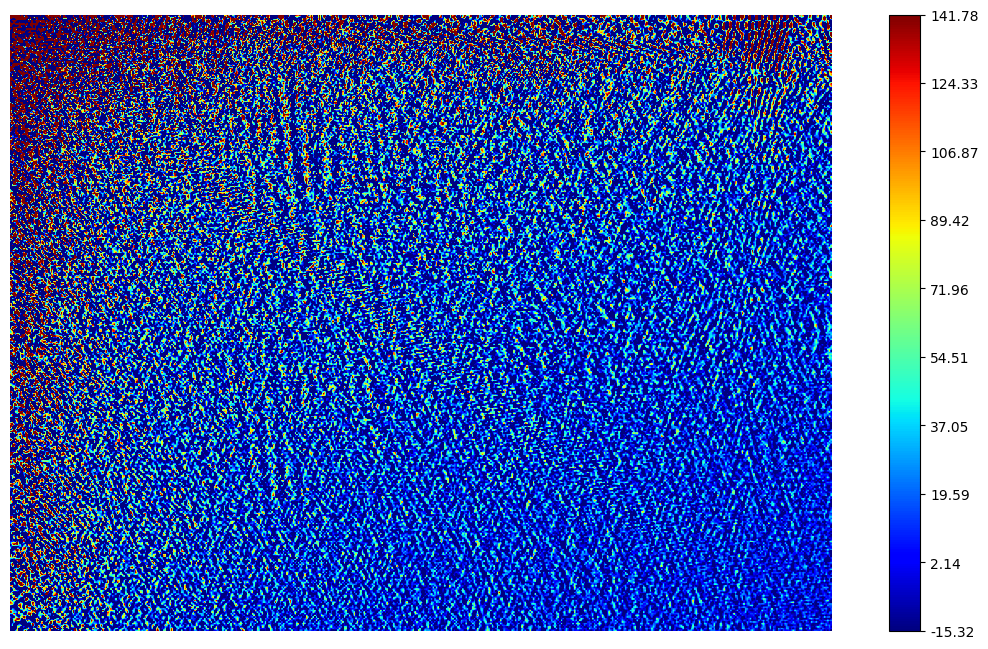}
    \end{minipage}
	\caption{Visualization of DCT matrix on ImageNet and Pascal examples}
	\label{fig:dct_coeffs}
\end{figure*}
Similar to fig. 3 of the main body, which visualizes the computed dataset metrics on two selected images from the Pascal dataset, we also selected an 'easy' and 'difficult' image from ImageNet to visualize the metrics in \cref{fig:metrics_imagenet}, as well as computed the DCT coefficients matrix for both chosen Pascal and ImageNet examples in \cref{fig:dct_coeffs}. What we see is that the more cluttered the image, the more irregular the entropy and segment image becomes. Cluttered images lead to higher amount of edges, but the edge strenghts of non-cluttered ones can be more prominent, which in summation may lead to similar sum of edge strengths. The DCT coefficients in \cref{fig:dct_coeffs} show that the more clutter there is, the higher the coefficients in all directions. With less clutter, but more prominent horizontal or vertical variations in the image, like the wings of the bird, lead to higher values in the DCT coefficient matrix for these horizontal and vertical directions.

\begin{figure*}
    \centering
    \begin{minipage}[t]{0.32\textwidth}
	\includegraphics[width = 0.99 \textwidth]{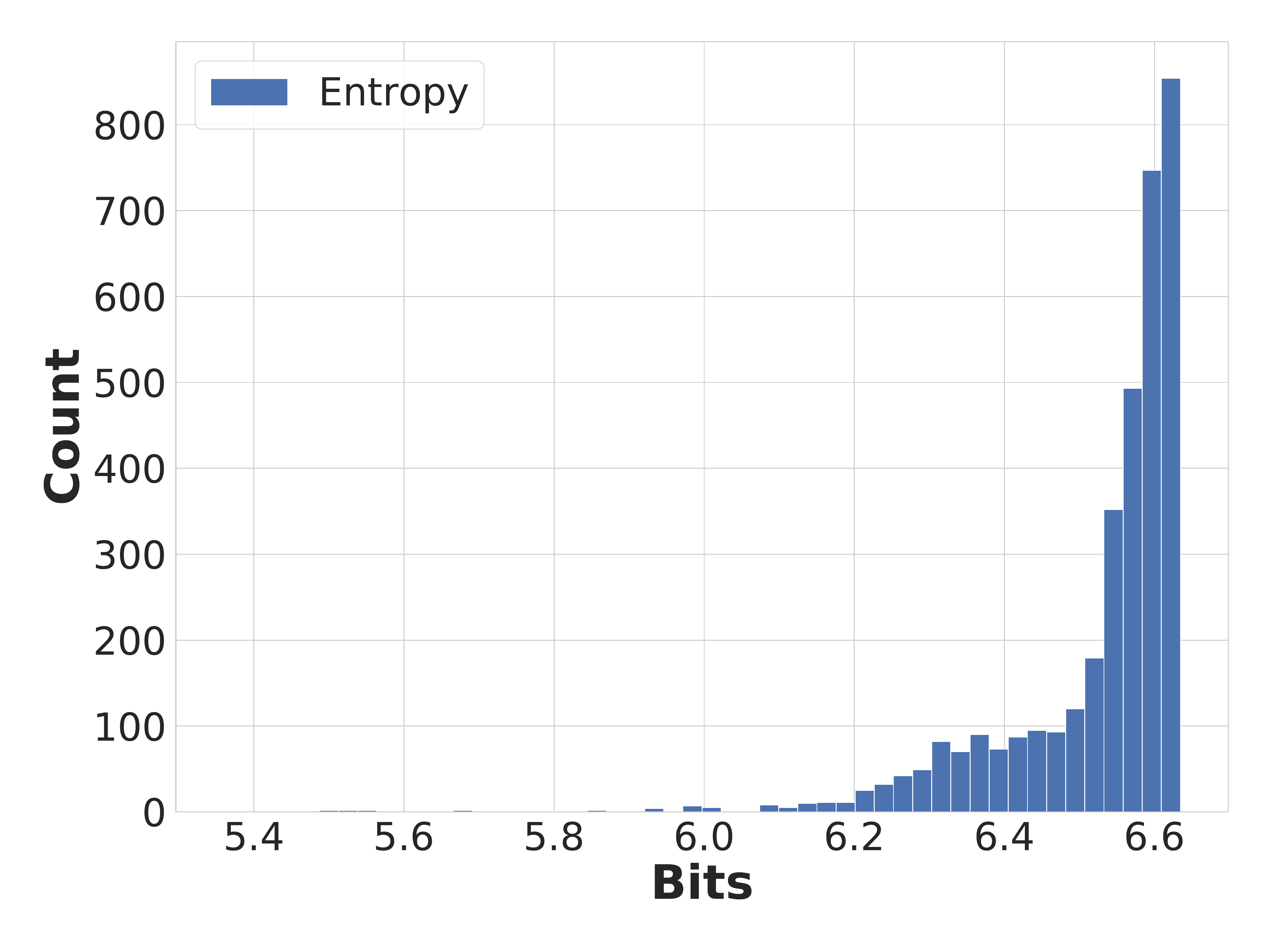}
	\, 
	\includegraphics[width = 0.99 \textwidth]{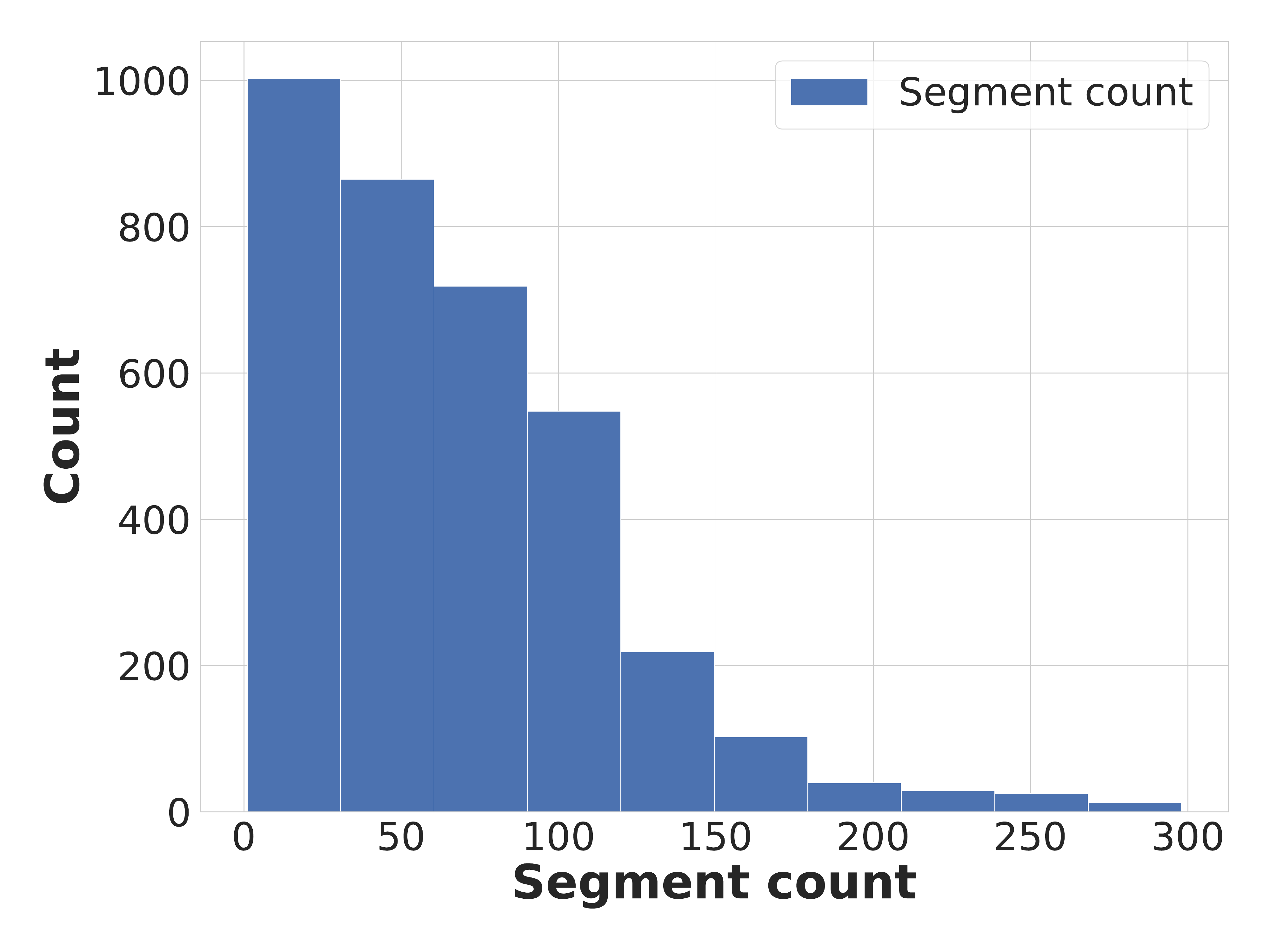}
	\,
	\includegraphics[width = 0.99 \textwidth]{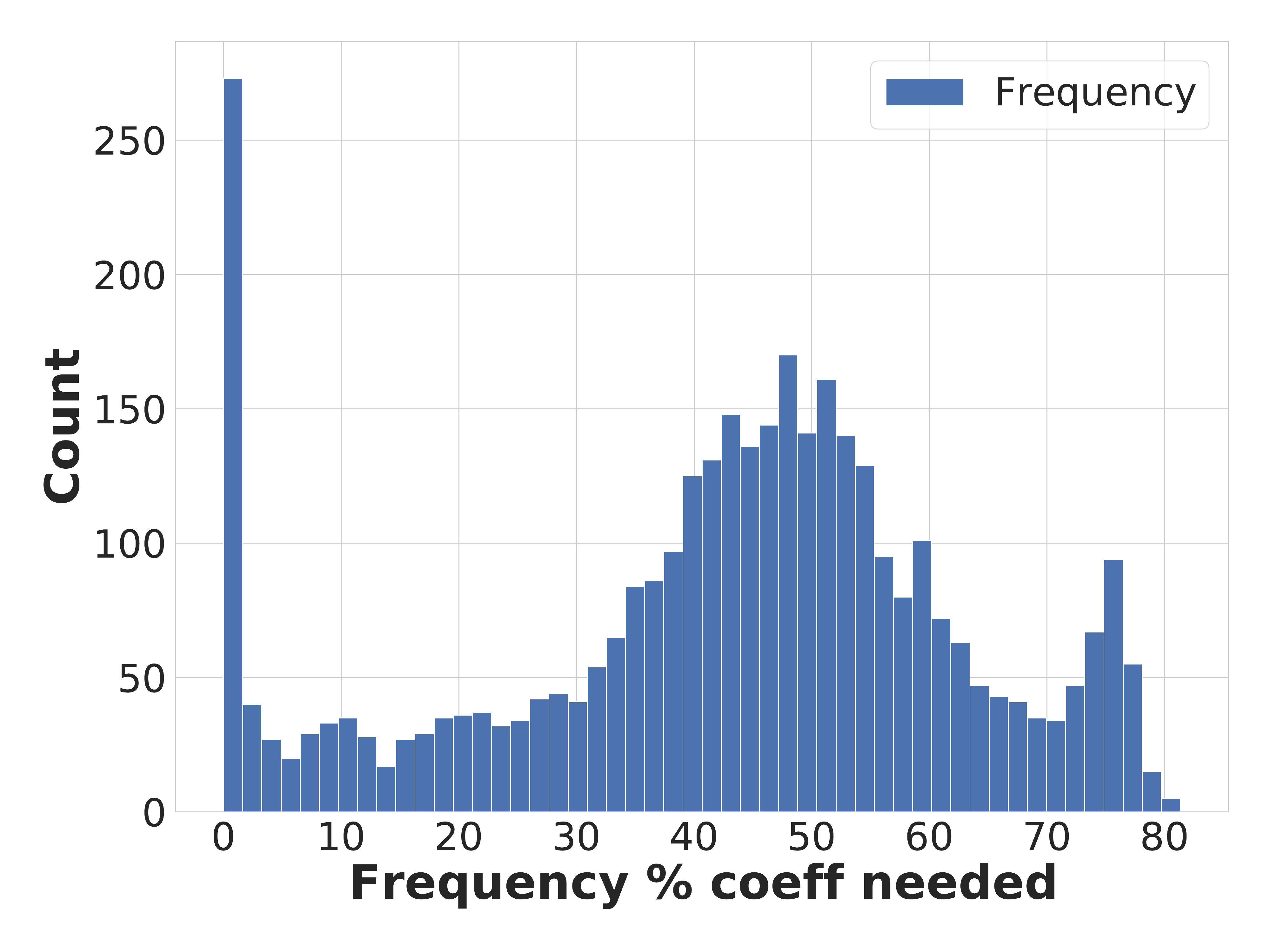}
	\\
	\includegraphics[width = 0.99 \textwidth]{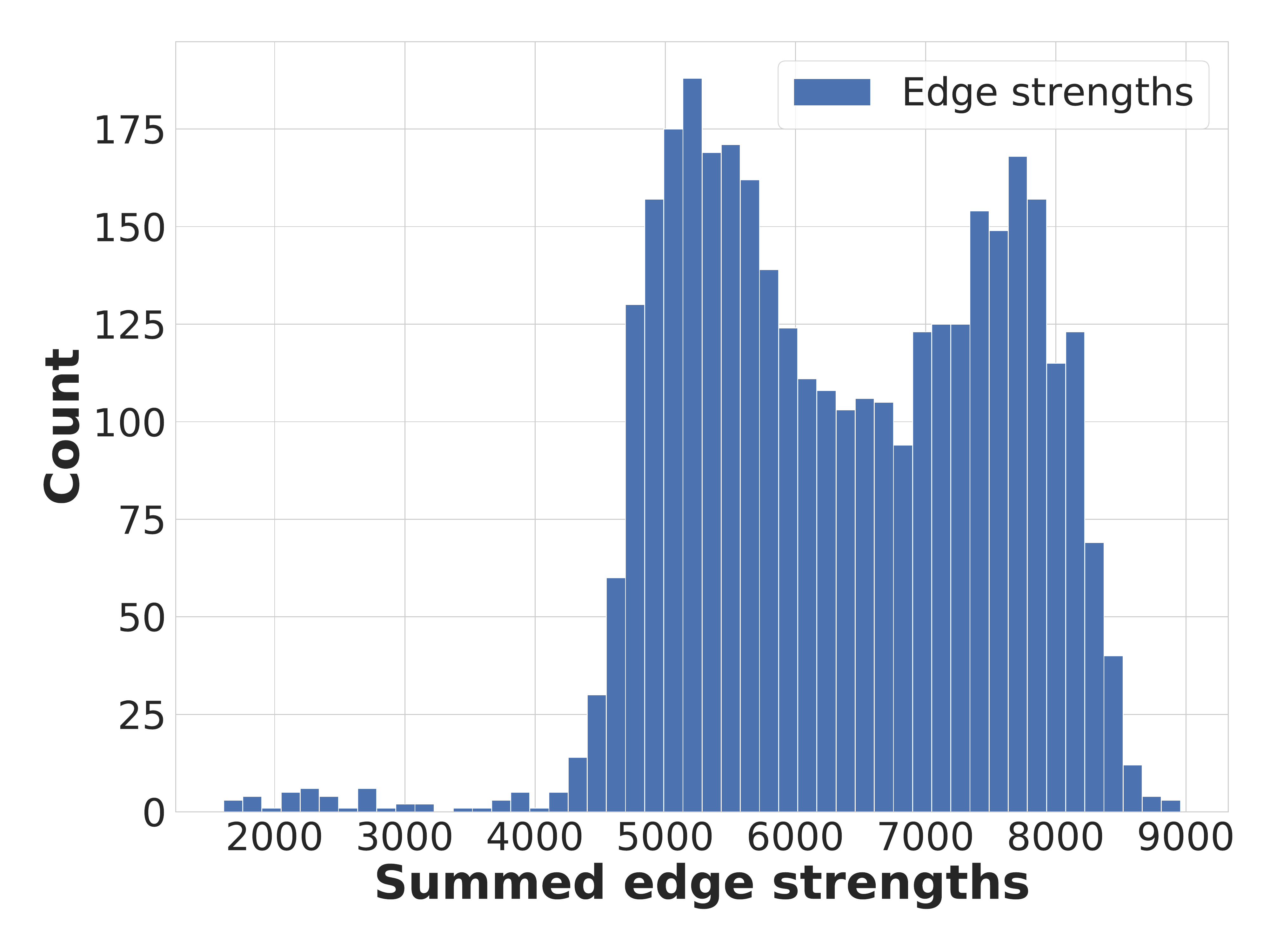}
	\caption{\textbf{KTH-TIPS2b}: Dataset metrics histograms on \emph{train set}}
	\label{kth_tips_hist}
	\end{minipage}
	\begin{minipage}[t]{0.32\textwidth}
	 \includegraphics[width = 0.99 \textwidth]{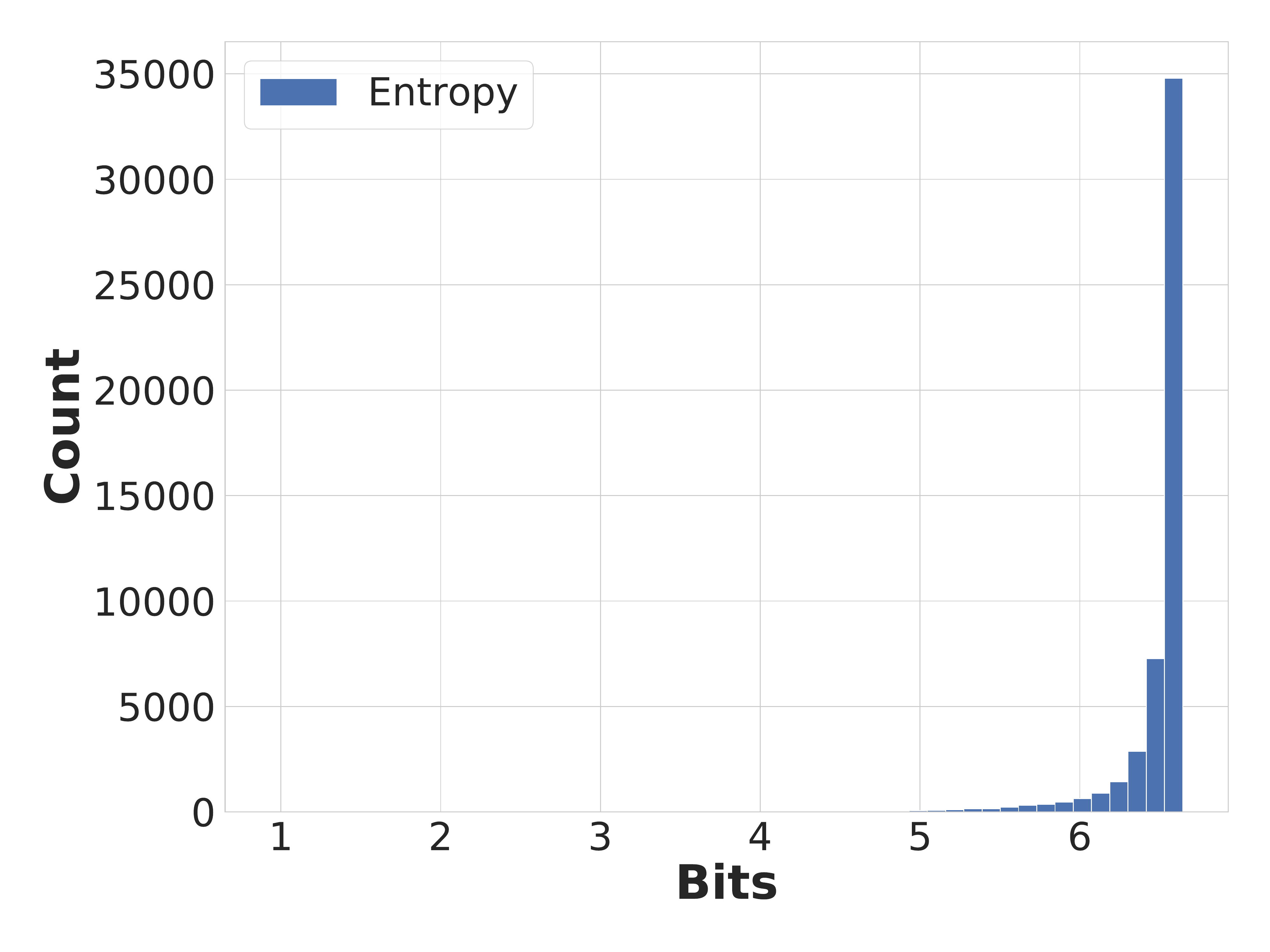}
	\, 
	\includegraphics[width = 0.99 \textwidth]{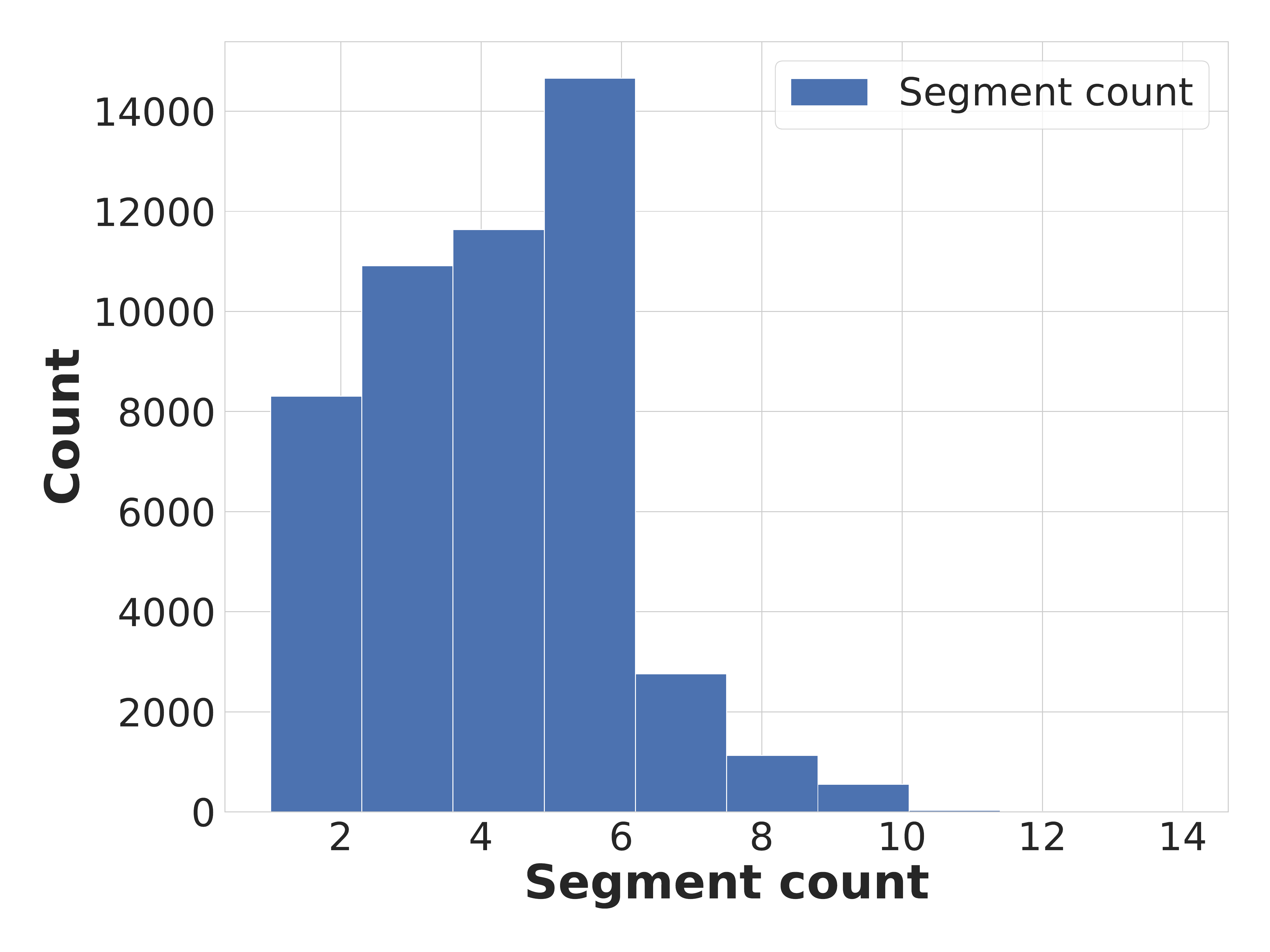}
	\, 
	\includegraphics[width = 0.99 \textwidth]{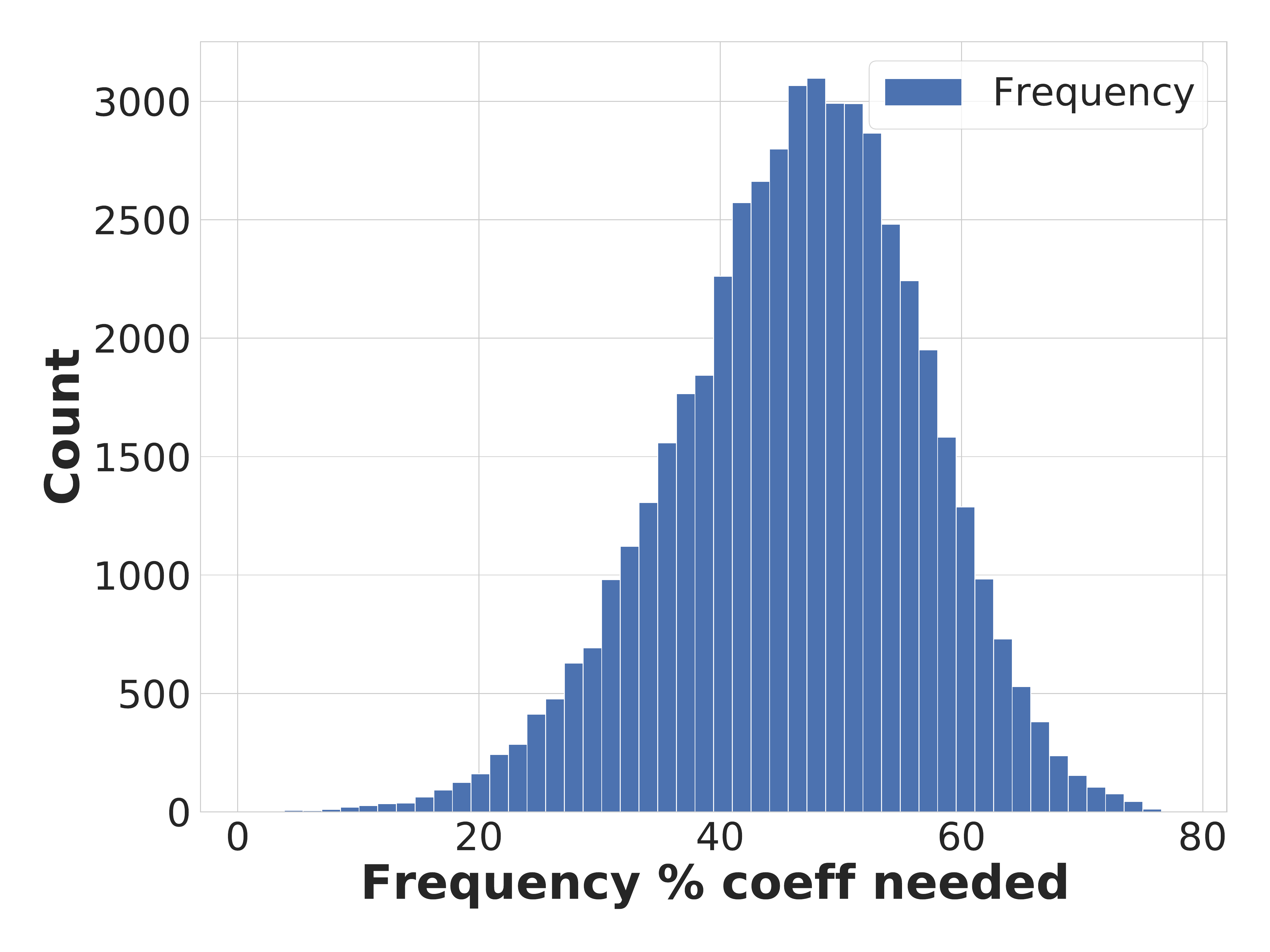}
	\\
	\includegraphics[width = 0.99 \textwidth]{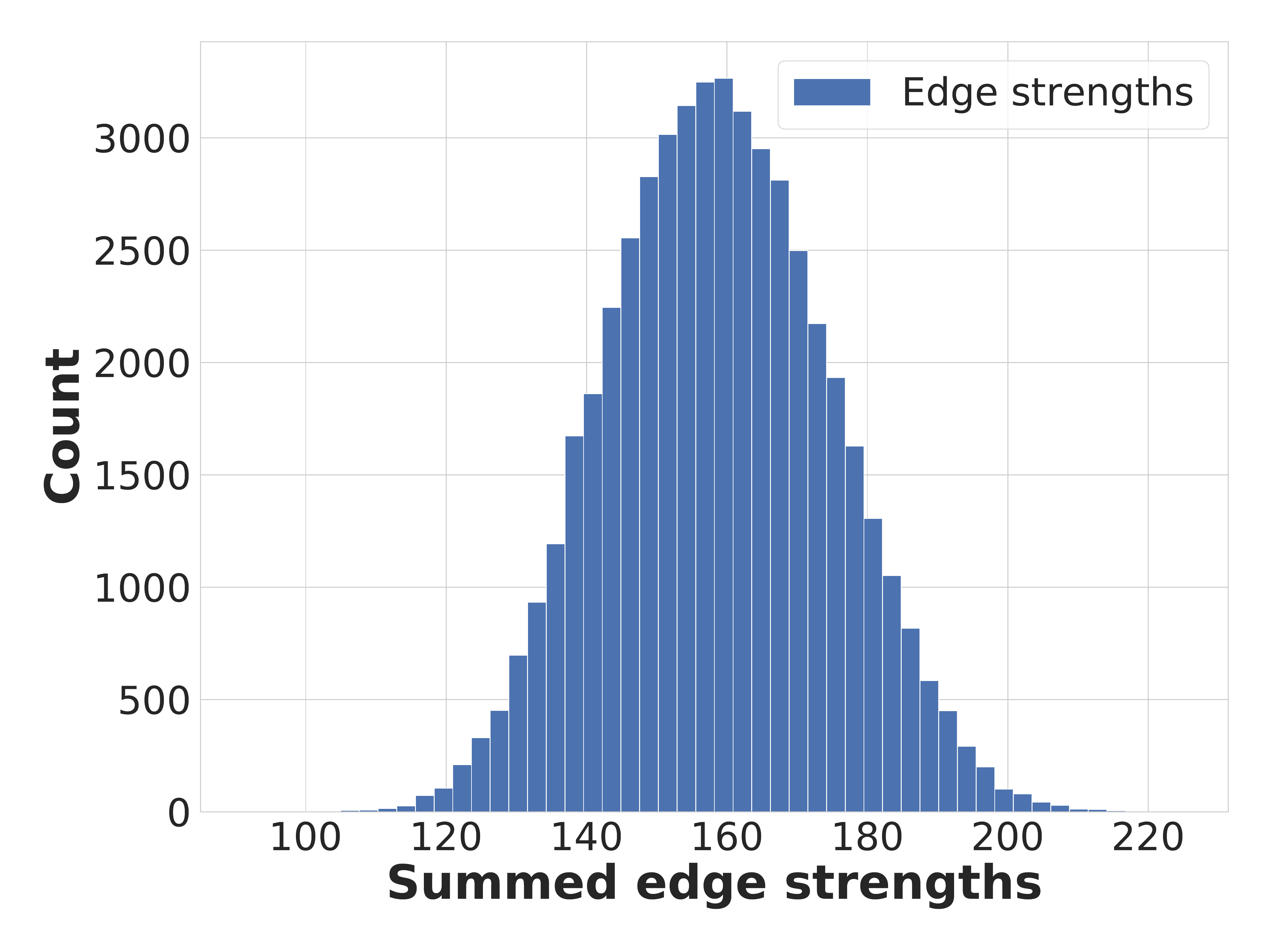}
	\, 
	\includegraphics[width = 0.99 \textwidth]{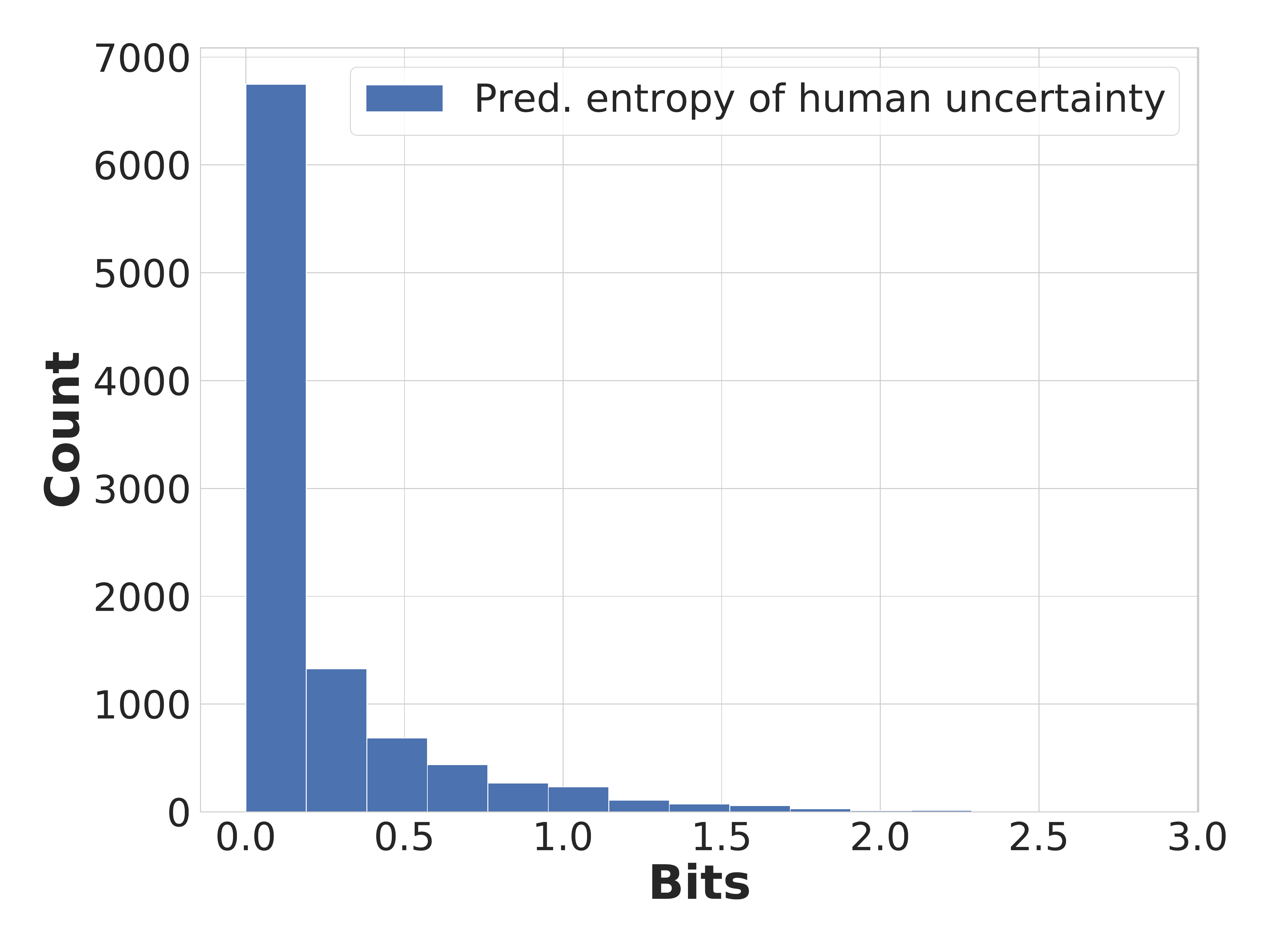}
	\caption{\textbf{CIFAR10}: Dataset metrics histograms}
	\label{cifar_hist}
	\end{minipage}
	\begin{minipage}[t]{0.32\textwidth}
	 \centering
	\includegraphics[width = 0.99 \textwidth]{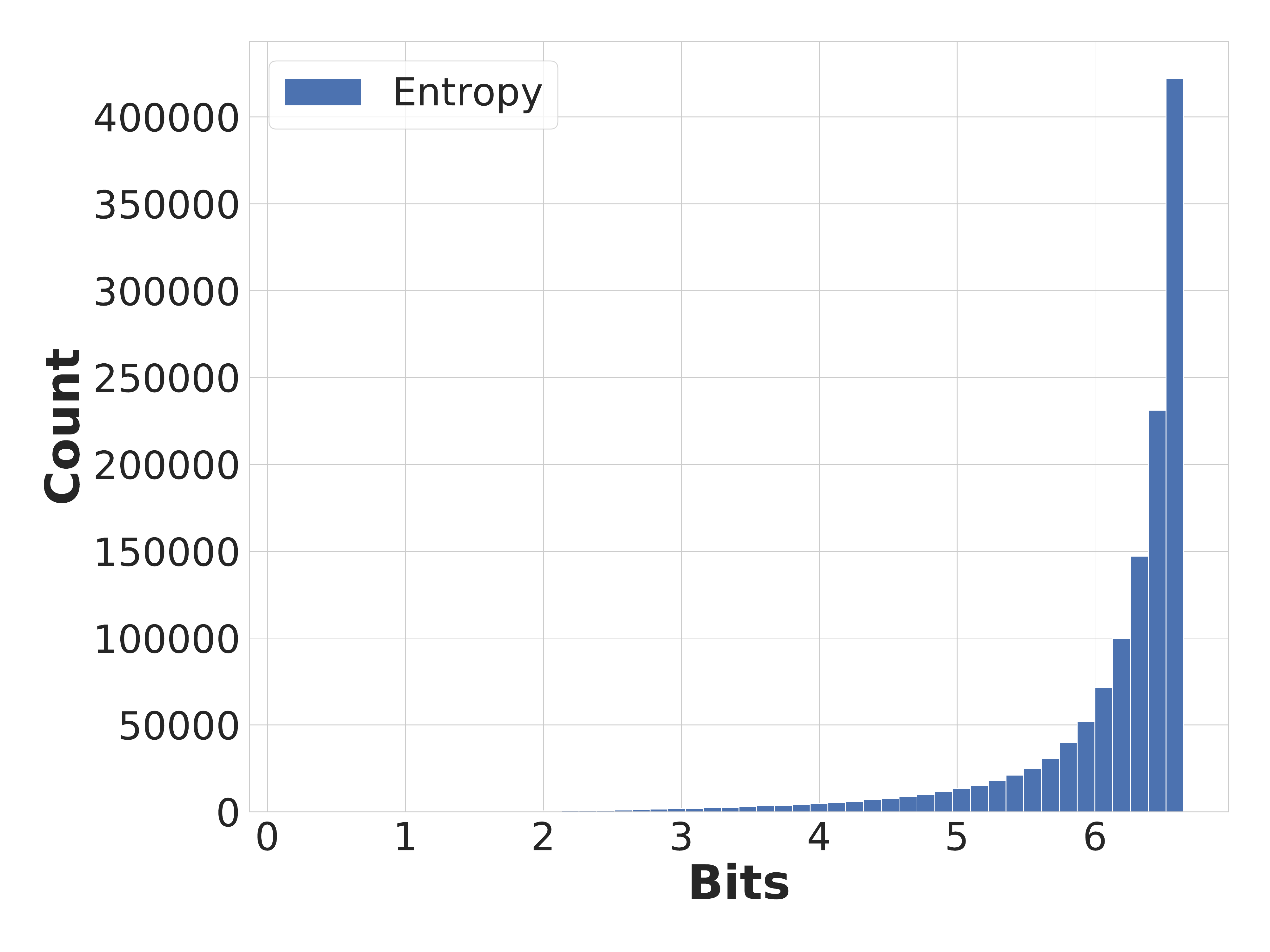}
	\, 
	\includegraphics[width = 0.99 \textwidth]{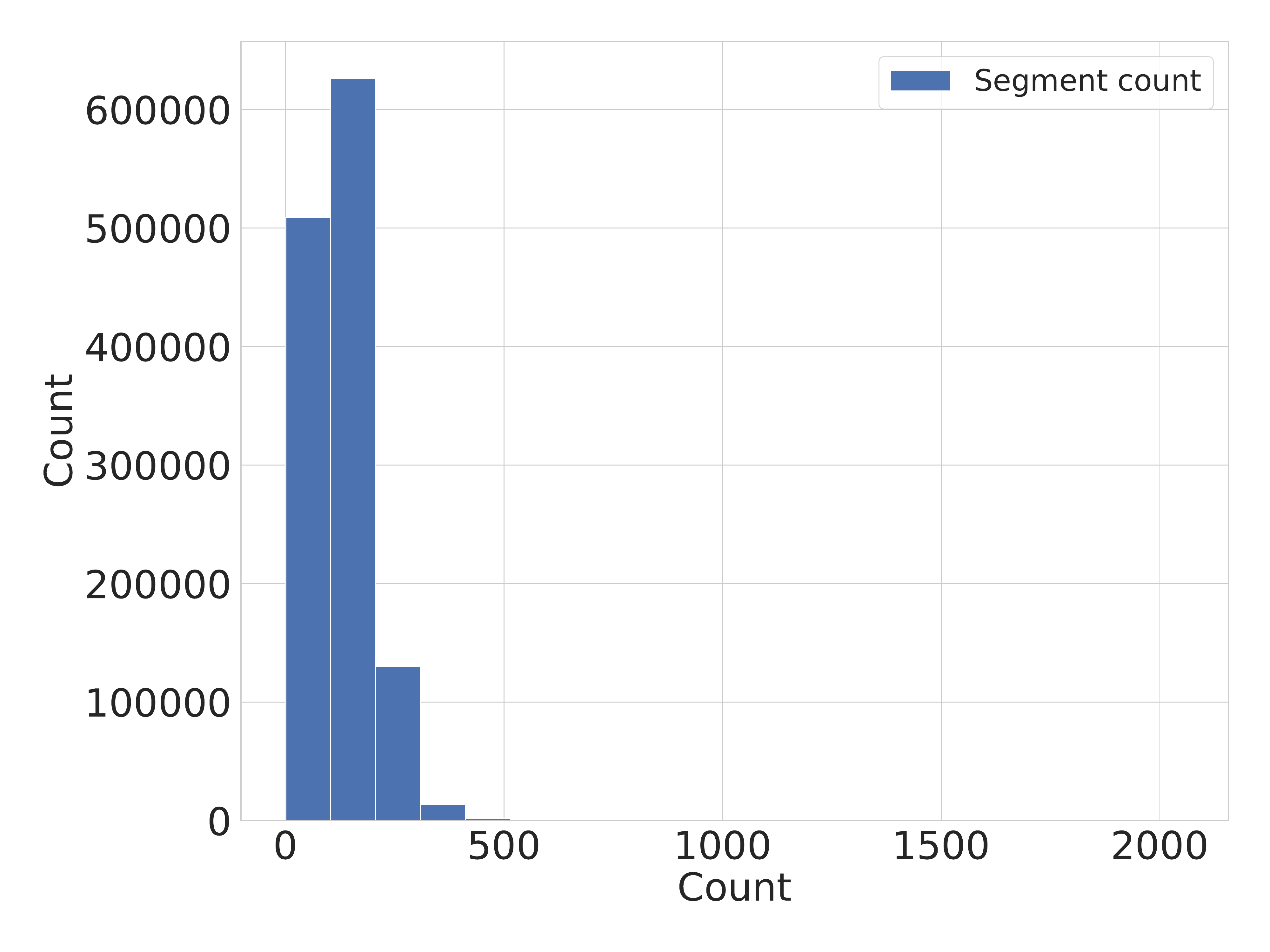}
	\, 
	\includegraphics[width = 0.99 \textwidth]{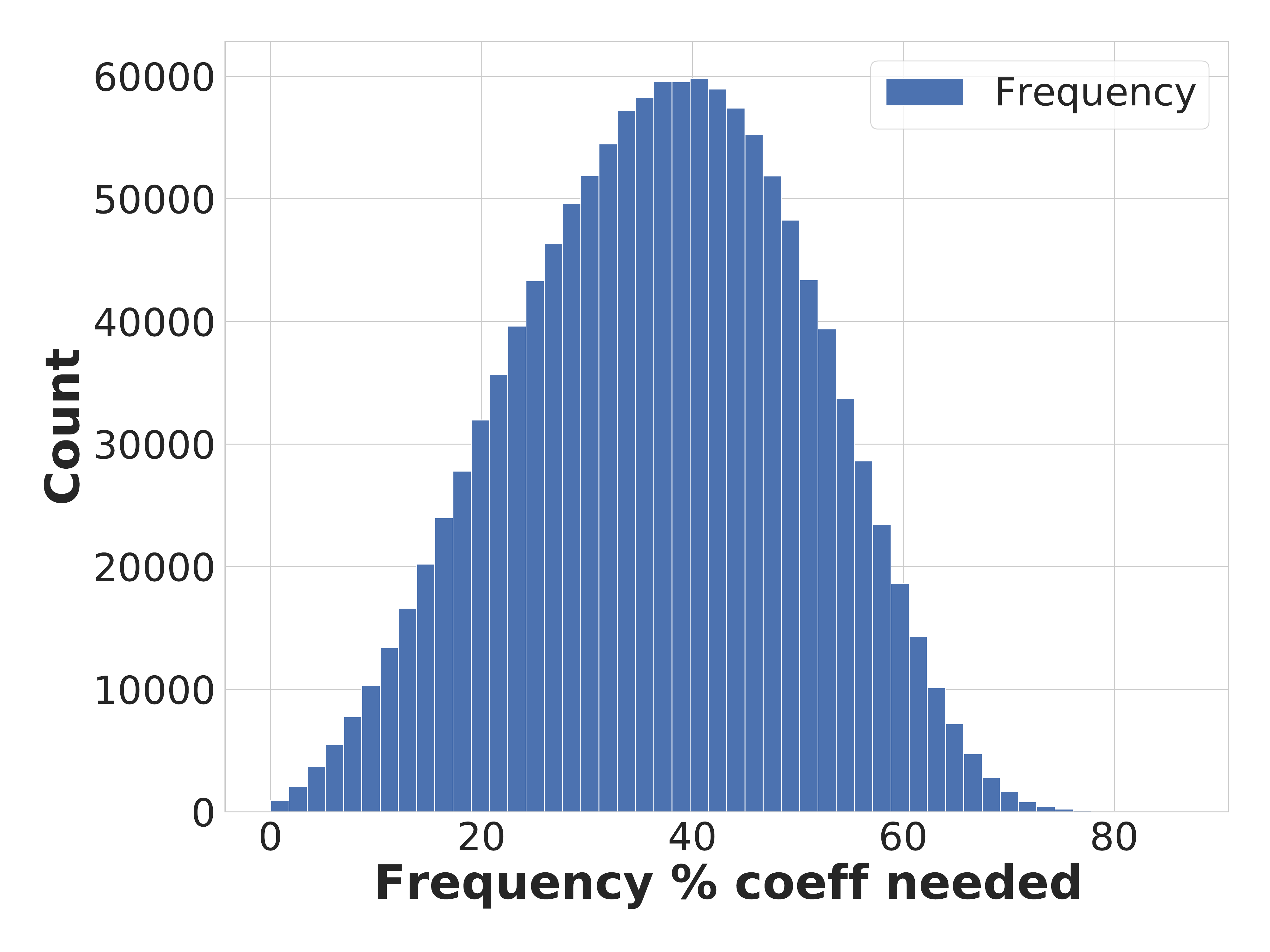}
	\caption{\textbf{ImageNet}: Dataset metrics histograms on \emph{train set}}
	\label{imagenet_hist}
	\end{minipage}
	\caption{Dataset metrics histograms on \emph{train set} for \textbf{KTH-TIPS2b}, \textbf{CIFAR10} and \textbf{ImageNet}, as well as for the \emph{Pred. entropy of human uncertainty} on the \emph{test set} of \textbf{CIFAR10}}
\end{figure*}

\begin{figure*}
    \centering
	\includegraphics[width = 0.31 \textwidth]{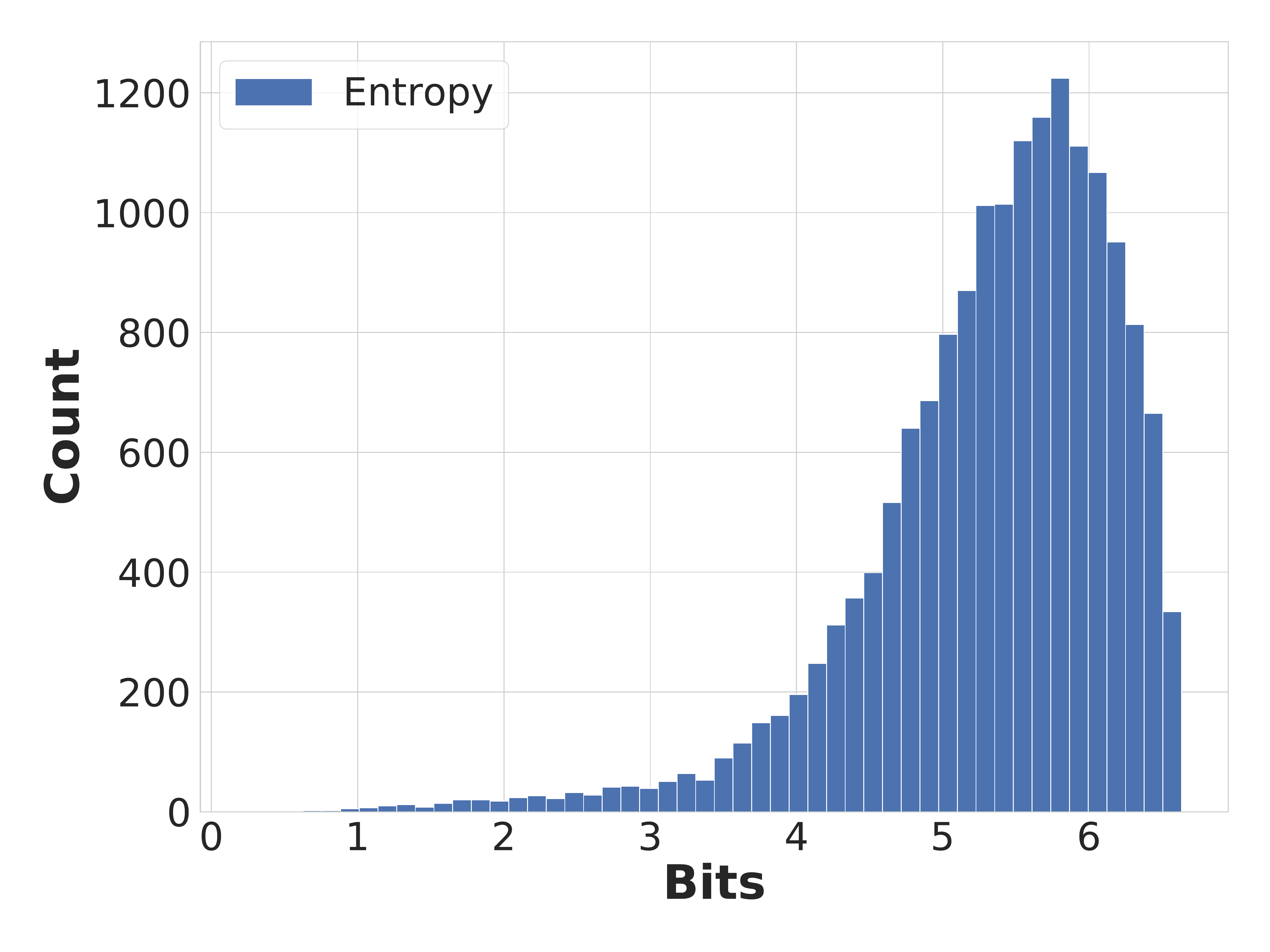}
	\, 
	\includegraphics[width = 0.31 \textwidth]{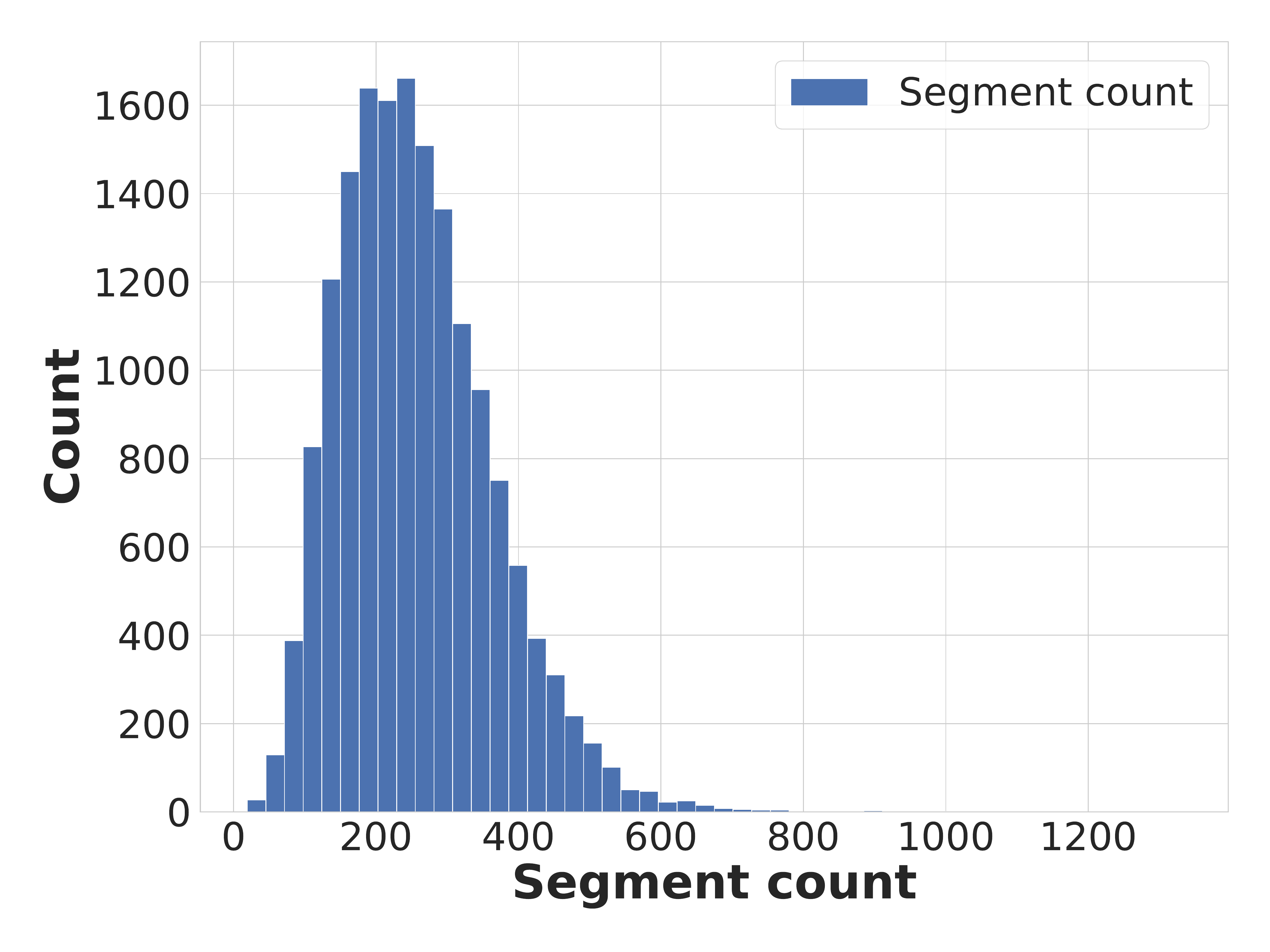}
	\, 
	\includegraphics[width = 0.31 \textwidth]{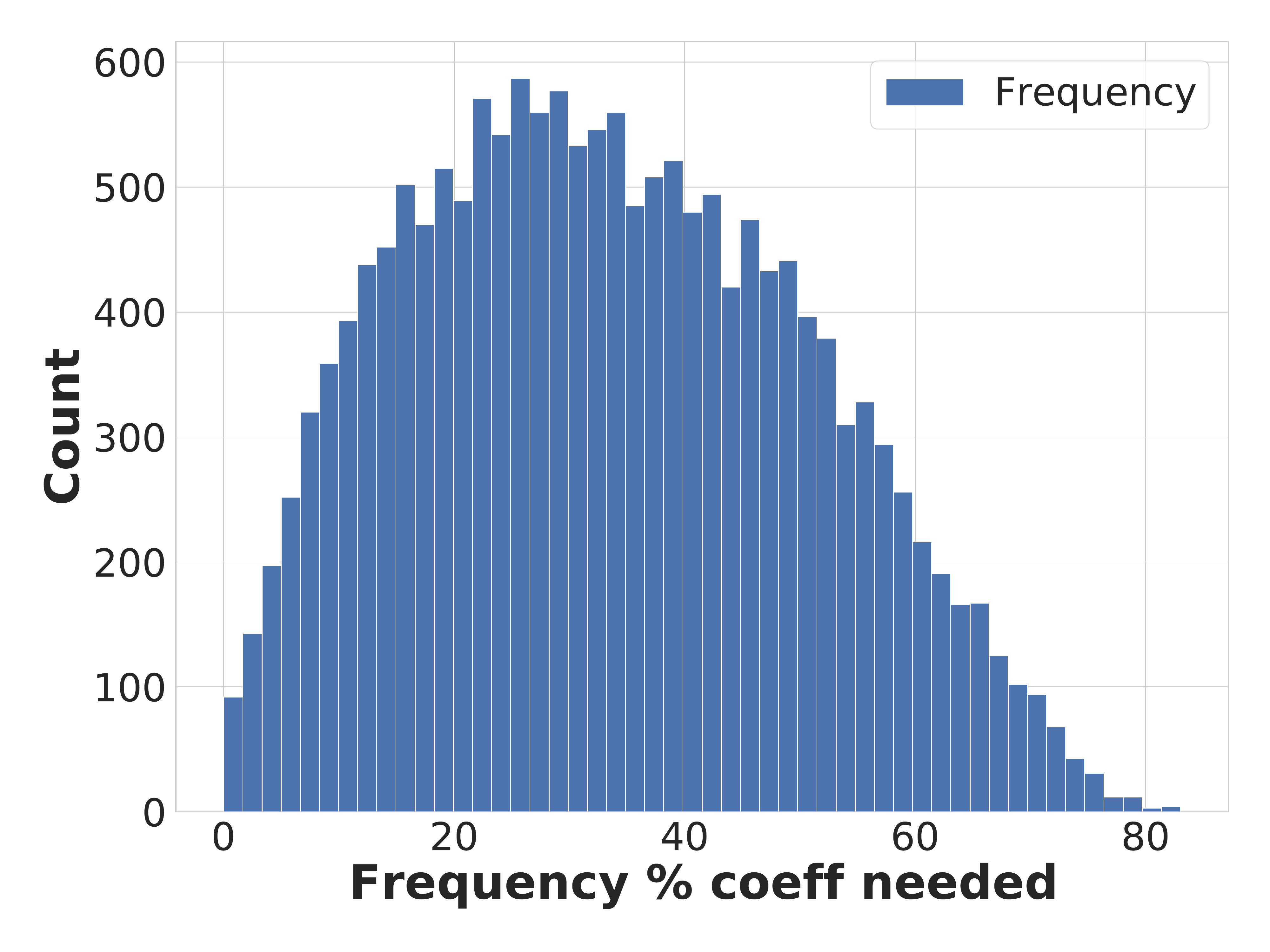}
	\\
	\includegraphics[width = 0.31 \textwidth]{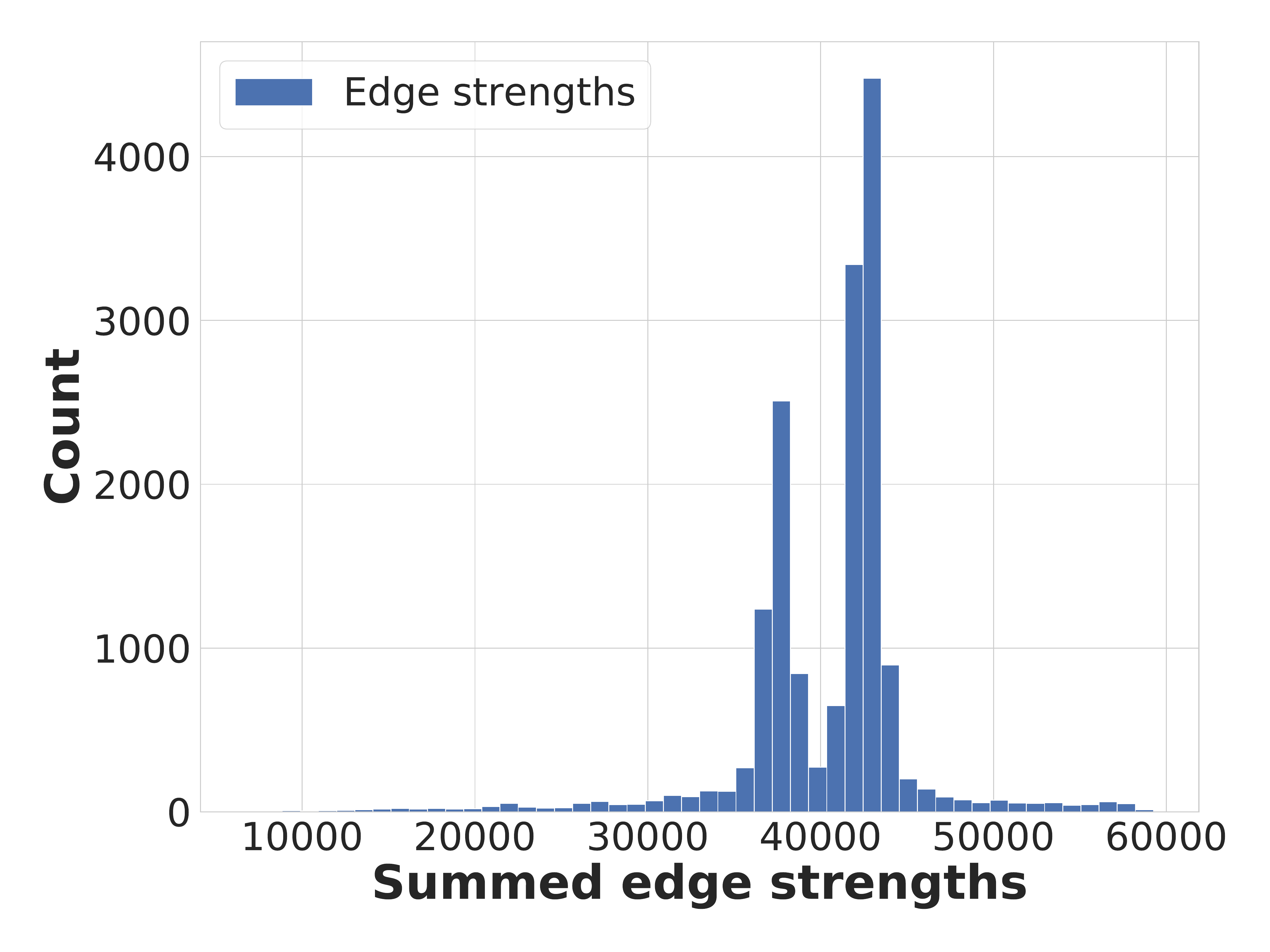}
	\, 
	\includegraphics[width = 0.31 \textwidth]{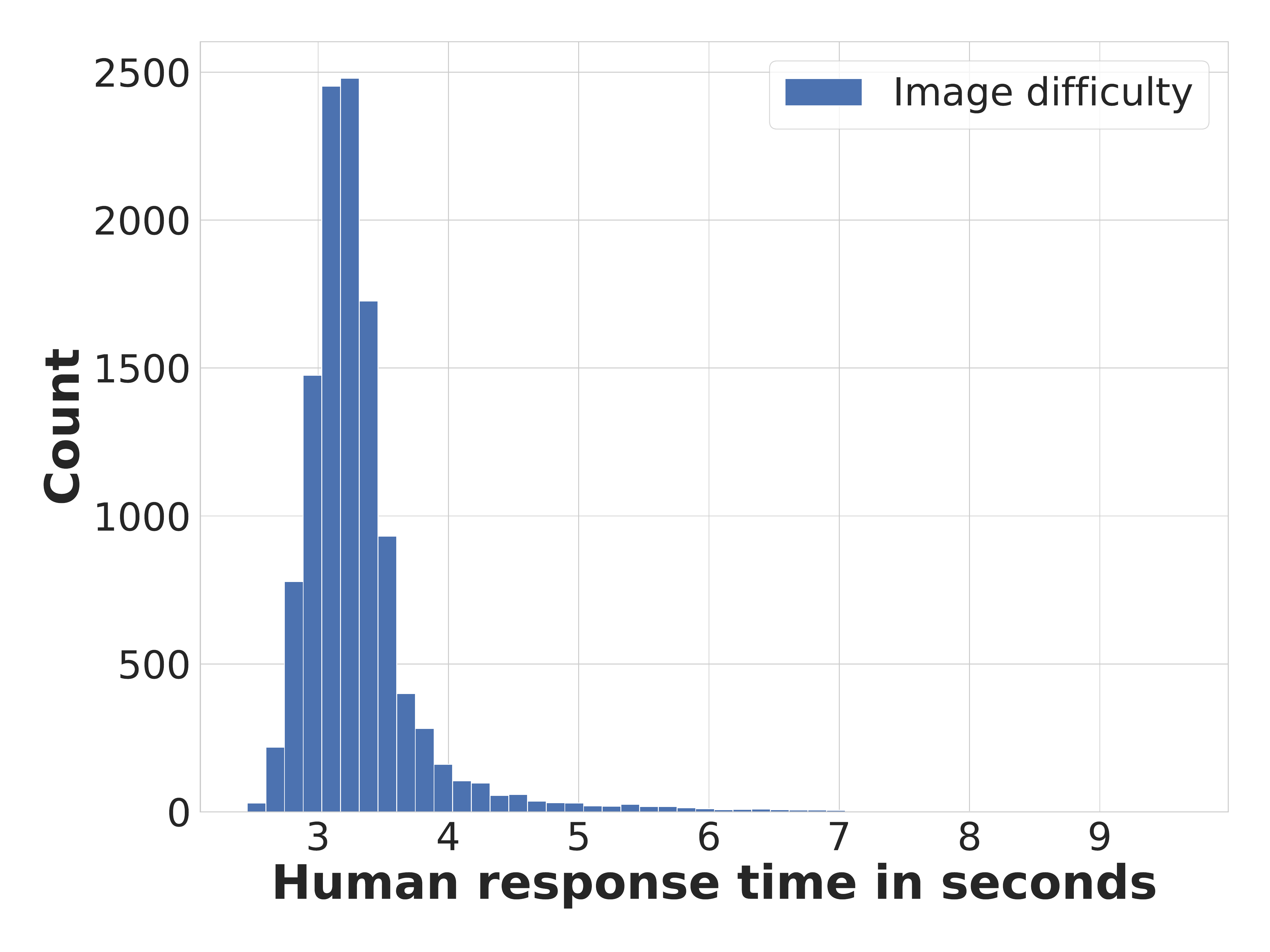}
	\,
	\includegraphics[width = 0.31 \textwidth]{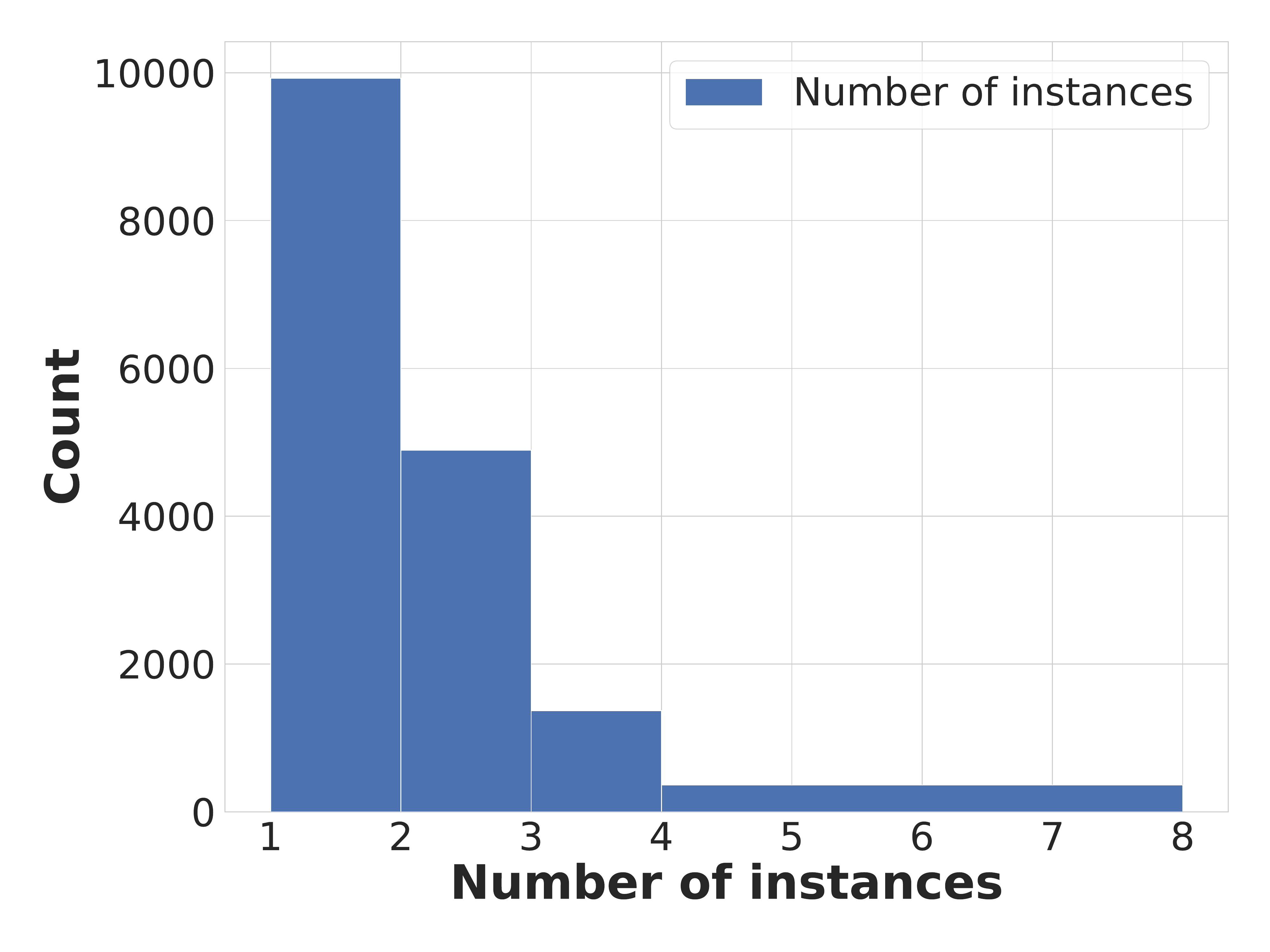}
	\\
	\includegraphics[width = 0.31 \textwidth]{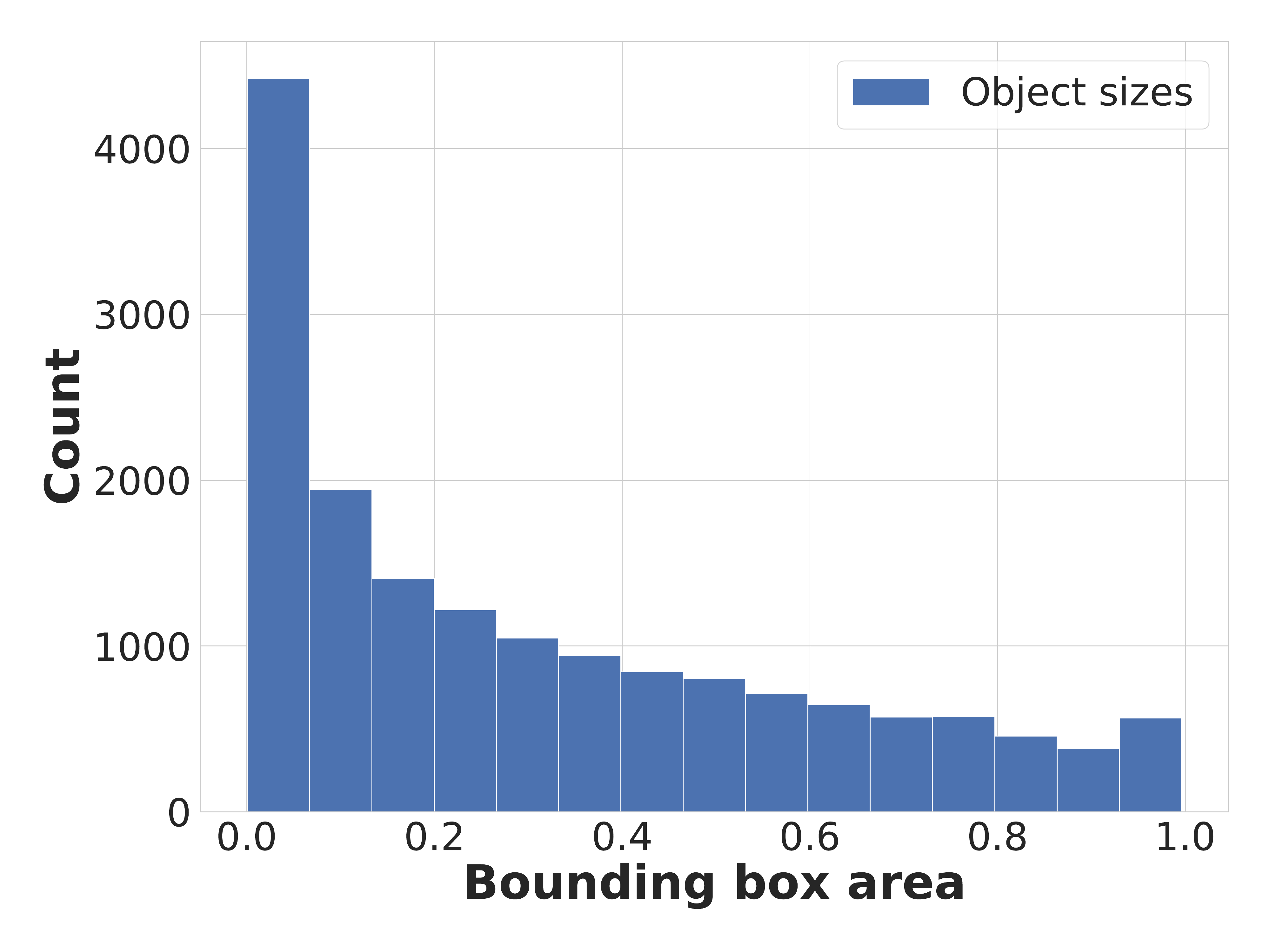}
	\caption{\textbf{Pascal}: Dataset metrics histograms on \emph{train set}}
	\label{pascal_hist}
\end{figure*}

\section{Dataset metrics histograms}
In order to assess the relevance of the results reported in the main body, we computed the histograms of the dataset metrics for CIFAR10 (\cref{cifar_hist}), Pascal (\cref{pascal_hist}), KTH-TIPS2b ( \cref{kth_tips_hist}) and ImageNet (\cref{imagenet_hist}) train sets. The \textbf{CIFAR10} histograms show that the frequency and edge strengths distributions are Gaussian, while the entropy and segment count are more or less centered on one value. Particularly for the metrics, which do not recognizably follow a certain distribution, the interpretation of the correlations is more difficult. The histogram for the predictive entropy of human uncertainty is not for the train, but for the test set, the corresponding correlation is in \cref{cifar_pred_uncertainty}. We see that predictive entropy is low for most instances.

The \textbf{ImageNet} histograms in \cref{imagenet_hist} resemble those of CIFAR10 more than those of KTH-TIPS2b or Pascal, in that there is no skew of the frequency Gaussian and the segment count distribution is irregular-shaped.

The \textbf{KTH-TIPS2b} histograms in \cref{kth_tips_hist} are more nuanced. The frequency and edge strengths distributions show several peaks, while the entropy and segment count exhibit an exponential course.

The \textbf{Pascal} histograms in \cref{pascal_hist} are skewed Gaussians for entropy, segment count, frequency and human response time, multi-peak Gaussian for edge strenghts, similar to KTH-TIPS2b, as well as more or less centered around one value for number of instances and bounding box area.

%
%
\bibliographystyle{splncs04}
\bibliography{references}
\end{document}